\documentclass[10pt,twocolumn,letterpaper]{article}

\usepackage{wacv}               % To produce the CAMERA-READY version
\usepackage[export]{adjustbox}
\usepackage[linesnumbered,lined,longend,algoruled]{algorithm2e}
\usepackage{array}
\usepackage{amsmath}
\usepackage{amssymb}
\usepackage{booktabs}
\usepackage{fancyhdr}
\usepackage{graphicx}
\usepackage{tikz}
\usepackage{multirow}
\usepackage{nicefrac}
\usepackage{silence}
\usepackage{soul}

\usepackage[linesnumbered,lined,longend,algoruled]{algorithm2e}
\usepackage{comment}
\usepackage{fouridx}
\usepackage{mathtools}
\SetKwComment{AComment}{/* }{ */}

\usetikzlibrary{spy, calc, tikzmark}
\usepackage[pagebackref,breaklinks,colorlinks]{hyperref}

\usepackage[capitalize]{cleveref}
\crefname{section}{Sec.}{Secs.}
\Crefname{section}{Section}{Sections}
\crefname{table}{Tab.}{Tabs.}
\Crefname{table}{Table}{Tables}
\crefname{algorithm}{Alg.}{Algs.}
\Crefname{algorithm}{Algorithm}{Algorithms}

\def\wacvPaperID{395}
\def\confName{WACV}
\def\confYear{2025}

\mathchardef\mhyphen="2D
\newcommand{\tref}{{t_{\text{ref}}}}
\newcommand{\origevents}{{\mathcal{E}}}
\newcommand{\warpedevents}{{\mathcal{E}_{\tref}'}}
\newcommand{\boxscaler}{0.92}
\newcommand{\labelscaler}{0.18}
\newcommand{\uln}{\underline}

\DeclareMathOperator*{\argmax}{arg\,max}

\makeatletter
\newcommand{\removelatexerror}{\let\@latex@error\@gobble}
\makeatother

\newcommand\copyrighttext{%
\footnotesize \textcopyright 2025 IEEE. Personal use of this material is
permitted. Permission from IEEE must be obtained for all other uses, in any
current or future media, including reprinting/republishing this material for
advertising or promotional purposes, creating new collective works, for resale
or redistribution to servers or lists, or reuse of any copyrighted component of
this work in other works.
}
\newcommand\copyrightnotice{%
\begin{tikzpicture}[remember picture,overlay]
\node[anchor=south,yshift=15pt] at (current page.south) {\fbox{\parbox{\dimexpr\textwidth-\fboxsep-\fboxrule\relax}{\copyrighttext}}};
\end{tikzpicture}%
}

\makeatletter
\def\thanks#1{\protected@xdef\@thanks{\@thanks
  \protect\footnotetext{#1}}}
\makeatother

\begin{document}
\pagestyle{fancy}
\fancyhf{}
\renewcommand{\headrulewidth}{0pt}%
\renewcommand{\footrulewidth}{0pt}%
\fancyfoot[C]{\thepage}

% ---------------------------------------------------------------
\title{Secrets of Edge-Informed Contrast Maximization for Event-Based Vision} 

\author{
Pritam P. Karmokar$^{*}$, Quan H. Nguyen$^{*}$, and William J. Beksi\\
The University of Texas at Arlington\\
Arlington, TX, USA\\
{\tt\small \{pritam.karmokar,quan.nguyen4\}@mavs.uta.edu, william.beksi@uta.edu}
\thanks{$^{*}$ Indicates equal contribution.}
}

\maketitle
\copyrightnotice

% --------------------------------------------------------------------------------
% ABSTRACT
% --------------------------------------------------------------------------------
\begin{abstract}
Event cameras capture the motion of intensity gradients (edges) in the image
plane in the form of rapid asynchronous events. When accumulated in 2D
histograms, these events depict overlays of the edges in motion, consequently
obscuring the spatial structure of the generating edges. Contrast maximization
(CM) is an optimization framework that can reverse this effect and produce
sharp spatial structures that resemble the moving intensity gradients by
estimating the motion trajectories of the events. Nonetheless, CM is still an
underexplored area of research with avenues for improvement. In this paper, we
propose a novel hybrid approach that extends CM from uni-modal (events only) to
bi-modal (events and edges). We leverage the underpinning concept that, given a
reference time, optimally warped events produce sharp gradients consistent with
the moving edge at that time. Specifically, we formalize a correlation-based
objective to aid CM and provide key insights into the incorporation of
multiscale and multireference techniques. Moreover, our edge-informed CM method
yields superior sharpness scores and establishes new state-of-the-art event
optical flow benchmarks on the MVSEC, DSEC, and ECD datasets.
\end{abstract}

%%%%%%%%%%%%%%%%%%%%%%%%%%%%%%%%%%%%%%%%%%%%%%%%%%%%%%%%%%%%%%%%%%%%%%%%%%%%%%%%
% Figure: Intro
%\input{figures/intro}
%%%%%%%%%%%%%%%%%%%%%%%%%%%%%%%%%%%%%%%%%%%%%%%%%%%%%%%%%%%%%%%%%%%%%%%%%%%%%%%%

% ------------------------------------------------------------------------------
% INTRODUCTION
% ------------------------------------------------------------------------------
\section{Introduction}
\label{sec:intro}
An event camera is a bio-inspired device that is fundamentally different from a
conventional frame-based camera in both its functionality and how its output
data is processed. Instead of capturing frames at regular intervals, event
cameras asynchronously measure per-pixel brightness changes. Such signals
encode scene motion, and they are particularly useful for optical flow
estimation. However, estimating event-based optical flow is challenging because
of the temporally dense yet spatially sparse nature of events. Many methods
have been proposed for event-based optical flow, which can be broadly
summarized as follows: (i) applying classical frame-based algorithms; (ii)
studying the principles and characteristics of event data; (iii) learning
correlations between event data and supervisory signals. 

Event contrast maximization (CM)
\cite{gallego2018unifying,stoffregen2019analysis, shiba2022secrets} is an
optimization framework to estimate the motion trajectories of events. In CM,
events are warped according to estimated motion parameters that optimize a
\emph{contrast} objective for a given reference time. Concretely, the
coordinates of the events are warped using motion parameters to obtain
\emph{warped events}, which are then used to construct an \emph{image of warped
events} (IWE). Typically, the \emph{contrast} objective represents the
sharpness of the IWE. 

Since events are generated mainly due to moving intensity gradients (\ie,
edges) across the image plane, each singular event can be associated with a
corresponding singular parent edgel (\ie, edge pixel). For a given reference
time, events warped with optimal motion parameters manifest good spatial
alignment with corresponding intensity gradients, leading events to spatially
coincide with their corresponding parent edgel. Consequently, there is a strong
spatial structure similarity between an IWE constructed using well-optimized
motion parameters and the respective edge image, which is the basis of our
work.

An \emph{image of unwarped events}\footnote{We refer to the IWE of zero-warped
events as the \emph{image of unwarped events}, not to be confused with inverse
warps.} constructed using the instantaneous output of an ideal event camera
should stencil out edge-fronts in motion. However, over larger intervals of
time this is captured as overlays of moving edge fronts smeared across the
image plane, thus obscuring the 2D visual appearance of the edges in the scene.
The motivation behind warping events is not only to obtain sharp edges in the
IWE, but also to have these IWE edges (i) located in the image plane at
appropriate coordinates corresponding to the warp reference time and (ii)
consistent with the intensity edges (\cref{fig:intro}).

In this work, we introduce a novel approach that makes use of both event and
frame modalities. Our method enforces \emph{contrast} and correlation
constraints in a model-based setting. To our knowledge, we are the first to
couple the two modalities and to leverage existing approaches under a unifying
model-based optical flow framework. Specifically, we unravel the \emph{secrets}
of using multiple scales, regularization, and sequential handover techniques
from frame-based computer vision to significantly improve event CM. Our
contributions are summarized as follows.
\begin{enumerate}
  \item We extend CM in a model-based setting by simultaneously maximizing the
  IWE \emph{contrast} and event-edge \emph{correlation} objectives.
  \item We refine existing multiscale and multireference techniques for the
  bi-modal case, as well as develop more sophisticated sequential processing
  strategies to improve convergence and enhance performance.
  \item We establish state-of-the-art event-based optical flow benchmarks via
  our hybrid framework on the MVSEC, DSEC, and ECD datasets.
\end{enumerate}
Our source code is available at \cite{eincm}.

% --------------------------------------------------------------------------------
% PRIOR WORK
% --------------------------------------------------------------------------------
\section{Prior Work}
\label{sec:prior_work}
% --------------------------------------------------------------------------------
\subsection{Model-Based Approaches}
\label{subsec:model-based-approaches}
Based on a brightness constancy assumption, Benosman \etal
\cite{benosman2012asynchronous} estimated spatial and temporal gradients in an
infinitesimal spatio-temporal neighborhood of the most recent events. This led
to an overdetermined system of linear equations to solve for the optical flow
vector (Lucas-Kanade). In a later work, Benosman \etal\cite{benosman2013event}
considered events in a spatio-temporal window as points in an $x$-$y$-$t$
coordinate system where a plane was fitted and its slopes in the $x$-$t$ and
$y$-$t$ cross sections encoded the normal flow. 

Akolkar \etal \cite{akolkar2020real} proposed a multiscale plane fitting method
that is more robust to the aperture problem. A sliding window
optimization-based technique developed by Bardow \etal
\cite{bardow2016simultaneous} jointly recovered the intensity frame and the
motion field. Brebion \etal \cite{brebion2021realtimeflow} applied a distance
transform inspired by Almatrafi \etal \cite{almatrafi2020distancesurface} to
construct an inverse exponential distance surface (\ie, a dense image-like
representation with edge emphasis) from events, then ultimately utilized
frame-based optical flow methods.

Conversely, Gallego \etal \cite{gallego2018unifying} relied on the concept of
warping events according to a parameterized motion model. This is known as
motion compensation (or CM) where the idea is to optimize the motion parameters
such that warped events achieve maximum alignment, thus resulting in an IWE
with higher \emph{contrast}. These parameters, however, may converge to a
global optimum that warps events to too few pixels (\ie, \emph{event collapse}
\cite{shiba2022eventcollapse}), which was mitigated by Shiba \etal
\cite{shiba2022secrets} using multiple warp reference times.

% --------------------------------------------------------------------------------
\subsection{Learning-Based Approaches}
\label{subsec:learning-based_approaches}
Learning-based methods have been driven by adaptations of frame-based neural
network architectures. These approaches require additional preprocessing steps
to turn event data into grid-like representations such as event frames
\cite{zhu2018evflownet,ye2020unsupervised}, event volumes
\cite{gehrig2019evflownetest,zhu2019evflownet,gehrig2021eraft,paredes2021back,wan2023rpeflow,gehrig2024dense},
and per-pixel temporal Gaussian fits \cite{ding2022steflownet}. Supervised
learning techniques (\eg,
\cite{gehrig2019evflownetest,gehrig2021eraft,wan2023rpeflow,gehrig2024dense})
train their networks via a photometric loss computed from the predicted and
ground-truth flows. As such, their performance may vary across lightning
conditions and scene dynamics, especially when the ground truth is not accurate
or when testing data comes from a different distribution.

Self-supervised methods \cite{ding2022steflownet, zhu2018evflownet} warp
corresponding grayscale images using the predicted flow and compute their
photometric consistency loss as a supervisory signal. Consequently, these
methods can suffer from poor image quality (\eg, due to motion blur and low
dynamic range) and inaccurate registration between frames from disparate
cameras (with different resolutions, field of views, etc.). Unsupervised
learning techniques
\cite{zhu2019evflownet,hagenaars2021convgruevflownet,paredes2021back}, similar
to model-based CM, train their network with a CM objective that measures the
event alignment error. These approaches show limited generalization on unseen
data, and theoretically cannot surpass the performance of model-based CM.

% --------------------------------------------------------------------------------
\subsection{Event Datasets for Optical Flow}
\label{subsec:event_datasets_for_optical_flow}
Currently, there are few real-world event datasets with high-quality optical
flow ground truth (\eg, MVSEC \cite{zhu2018evflownet} and DSEC
\cite{gehrig2021eraft}). The MVSEC dataset contains stereo sequences of indoor
and outdoor environments, captured by a pair of DAVIS346B cameras, each of
which records frames and events on the same $346\times 260$ pixel array.
However, the optical flow ground truth lacks accuracy due to synchronization
deficiencies. Furthermore, the dataset only has very small displacements (\eg,
less than 10 pixels). The DSEC dataset provides outdoor stereo driving
sequences that exhibit larger displacements (up to 210 pixels) with more
accurate sparse optical flow ground truth and a higher resolution camera setup.
However, since events and RGB frames were recorded on different cameras, it is
difficult to obtain perfect event-frame alignment. Simulation suites are also
proposed for event vision research (\eg, ESIM \cite{rebecq2018esim},
DVS-Voltmeter \cite{lin2022dvs}, and V2E \cite{hu2021v2e}). These simulations
have different event generation models with distinct trade-offs (\eg, speed
versus realism).

% --------------------------------------------------------------------------------
\subsection{Contrast Maximization and Hybrid Frameworks}
\label{subsec:contrast_maximization_and_hybrid_frameworks}
Previous CM works (\eg,
\cite{gallego2019focus,stoffregen2019analysis,shiba2022secrets}) have
demonstrated that with a well-regularized \emph{contrast} objective function,
we can achieve good predicted flow and avoid the \emph{event collapse} problem
mentioned in \cref{subsec:model-based-approaches}. For example, MultiCM
\cite{shiba2022secrets} employed multiscale and multireference techniques to
obtain state-of-the-art performance in model-based estimation of event optical
flow. On the other hand, Wang \etal \cite{wang2020jointfiltering} developed a
bi-modal technique in which a CM objective was optimized with respect to a
weighted sum of the IWE and Sobel edge image obtained from the corresponding
grayscale image. In contrast, we provide further assistance to the CM objective
function by simultaneously maximizing the \emph{contrast} of the IWE and its
spatial correlation with the corresponding edge image. Furthermore, our
approach refines existing pyramidal and multireference techniques in the CM
framework while extending them from a uni-modal to a bi-modal setting. 

% ------------------------------------------------------------------------------
% METHOD
% ------------------------------------------------------------------------------
\section{Method}
\label{sec:method}

% ------------------------------------------------------------------------------
\subsection{Events and Contrast Maximization}
\label{subsec:events_and_contrast_maximization}
%%%%%%%%%%%%%%%%%%%%%%%%%%%%%%%%%%%%%%%%%%%%%%%%%%%%%%%%%%%%%%%%%%%%%%%%%%%%%%%%
% Figure: Edge Extraction Pipeline
\begin{figure*}%[!ht]
    \centering
    \begin{adjustbox}{max width=0.9\textwidth}
    \begin{tabular}{@{}c@{\thinspace}c@{\thinspace}c@{\thinspace}c@{\thinspace}c@{\thinspace}c@{\thinspace}c@{}}
    \begin{tabular}{@{}c@{}} \begin{adjustbox}{max width=0.142\textwidth} Original \end{adjustbox}\end{tabular} & 
    \begin{tabular}{@{}c@{}} \begin{adjustbox}{max width=0.142\textwidth} Denoise  \end{adjustbox}\end{tabular} &
    \begin{tabular}{@{}c@{}} \begin{adjustbox}{max width=0.142\textwidth} CLAHE    \end{adjustbox}\end{tabular} &
    \begin{tabular}{@{}c@{}} \begin{adjustbox}{max width=0.142\textwidth} Sharpen  \end{adjustbox}\end{tabular} &
    \begin{tabular}{@{}c@{}} \begin{adjustbox}{max width=0.142\textwidth} Filter   \end{adjustbox}\end{tabular} &
    \begin{tabular}{@{}c@{}} \begin{adjustbox}{max width=0.142\textwidth} Canny    \end{adjustbox}\end{tabular} &
    \begin{tabular}{@{}c@{}} \begin{adjustbox}{max width=0.142\textwidth} Blur     \end{adjustbox}\end{tabular} \\

    \begin{tabular}{@{}c@{}} \includegraphics[width=0.142\textwidth, cfbox=gray 0.1pt 0pt]{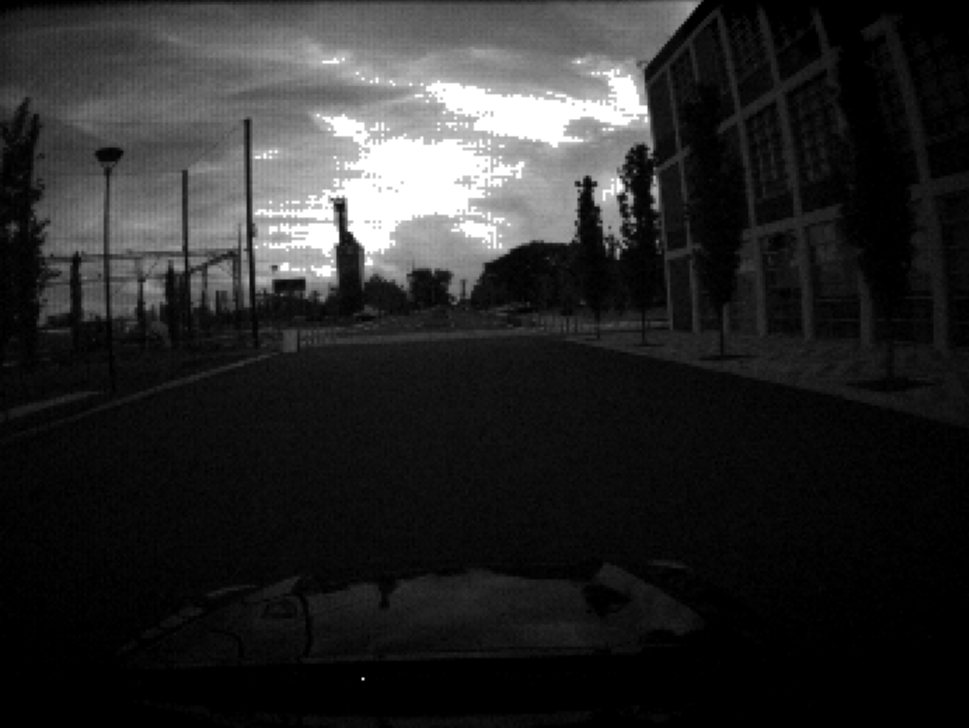} \end{tabular} &
    \begin{tabular}{@{}c@{}} \includegraphics[width=0.142\textwidth, cfbox=gray 0.1pt 0pt]{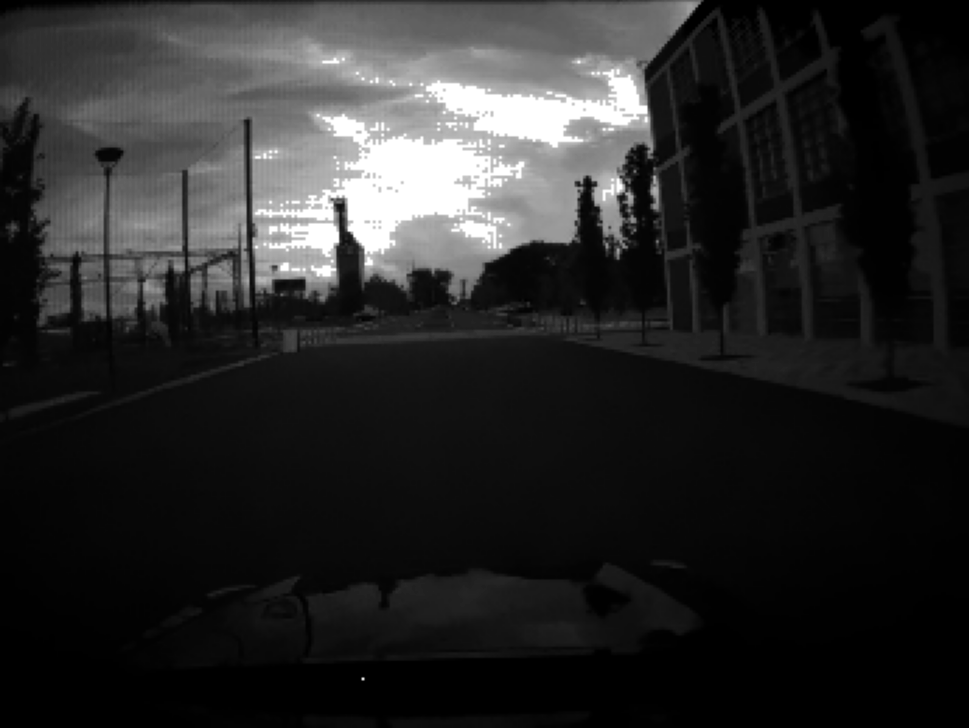} \end{tabular} &
    \begin{tabular}{@{}c@{}} \includegraphics[width=0.142\textwidth, cfbox=gray 0.1pt 0pt]{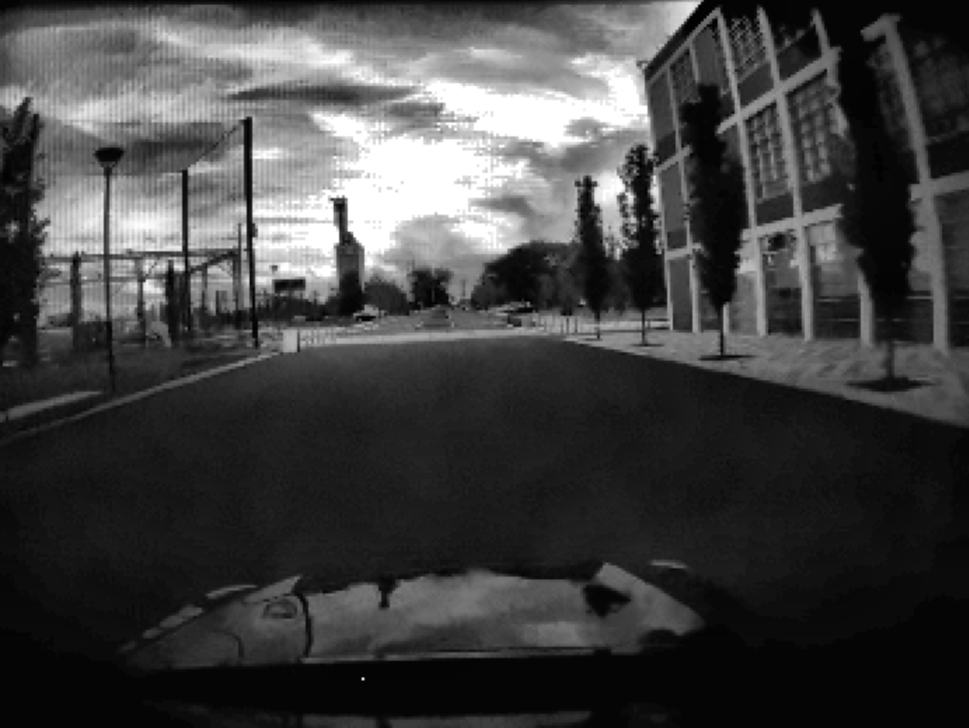} \end{tabular} &
    \begin{tabular}{@{}c@{}} \includegraphics[width=0.142\textwidth, cfbox=gray 0.1pt 0pt]{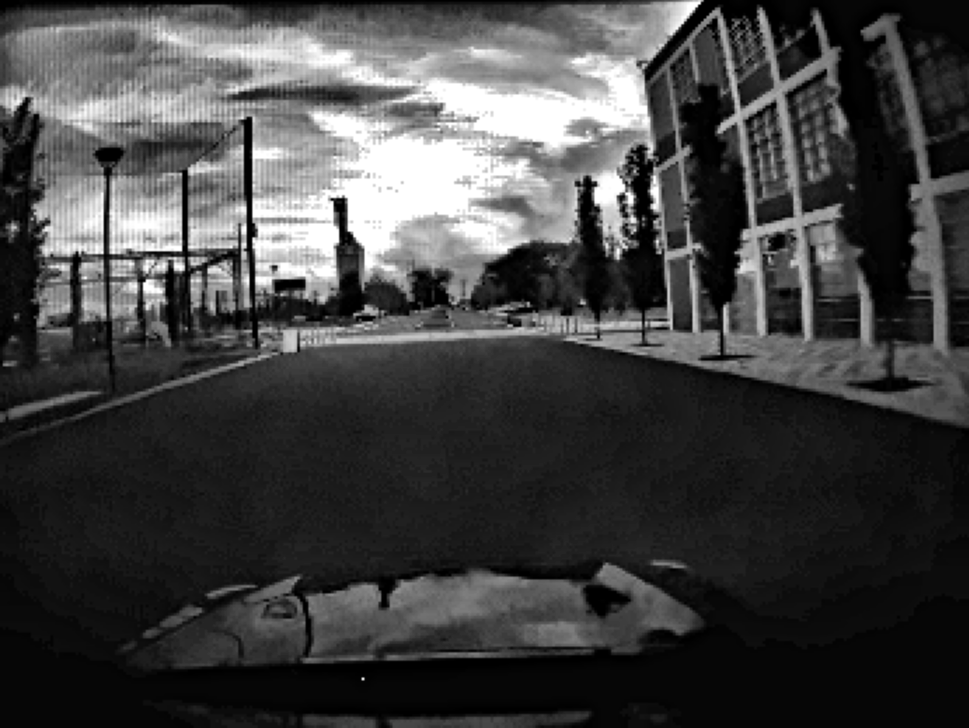} \end{tabular} &
    \begin{tabular}{@{}c@{}} \includegraphics[width=0.142\textwidth, cfbox=gray 0.1pt 0pt]{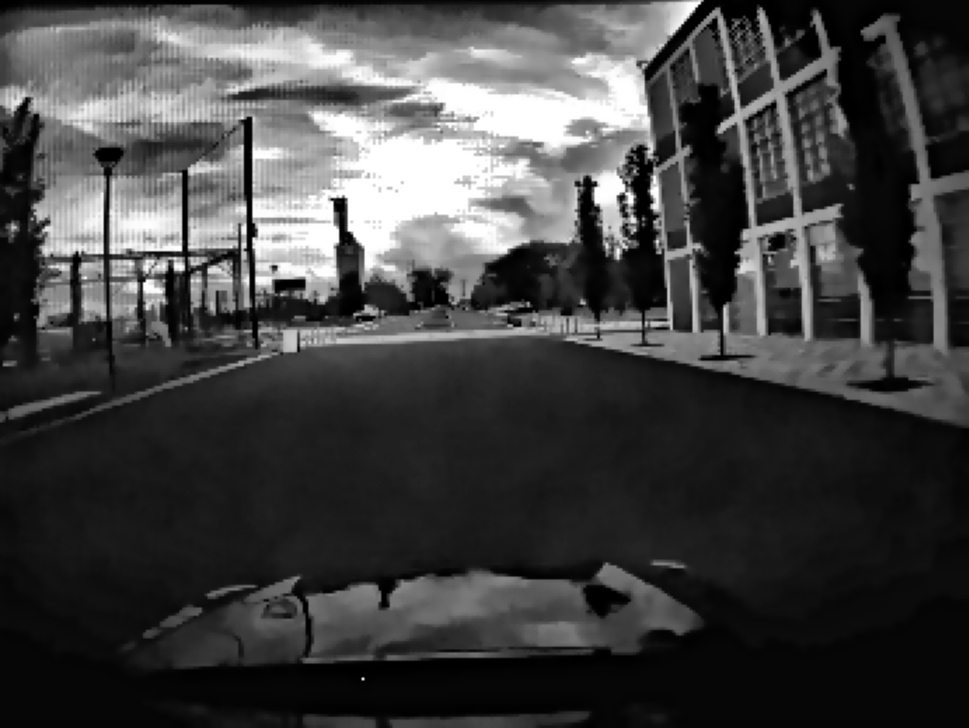} \end{tabular} &
    \begin{tabular}{@{}c@{}} \includegraphics[width=0.142\textwidth, cfbox=gray 0.1pt 0pt]{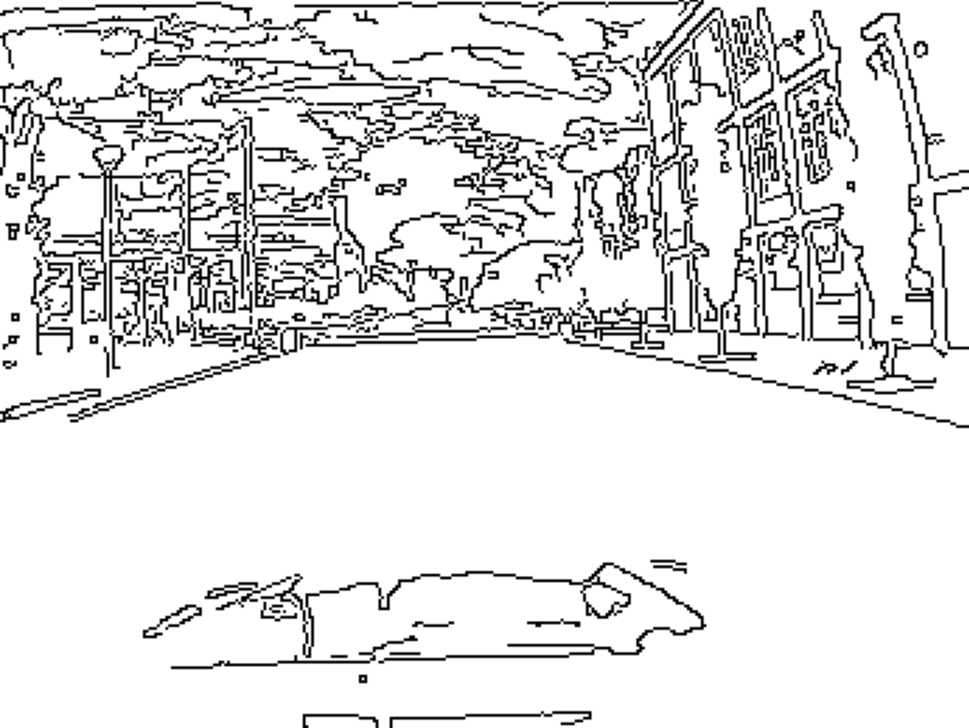} \end{tabular} &
    \begin{tabular}{@{}c@{}} \includegraphics[width=0.142\textwidth, cfbox=gray 0.1pt 0pt]{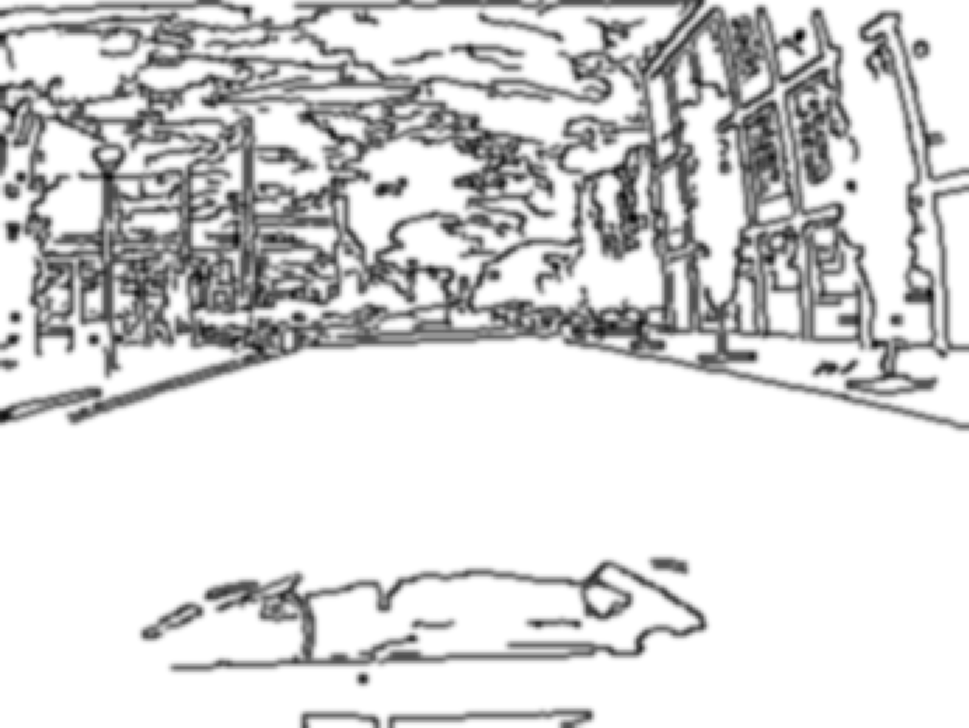} \end{tabular} \\
    
    \end{tabular}
    \end{adjustbox}
    \caption{The edge extraction pipeline. To extract a viable edge image from a grayscale image, we sequentially apply non-local means denoising, contrast-limited adaptive histogram equalization (CLAHE), Gaussian sharpening, bilateral filtering, Canny edge detection, and finally Gaussian blur for edge smoothing (please see supplementary material for an edge smoothing sensitivity analysis). Our pipeline was tested extensively for images in the MVSEC (low-quality) and DSEC (high-quality) datasets, and can be adjusted for other use cases.}
    \label{fig:edge_extraction_pipeline}
\end{figure*}
%%%%%%%%%%%%%%%%%%%%%%%%%%%%%%%%%%%%%%%%%%%%%%%%%%%%%%%%%%%%%%%%%%%%%%%%%%%%%%%%

In a CM setting, we process all the data from an event stream in parts based on
time intervals. Specifically, we work on a single ($i$-th) event set
$\mathcal{E}^{(i)} \doteq \{e^{(i)}_k\}^{N^{(i)}_{e}}_{k=1}$ at a time,
consisting of $N^{(i)}_{e}$ individual events. For brevity, we omit the
superscript ($\cdot^{(i)}$) from the remainder of this paper. Each event $e_k
\doteq (\mathbf{x}_k, t_k, p_k)$ is recorded as a 4-tuple comprising the
$x\mhyphen y$ coordinates of the location of the triggered pixel $\mathbf{x}_k
= (x_k, y_k)$ as well as the timestamp $t_k$ and the polarity $p_k$ of the
event. 

Let $\mathbf{X}_{\origevents} = \{\mathbf{x}_k\}^{N_e}_{k=1}$ and
$\mathbf{X}_{\warpedevents} = \{\mathbf{x}_k'\}^{N_e}_{k=1}$ denote the set of
event coordinates that belong to the original event set $\origevents$ and the
warped event set $\warpedevents$ at the reference time $\tref$, respectively.
Within a small enough time window, we assume that all local motions are linear.
A warping function $\mathbf{W}(\mathbf{x}_{k}, t_k; \boldsymbol{\theta}_k,
\tref)$ defines a mapping $\origevents \mapsto \warpedevents$ that transports
$\mathbf{x}_k \in \mathbf{X}_{\origevents}$ to $\mathbf{x}_k' \in
\mathbf{X}_{\warpedevents}$ at the reference time $\tref$ according to 
\begin{equation}
\begin{aligned}
  \mathbf{x}_k' &\doteq \mathbf{W}(\mathbf{x}_{k}, t_k; \boldsymbol{\theta}_k, \tref) \\
                &= \mathbf{x}_k + \boldsymbol{\theta}_k(\tref - t_k), \quad 1 \leq k \leq N_e,
  \label{eq:warp_function}
\end{aligned}
\end{equation}
along motion trajectories modeled by motion parameters $\boldsymbol{\Theta} =
\{\boldsymbol{\theta}_k\}^{N_e}_{k=1}$, where $\boldsymbol{\theta}_k =
\mathbf{v}(\mathbf{x}_k)$ is the velocity vector at $\mathbf{x}_k$. Next, these
warped events $\warpedevents$ are utilized to construct the IWE by organizing
them by their coordinates $\mathbf{x}_k' \in \mathbf{X}_{\warpedevents}$ and
aggregating them on the image plane as
\begin{equation}
  I_{\text{events}}(\mathbf{x}; \boldsymbol{\Theta}, \tref) \doteq \sum^{N_e}_{k=1} \delta(\mathbf{x} - \mathbf{x}_k'), 
  \label{eq:iwe}
\end{equation}
where $\delta(\cdot)$ is the Dirac delta function. In practice, for CM the
function $\delta$ is replaced by a smooth approximation $\delta_{\sigma}$ such
as a Gaussian, \ie, $\delta_{\sigma}(\mathbf{x} - \boldsymbol{\mu}) \doteq
\left. \mathcal{N}(\mathbf{x}; \boldsymbol{\mu}, \sigma^2 \texttt{Id})
\right\vert_{\sigma=1 \text{\,pixel}}$. Note that we opt to ignore event
polarities when constructing the IWE, but one could choose to construct a
two-channel IWE or a summation of the two polarities. Next, the \emph{contrast}
of this IWE is defined using a CM objective function $f(\boldsymbol{\Theta})$
such as the variance function 
\begin{equation}
  \begin{aligned}
  & \text{Var}(I_{\text{events}}(\boldsymbol{\Theta}; \tref)) \\
  & \doteq \frac{1}{\vert \Omega \vert} \int_{\Omega} \left( I_{\text{events}}(\mathbf{x}; \boldsymbol{\Theta}, \tref) - \mu_{I_{\text{events}}(\boldsymbol{\Theta}; \tref)} \right)^2 d\mathbf{x},
  \end{aligned}
  \label{eq:variance}
\end{equation}
with mean $\mu_{I_{\text{events}}(\boldsymbol{\Theta}; \tref)} \doteq
\frac{1}{\vert \Omega \vert} \int_{\Omega} I_{\text{events}}(\mathbf{x};
\boldsymbol{\Theta}, \tref) d\mathbf{x}$, where $\Omega$ denotes the sensor
coordinate subspace. The CM objective function is intended to implicitly
measure how well the motion parameters $\boldsymbol{\Theta}$ model the original
motion of the intensity gradients that generated the original events. A higher
measure of \emph{contrast} is associated with a better alignment of warped
events as well as more accurate motion parameters. Finally, CM is the task of
maximizing this \emph{contrast} with respect to $\boldsymbol{\Theta}$ to find
the optimal motion parameters $\boldsymbol{\Theta}^{\ast} =
\argmax_{\boldsymbol{\Theta}} f(\boldsymbol{\Theta})$.

Inspired by Gallego \etal \cite{gallego2019focus}, we use the mean squared
magnitude of the IWE gradient,
\begin{equation}
  G(\boldsymbol{\Theta}; \tref)) \doteq \frac{1}{\vert \Omega \vert} \int_{\Omega} \Vert \nabla I_{\text{events}}(\mathbf{x}; \boldsymbol{\Theta}, \tref) \Vert^2 d\mathbf{x},
  \label{eq:gradient_magnitude}
\end{equation}
as our CM objective function. We chose $G(\cdot)$ because (i) in contrast to
the zeroth-order with $\text{Var}(\cdot)$, it encodes a first-order spatial
constraint by its definition, and (ii) it empirically provides a faster and
better convergence in our evaluation. Finally, we calculate \emph{relative
contrast} to be used as our \emph{contrast} objective,
\begin{equation}
  f_{\text{rel}}(\boldsymbol{\Theta}) = \frac{G(\boldsymbol{\Theta}; \tref)}{G(\mathbf{0}_{\boldsymbol{\Theta}};-)}.
  \label{eq:rel_contrast}
\end{equation}

% ------------------------------------------------------------------------------
\subsection{Edge-Informed Contrast Maximization}
\label{subsec:edge-informed_contrast_maximization}
We hypothesize that the underlying objective of estimating the true parameters
that model the original motion is only partially informed by \emph{contrast}.
Based on this hypothesis, we extract edges from image frames
(\cref{fig:edge_extraction_pipeline}) to introduce an additional
correlation-based objective $g(\boldsymbol{\Theta})$ that constrains the space
of admissible solutions. In doing so, we extend CM from the existing uni-modal
structure to a bi-modal edge-informed CM (EINCM) framework. As stated in
\cref{sec:intro}, we incorporate edge consistency by extracting edges
$E(\mathbf{x}, \tref)$ from a frame-based camera and enforce \emph{correlation}
between the edges and the IWE. We use \emph{mean squared error}
(MSE){\footnote{Both the IWE and the edge image are normalized before measuring
the error.}} to measure \emph{correlation} at the reference time $\tref$
according to
\begin{equation}
\begin{aligned}
  & \text{MSE}(\boldsymbol{\Theta}; \tref)) \\
  & = \frac{1}{\vert \Omega \vert} \int_{\Omega} \left( I_{\text{events}}(\mathbf{x}; \boldsymbol{\Theta}, \tref) - E(\mathbf{x}, \tref) \right)^{2} d\mathbf{x}.
\end{aligned}
  \label{eq:mse}
\end{equation}

The additive inverse of the MSE was chosen as our correlation objective
function over other functions (\eg, sum of squared errors, mean absolute
differences, sum of absolute differences, mean of the Hadamard product, or the
sum of the Hadamard product) since it provided faster and better convergence
during the optimization process. Finally, we calculate the \emph{relative
correlation},
\begin{equation}
  g_{\text{rel}}(\boldsymbol{\Theta}) = -\frac{\text{MSE}(\boldsymbol{\Theta}; \tref)}{\text{MSE}(\mathbf{0}_{\boldsymbol{\Theta}}; -)},
  \label{eq:rel_correlation}
\end{equation}
to be used as our \emph{correlation} objective.

% ------------------------------------------------------------------------------
\subsubsection{Multiple Reference Times for Events and Edges}
\label{subsubsec:multiple_reference_times_for_events_and_edges}
An event set $\mathcal{E}$ spans a temporal window $[t_0, t_1]$, where $t_0 =
t_{\text{min}}$ and $t_1 = t_{\text{max}}$. Typically, events are warped to the
reference time $t_0$ during optimization to measure \emph{contrast}.
Multireference in model-based CM was first introduced in MultiCM
\cite{shiba2022secrets}, where this \emph{contrast} is computed at three
different reference times $t_0$, $t_{\text{mid}}$, and $t_1$ to discourage
overfitting of the motion parameters or flow field. Similar to MultiCM, we
compute \emph{contrast} at multiple reference times. Furthermore, we also
compute correlations at multiple reference times, identical to the exact
timestamps of the $N_{\text{img}}$ frames at
$\{t_{j}\}^{N_{\text{img}}}_{j=1}$. Specifically, both \emph{relative contrast}
and \emph{relative correlation} are computed at multiple reference times
according to
\begin{align}
  f_{\text{rel}}(\boldsymbol{\Theta}) & = \frac{1}{N_{\text{img}} \cdot G(\mathbf{0}_{\boldsymbol{\Theta}}; -)} \sum^{N_{\text{img}}}_{j=1} G(\boldsymbol{\Theta}; t_j), \\
  g_{\text{rel}}(\boldsymbol{\Theta}) & = -\frac{1}{N_{\text{img}}} \sum^{N_{\text{img}}}_{j=1} \frac{\text{MSE}(\boldsymbol{\Theta}; \tref)}{\text{MSE}(\mathbf{0}_{\boldsymbol{\Theta}}; -)}.
  \label{eq:multi_reference}
\end{align}
Our hybrid CM objective, $\mathcal{F}(\boldsymbol{\Theta})$, can be summarized
as a function that maximizes IWE \emph{contrast} and edge \emph{correlation}
simultaneously,
\begin{equation}
    \boldsymbol{\Theta}^{\ast} = \argmax_{\boldsymbol{\Theta}} ( \underbrace{\alpha f_{\text{rel}}(\boldsymbol{\Theta}) + \beta g_{\text{rel}}(\boldsymbol{\Theta})}_{\mathcal{F}(\boldsymbol{\Theta})} + \gamma \mathcal{R}(\boldsymbol{\Theta}) ),
\end{equation}
where $\alpha$, $\beta$, and $\gamma$ are balancing coefficients.
$\mathcal{R}(\boldsymbol{\Theta})$ is a regularization term to discount
non-smooth solutions.  Following previous work \cite{shiba2022secrets}, we use
the additive inverse of the total variation of $\boldsymbol{\Theta}$.
Furthermore, when edge images cannot be reliably obtained, $\beta$ can be set
to zero, which enables our framework use only events similar to
\cite{gallego2018unifying,shiba2022secrets}.

% ------------------------------------------------------------------------------
\subsubsection{Multiple Scales and Handovers}
\label{subsubsec:multiple_scales_and_handovers}
%%%%%%%%%%%%%%%%%%%%%%%%%%%%%%%%%%%%%%%%%%%%%%%%%%%%%%%%%%%%%%%%%%%%%%%%%%%%%%%%
% Figure: Multiscale Strategy
\begin{figure}
\centering
\includegraphics[width=\columnwidth]{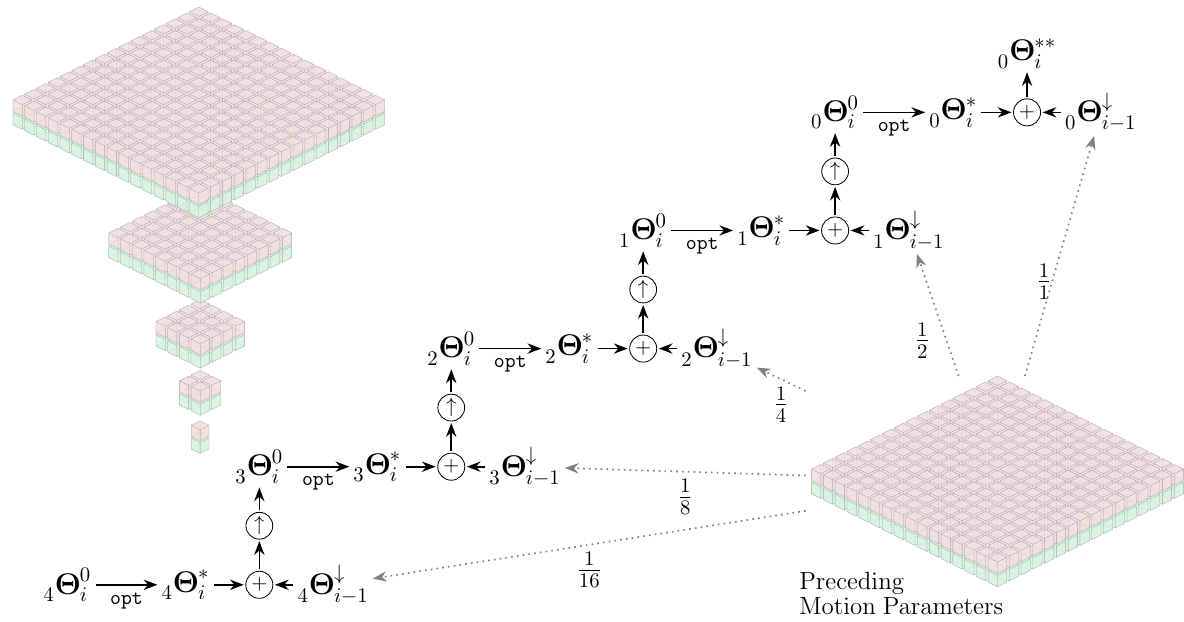}
\caption{Our \emph{pre-handover} multiscale strategy with an
optimize-handover-upsample pipeline. Notation:
${}_{p}\boldsymbol{\Theta}^{q}_{r}$, the number $p \in \{0, 1, 2, 3, 4\}$
denotes the pyramid level, $q \in \{0, \ast, \downarrow\}$ indicates different
versions of the motion parameters, where $q=0$ and $q=\ast$ represents pre- and
post-optimization, respectively. $q=\,\downarrow$ indicates downsampled from
the preceding iteration. $r \in \{i, i-1\}$ represents the iteration. The
symbols \protect\tikz[baseline] \protect\node[circle, draw=black, inner
sep=0.1ex, outer sep=0.1ex] at (0ex,0.5ex) {\tiny$+$}; and
\protect\tikz[baseline] \protect\node[circle, draw=black, inner sep=0.1ex,
outer sep=0.1ex] at (0ex,0.5ex) {\tiny$\uparrow$}; denote handover and
upsampling operations, respectively.}
\label{fig:multi_scale}
\end{figure}

%%%%%%%%%%%%%%%%%%%%%%%%%%%%%%%%%%%%%%%%%%%%%%%%%%%%%%%%%%%%%%%%%%%%%%%%%%%%%%%%

Ever since the inception of multiresolution image processing, it has been used
to enable faster and better convergence, allowing for smoother and more robust
solutions. Today, it is used in several model-based and learning-based optical
flow estimation approaches
\cite{akolkar2020real,ding2022steflownet,shiba2022secrets}. Although there is
no best-known strategy to apply multiscale to the CM problem, in this work we
outline our multiscale strategy as depicted in detail in
\cref{fig:multi_scale}.

Upsampling is necessary during multiscaling; however, it may not be sufficient.
As such, an aggregation operation is also used to incorporate solutions from
preceding iterations. We call this a \emph{handover}. Typically, at a given
coarse pyramid level, a \emph{handover} is performed between the optimized
$\boldsymbol{\Theta}^{\ast}_{i}$ and the downscaled
$\boldsymbol{\Theta}^{\downarrow}_{i-1}$ from the preceding iteration. In this
work, we used the \texttt{lanczos3} algorithm to downsample and
\texttt{repeat}\footnote{Both \texttt{lanczos3} and \texttt{repeat} are
available in JAX and Numpy.} for intra-scale upsampling, as it enabled better
convergence in our method. In our multiscale strategy, the processing pipeline
consists of optimize-handover-upsample cycles. We refer to this as a
\emph{pre-handover} scheme. This is in stark contrast to the
optimize-upsample-handover cycle (post-handover scheme) used in MultiCM. We
found that replacing a \emph{post-handover} with a \emph{pre-handover} scheme
enabled superior overall performance for our method. As depicted in
\cref{tab:mvsec_accuracies}, our \textit{events only} baseline beats the
previous \textit{events only} state of the art even without utilizing edges.

MultiCM performs an averaging operation during handovers. We add flexibility to
this design as follows. We consider two sub-strategies where the
\emph{handover} operation can be performed using (i) fixed weights or (ii)
solved weights. In the fixed-weight case, we use a fixed-handover weight
$w_{\text{ho}}$, while in the solved-weight case we solve for
$w_{\text{ho}}^{\ast} = \argmax_{w_{\text{ho}}} \mathcal{F} \left(
w_{\text{ho}} \cdot \boldsymbol{\Theta}_{i-1} + (1-w_{\text{ho}}) \cdot
\boldsymbol{\Theta}_{i} \right)$ to optimize the overall CM objective. Lastly,
we use a fixed $w_{\text{ho}}$ for coarser pyramid levels $\{4,3,2\}$ and solve
$w_{\text{ho}}^{\ast}$ on finer levels $\{1,0\}$.

% ------------------------------------------------------------------------------
% EXPERIMENTS
% ------------------------------------------------------------------------------
\section{Experiments}
\label{sec:experiments}
\subsection{Experimental Setup}
\label{subsec:experimental_setup}
All experimental runs were executed using Python 3.11 on a workstation with an
Intel Core i7-8700 CPU, 32 GB RAM, and NVIDIA Quadro P4000 8GB GPU with Ubuntu
22.04. Image processing was performed using OpenCV 4.8. For the optimization
tasks, we employed JAX 0.4 \cite{jax2018github,frostig2018highleveltracing}
with its support library \texttt{jaxlib} 0.4.19+cuda12.cudnn89 and JAXopt 0.8
\cite{jaxopt_implicit_diff} with CUDA toolkit 12.3. We used a quasi-Newton full
batch gradient-based optimization algorithm `BFGS'
\cite{shanno1970conditioning,shanno1985example}, which is available in JAXopt
through its SciPy wrapper.

% --------------------------------------------------------------------------------
\subsection{Datasets, Metrics, and Hyperparameters}
\label{subsec:datasets_metrics_and_hyperparameters}
We evaluated our method on the following publicly available datasets for
optical flow benchmarking: MVSEC \cite{zhu2018multivehicle}, DSEC
\cite{gehrig2021dsec,gehrig2021eraft}, and ECD \cite{mueggler2017event}. The
MVSEC dataset consists of indoor scenes on a flying platform and outdoor scenes
on a mobile vehicle. MVSEC was expanded by \cite{zhu2018evflownet} to include
ground-truth optical flow calculated as a motion field
\cite{trucco1998introductory,rueckauer2016evaluation} derived from camera
velocity and scene depth. Both image and event data were captured through the
same sensor array of the DAVIS 346B ($346\times260\,$pix\textsuperscript{2}).
The DSEC dataset captures driving scenes under different lighting conditions.
It was updated in \cite{gehrig2021eraft} to provide ground-truth optical flow
for a limited number of sequences and a subset of the overall time intervals
within them. DSEC was collected using separate sensors to capture images (FLIR
Blackfly S USB3 $1440\times1080\,$pix\textsuperscript{2}) and events (Prophesee
Gen 3.1 $640\times480\,$pix\textsuperscript{2}) in a stereo configuration. For
benchmarking, the optical ground truth in DSEC evaluates the estimated optical
flow in the left event camera rectified frame. The ECD dataset contains a
variety of scenes and provides image and event data
($240\times180\,$pix\textsuperscript{2}) without ground-truth optical flow.

Following previous work
\cite{zhu2018evflownet,gehrig2021eraft,shiba2022secrets}, we used the average
endpoint error (AEE), the average $n$-pixel error percentage (outlier
percentage), and the flow warp loss (FWL) for quantitative comparisons. In all
experiments, we utilized five pyramid levels for the motion parameters, where
the resolutions range from $1\times1$ to $16\times16$. Bilinear upsampling was
used to upscale the finest scale ($16\times16$) ${}_0\boldsymbol{\Theta}$ to
the sensor size. During optimization on MVSEC, we employed 30 K and 40 K events
for indoor and outdoor scenes, respectively. For DSEC and ECD, we used 1.5 M
and 30 K events, respectively. For more details, please see the supplementary
material.

%%%%%%%%%%%%%%%%%%%%%%%%%%%%%%%%%%%%%%%%%%%%%%%%%%%%%%%%%%%%%%%%%%%%%%%%%%%%%%%%
% Table: MVSEC Accuracies
\begin{table}
    \centering
    \begin{adjustbox}{max width=\columnwidth}
        \begin{tabular}{@{\thinspace}c@{\medspace}l@{\thickspace}
        c@{\thickspace}c@{\thickspace}c 
        c@{\thickspace}c@{\thickspace}c 
        c@{\thickspace}c@{\thickspace}c 
        c@{\thickspace}c} 
    
            % ================================================================================
            \toprule
            & &  
            \multicolumn{2}{c}{\texttt{indoor\_flying1}}        & \thickspace &     % indoor_flying1
            \multicolumn{2}{c}{\texttt{indoor\_flying2}}        & \thickspace &     % indoor_flying2
            \multicolumn{2}{c}{\texttt{indoor\_flying3}}        & \thickspace &     % indoor_flying3
            \multicolumn{2}{c}{\texttt{outdoor\_day1}}          \\                  % outdoor_day1    
            % --------------------------------------------------------------------------------
            \cmidrule(){3-4} \cmidrule(){6-7} \cmidrule(){9-10} \cmidrule(){12-13} 
            % 
            % dt = 1
            % ------
            & \multicolumn{1}{c}{\emph{dt} = 1}  & 
            AEE $\downarrow$         & \%Out $\downarrow$       & \thickspace &     % indoor_flying1 
            AEE $\downarrow$         & \%Out $\downarrow$       & \thickspace &     % indoor_flying2 
            AEE $\downarrow$         & \%Out $\downarrow$       & \thickspace &     % indoor_flying3 
            AEE $\downarrow$         & \%Out $\downarrow$       \\                  % outdoor_day1      
            % --------------------------------------------------------------------------------
            \midrule 

            % SL
            \multirow{3}{*}{\rotatebox[origin=c]{90}{SL}} 
            & EV-FlowNet-EST \cite{gehrig2019evflownetest} & 
            0.97                     & 0.91                     & \thickspace &     % indoor_flying1
            1.38                     & 8.20                     & \thickspace &     % indoor_flying2
            1.43                     & 6.47                     & \thickspace &     % indoor_flying3
            --                       & --                       \\                  % outdoor_day1
            & EV-FlowNet$+$ \cite{stoffregen2020evflownetplus} & 
            0.56                     & 1.00                     & \thickspace &     % indoor_flying1
            0.66                     & 1.00                     & \thickspace &     % indoor_flying2
            0.59                     & 1.00                     & \thickspace &     % indoor_flying3
            0.68                     & 0.99                     \\                  % outdoor_day1
            & E-RAFT \cite{gehrig2021eraft} & 
            --                       & --                       & \thickspace &     % indoor_flying1
            --                       & --                       & \thickspace &     % indoor_flying2
            --                       & --                       & \thickspace &     % indoor_flying3
            \textbf{0.24}            & 1.70                     \\                  % outdoor_day1
            % ================================================================================
            \midrule

            % SSL
            \multirow{3}{*}{\rotatebox[origin=c]{90}{SSL}} 
            & EV-FlowNet (original) \cite{zhu2018evflownet} & 
            1.03                     & 2.20                     & \thickspace &     % indoor_flying1
            1.72                     & 15.1                     & \thickspace &     % indoor_flying2
            1.53                     & 11.90                    & \thickspace &     % indoor_flying3
            0.49                     & 0.20                     \\                  % outdoor_day1
            & Spike-FlowNet \cite{lee2020spikeflownet} & 
            0.84                     & --                       & \thickspace &     % indoor_flying1
            1.28                     & --                       & \thickspace &     % indoor_flying2
            1.11                     & --                       & \thickspace &     % indoor_flying3
            0.49                     & --                       \\                  % outdoor_day1
            & STE-FlowNet \cite{ding2022steflownet} & 
            0.57                     & 0.10                     & \thickspace &     % indoor_flying1
            0.79                     & 1.60                     & \thickspace &     % indoor_flying2
            0.72                     & 1.30                     & \thickspace &     % indoor_flying3
            0.42                     & \textbf{0.00}            \\                  % outdoor_day1
            % ================================================================================
            \midrule

            % USL
            \multirow{3}{*}{\rotatebox[origin=c]{90}{USL}} 
            & EV-FlowNet \cite{zhu2019evflownet} & 
            0.58                     & \textbf{0.00}            & \thickspace &     % indoor_flying1
            1.02                     & 4.00                     & \thickspace &     % indoor_flying2
            0.87                     & 3.00                     & \thickspace &     % indoor_flying3
            \underline{0.32}         & \textbf{0.00}            \\                  % outdoor_day1
            & FireFlowNet \cite{paredes2021back} & 
            0.97                     & 2.60                     & \thickspace &     % indoor_flying1
            1.67                     & 15.30                    & \thickspace &     % indoor_flying2
            1.43                     & 11.00                    & \thickspace &     % indoor_flying3
            1.06                     & 6.60                     \\                  % outdoor_day1
            & ConvGRU-EV-FlowNet \cite{hagenaars2021convgruevflownet} & 
            0.60                     & 0.51                     & \thickspace &     % indoor_flying1
            1.17                     & 8.06                     & \thickspace &     % indoor_flying2
            0.93                     & 5.64                     & \thickspace &     % indoor_flying3
            0.47                     & 0.25                     \\                  % outdoor_day1
            % ================================================================================
            \midrule

            % MB
            \multirow{6}{*}{\rotatebox[origin=c]{90}{MB}} 
            & Nagata \etal \cite{nagata2021matchtimesurf} & 
            0.62                     & --                       & \thickspace &     % indoor_flying1
            0.93                     & --                       & \thickspace &     % indoor_flying2
            0.84                     & --                       & \thickspace &     % indoor_flying3
            0.77                     & --                       \\                  % outdoor_day1
            & Akolkar \etal \cite{akolkar2020real} & 
            1.52                     & --                       & \thickspace &     % indoor_flying1
            1.59                     & --                       & \thickspace &     % indoor_flying2
            1.89                     & --                       & \thickspace &     % indoor_flying3
            2.75                     & --                       \\                  % outdoor_day1
            & Brebion \etal \cite{brebion2021realtimeflow} & 
            0.52                     & 0.10                     & \thickspace &     % indoor_flying1
            0.98                     & 5.50                     & \thickspace &     % indoor_flying2
            0.71                     & 2.10                     & \thickspace &     % indoor_flying3
            0.53                     & \underline{0.20}         \\                  % outdoor_day1
            & Shiba \etal \cite{shiba2022secrets} & 
            0.42                     & 0.09                     & \thickspace &     % indoor_flying1
            0.60                     & 0.59                     & \thickspace &     % indoor_flying2
            0.50                     & 0.29                     & \thickspace &     % indoor_flying3
            0.72\tiny{5}             & 5.49\tiny{4}             \\                  % outdoor_day1
            & Ours (CM events only) & 
            \underline{0.37\tiny{8}} & 0.05\tiny{5}             & \thickspace &     % indoor_flying1
            \underline{0.50\tiny{9}} & \underline{0.17\tiny{2}} & \thickspace &     % indoor_flying2
            \underline{0.44\tiny{3}} & \underline{0.05\tiny{7}} & \thickspace &     % indoor_flying3
            {0.64\tiny{9}}           & {4.22\tiny{6}}           \\                  % outdoor_day1
            & Ours (EINCM) & 
            \textbf{0.37\tiny{2}}    & \underline{0.04\tiny{6}} & \thickspace &     % indoor_flying1
            \textbf{0.50\tiny{5}}    & \textbf{0.15\tiny{5}}    & \thickspace &     % indoor_flying2
            \textbf{0.43\tiny{7}}    & \textbf{0.02\tiny{9}}    & \thickspace &     % indoor_flying3
            {0.61\tiny{8}}           & {3.69\tiny{3}}           \\                  % outdoor_day1
            % ================================================================================
            \bottomrule
            %
            % dt = 4
            % ------
            &                                   & & & \thickspace & & & \thickspace & & & \thickspace & & \\
            & \multicolumn{1}{c}{\emph{dt} = 4} & & & \thickspace & & & \thickspace & & & \thickspace & & \\
            % --------------------------------------------------------------------------------
            \midrule
            
            \multirow{3}{*}{\rotatebox[origin=c]{90}{SSL}}
            & EV-FlowNet (original) \cite{zhu2018evflownet} & 
            2.25                     & 24.70                     & \thickspace &    % indoor_flying1
            4.05                     & 45.30                     & \thickspace &    % indoor_flying2
            3.45                     & 39.70                     & \thickspace &    % indoor_flying3
            \textbf{1.23}            & \textbf{7.30}             \\                 % outdoor_day1
            & Spike-FlowNet \cite{lee2020spikeflownet} & 
            2.24                     & --                        & \thickspace &    % indoor_flying1
            3.83                     & --                        & \thickspace &    % indoor_flying2
            3.18                     & --                        & \thickspace &    % indoor_flying3
            1.09                     & --                        \\                 % outdoor_day1
            & STE-FlowNet \etal \cite{ding2022steflownet} & 
            1.77                     & 14.70                     & \thickspace &    % indoor_flying1
            2.52                     & 26.10                     & \thickspace &    % indoor_flying2
            2.23                     & 22.10                     & \thickspace &    % indoor_flying3
            0.99                     & 3.90                      \\                 % outdoor_day1
            % --------------------------------------------------------------------------------
            \midrule
            
            \multirow{2}{*}{\rotatebox[origin=c]{90}{USL}}
            & EV-FlowNet \cite{zhu2019evflownet} & 
            2.18                     & 24.20                     & \thickspace &    % indoor_flying1
            3.85                     & 46.80                     & \thickspace &    % indoor_flying2
            3.18                     & 47.80                     & \thickspace &    % indoor_flying3
            \underline{1.30}         & \underline{9.70}          \\                 % outdoor_day1
            & ConvGRU-EV-FlowNet \cite{hagenaars2021convgruevflownet} & 
            2.16                     & 21.50                     & \thickspace &    % indoor_flying1
            3.90                     & 40.72                     & \thickspace &    % indoor_flying2
            3.00                     & 29.60                     & \thickspace &    % indoor_flying3
            1.69                     & 12.50                     \\                 % outdoor_day1
            % --------------------------------------------------------------------------------
            \midrule
            
            \multirow{3}{*}{\rotatebox[origin=c]{90}{MB}}
            & Shiba \etal \cite{shiba2022secrets} & 
            1.68                     & 12.79                     & \thickspace &    % indoor_flying1
            2.49                     & 26.31                     & \thickspace &    % indoor_flying2
            2.06                     & 18.93                     & \thickspace &    % indoor_flying3
            {2.07}                   & {19.99}                   \\                 % outdoor_day1
            & Ours (CM events only) & 
            \underline{1.53\tiny{9}} & \underline{9.48\tiny{2}}  & \thickspace &    % indoor_flying1
            \underline{2.38\tiny{2}} & \underline{21.84\tiny{9}} & \thickspace &    % indoor_flying2
            \underline{1.92\tiny{3}} & \underline{15.37\tiny{8}} & \thickspace &    % indoor_flying3
            {2.83\tiny{1}}           & {25.01\tiny{1}}           \\                 % outdoor_day1
            & Ours (EINCM) & 
            \textbf{1.43\tiny{9}}    & \textbf{7.78\tiny{9}}     & \thickspace &    % indoor_flying1
            \textbf{1.97\tiny{3}}    & \textbf{17.18\tiny{3}}    & \thickspace &    % indoor_flying2
            \textbf{1.70\tiny{6}}    & \textbf{12.33\tiny{9}}    & \thickspace &    % indoor_flying3
            {1.70\tiny{4}}           & {16.01\tiny{3}}           \\                 % outdoor_day1
            % ================================================================================
            \bottomrule
        
        \end{tabular}
    \end{adjustbox}
    \caption{Quantitative results on MVSEC. \emph{Bold} and \emph{underline}
    typefaces are used to indicate the \textbf{best} and the \underline{second
    best}, respectively. \%Out indicates a 3-pixel error percentage.}
    \label{tab:mvsec_accuracies}
\end{table}

%%%%%%%%%%%%%%%%%%%%%%%%%%%%%%%%%%%%%%%%%%%%%%%%%%%%%%%%%%%%%%%%%%%%%%%%%%%%%%%%

% ------------------------------------------------------------------------------
\subsection{MVSEC Evaluation}
\label{subsec:mvsec_evaluation}
We report quantitative results on the MVSEC dataset in
\cref{tab:mvsec_accuracies} against other prominent supervised learning (SL),
self-supervised learning (SSL), unsupervised learning (USL), and model-based
(MB) methods. The upper and lower parts of \cref{tab:mvsec_accuracies} show the
results corresponding to the $\emph{dt} = 1$ grayscale frame time interval
($\approx 22.2\,$ms) and $\emph{dt} = 4$ frames ($\approx 89\,$ms),
respectively. Our implementation of the \emph{events only} variant provides a
good baseline, which already beats the previous state-of-the-art MB method
\cite{shiba2022secrets} across sequences in both evaluation settings, except
for \texttt{outdoor\_day1}\footnote{In the literature, the evaluation of the
\texttt{outdoor\_day1} sequence is extremely inconsistent. Some works (\eg,
\cite{lee2020spikeflownet,ding2022steflownet}) use two disjoint collections of
evaluation points, while others (\eg, \cite{zhu2018evflownet,paredes2021back})
use a single one (please see the supplementary material for more details).}. 

%%%%%%%%%%%%%%%%%%%%%%%%%%%%%%%%%%%%%%%%%%%%%%%%%%%%%%%%%%%%%%%%%%%%%%%%%%%%%%%%
% Figure: MVSEC Comparisons
\begin{figure*}
    \centering
    \begin{adjustbox}{max width=\boxscaler\textwidth}
    \begin{tabular}{@{}c@{\thinspace}c@{\thinspace}c@{\thinspace}|@{\thinspace}c@{\thinspace}c@{\thinspace}c@{}}
        \vspace{-2pt}
        % --------------------------------------------------------------------------------
        % indoor_flying1
        \multirow{2}{*}{\rotatebox[origin=c]{90}{\begin{adjustbox}{max width=\labelscaler\textwidth} \texttt{indoor\_flying1} \end{adjustbox}}} &
        \begin{tabular}{@{}c@{}} \includegraphics[width=0.2\textwidth, cfbox=gray 0.1pt 0pt]{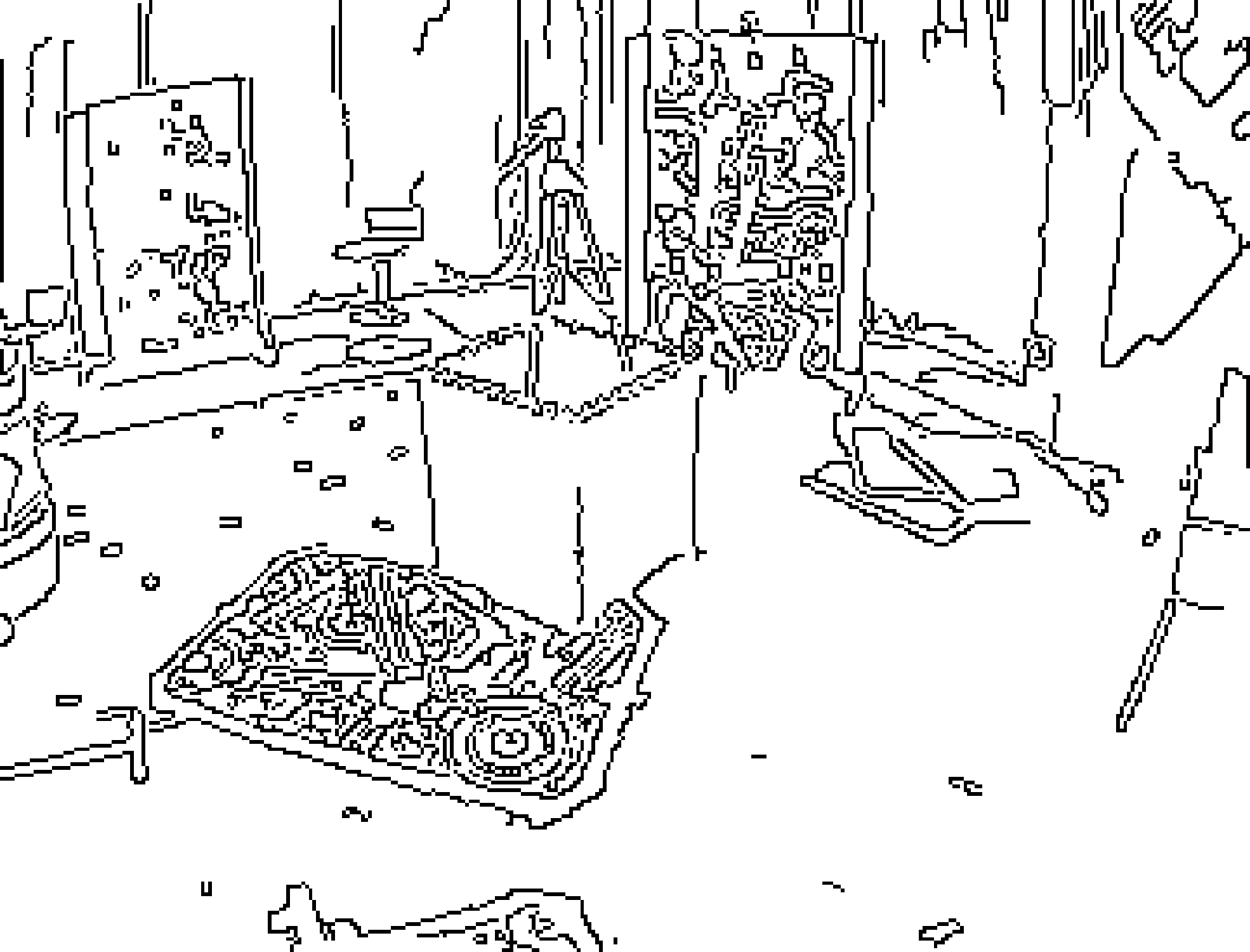} \end{tabular} &      % Edges
        \begin{tabular}{@{}c@{}} \includegraphics[width=0.2\textwidth, cfbox=gray 0.1pt 0pt]{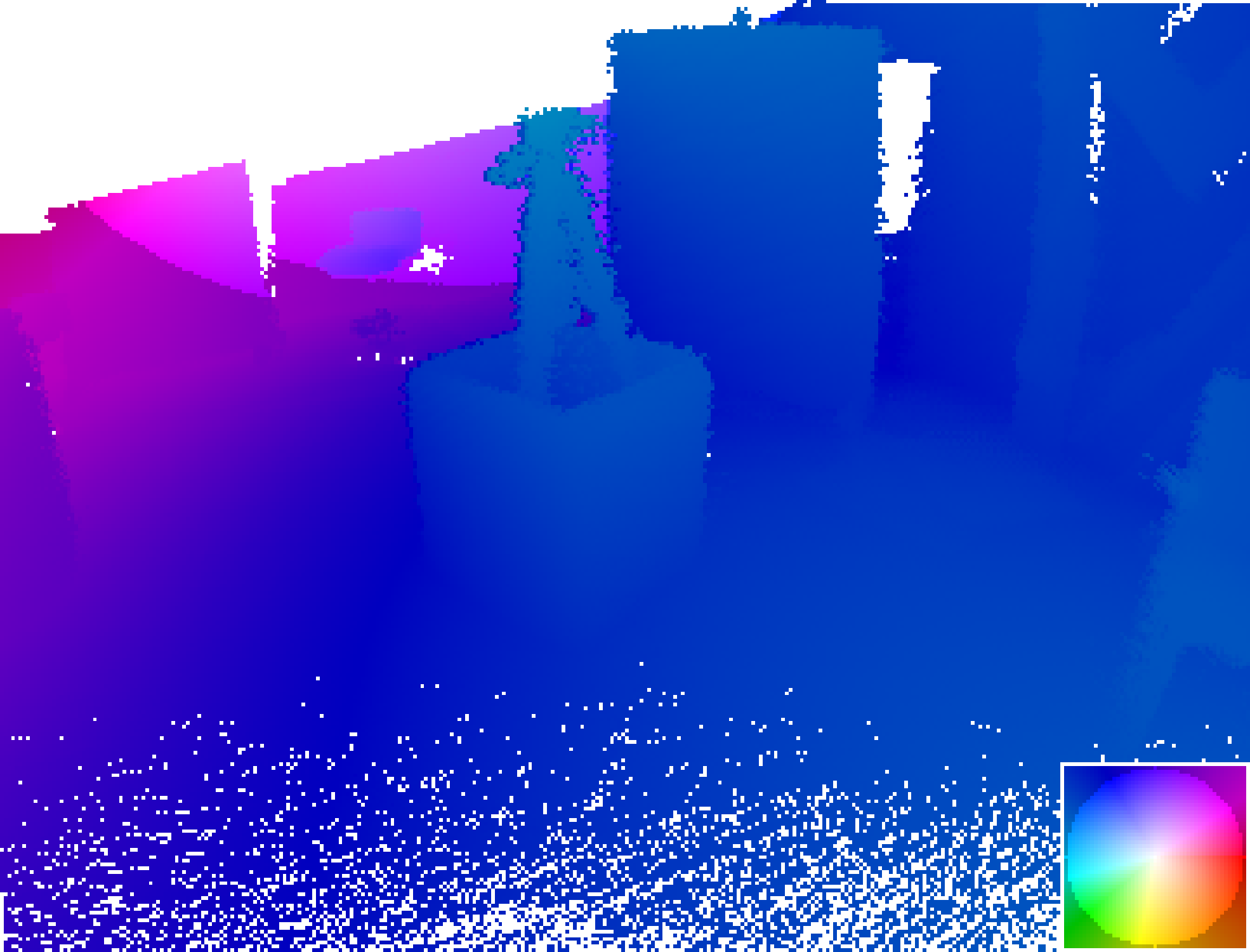} \end{tabular} &        % GT
        \begin{tabular}{@{}c@{}} \includegraphics[width=0.2\textwidth, cfbox=gray 0.1pt 0pt]{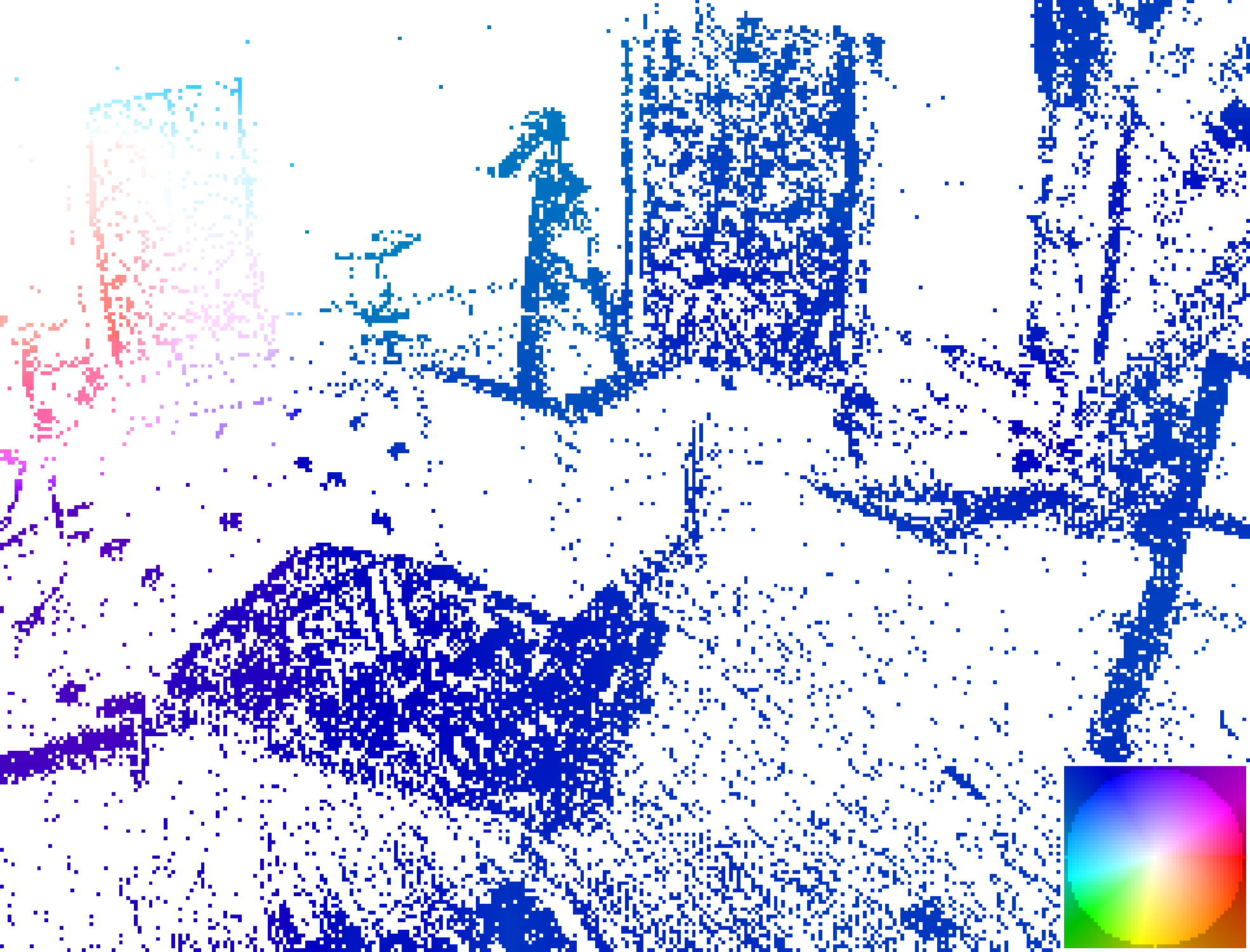} \end{tabular} &  % Ours (predicted flow masked by input events)
        \begin{tabular}{@{}c@{}} \includegraphics[width=0.203\textwidth, cfbox=gray 0.1pt 0pt]{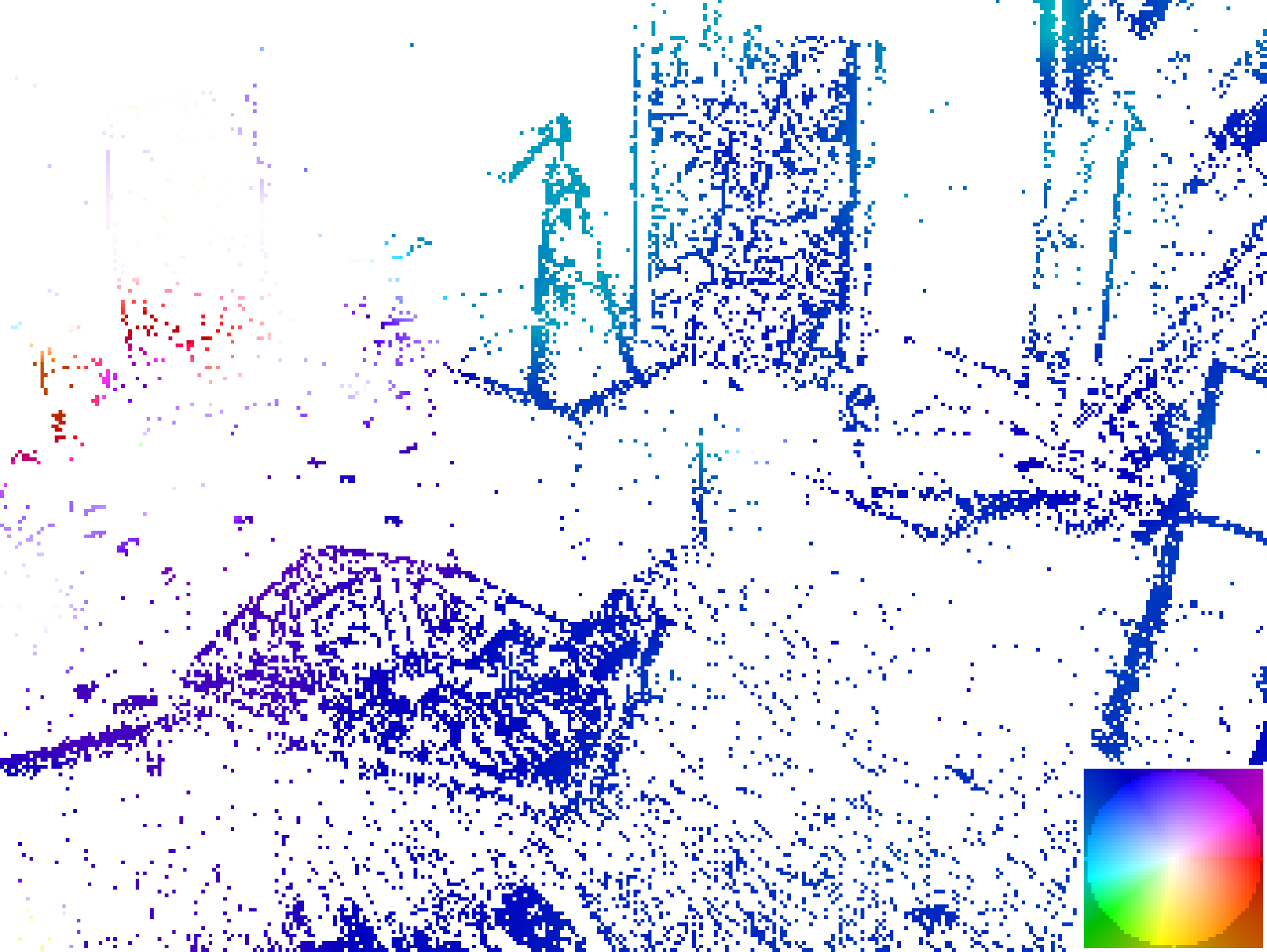} \end{tabular} &                         % MultiCM (predicted flow masked by input events)
        \begin{tabular}{@{}c@{}} \includegraphics[width=0.152\textwidth, cfbox=gray 0.1pt 0pt]{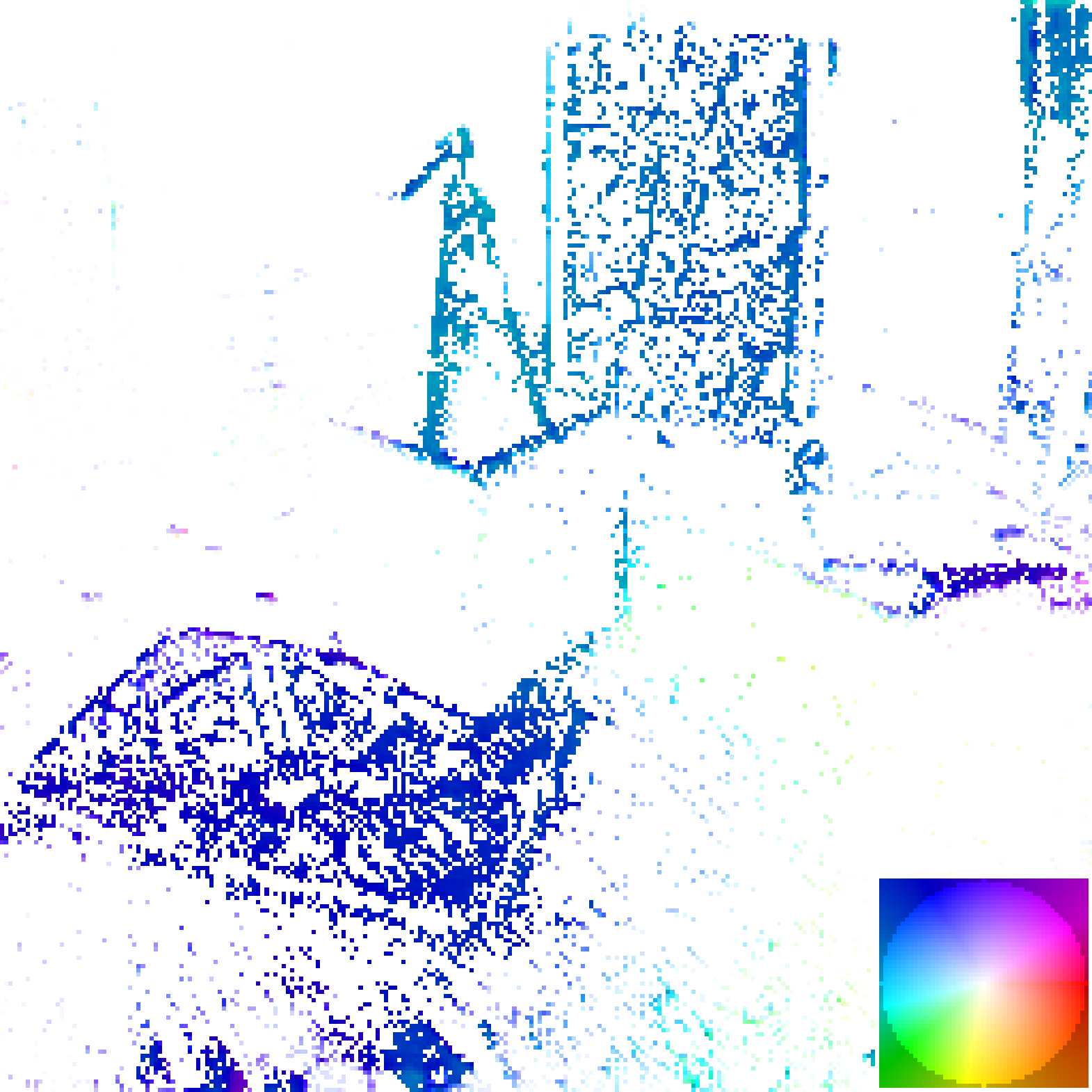} \end{tabular} \\ \vspace{-2pt}          % EV-FlowNet (predicted flow masked by input events)

        &
        \begin{tabular}{@{}c@{}} \includegraphics[width=0.2\textwidth, cfbox=gray 0.1pt 0pt]{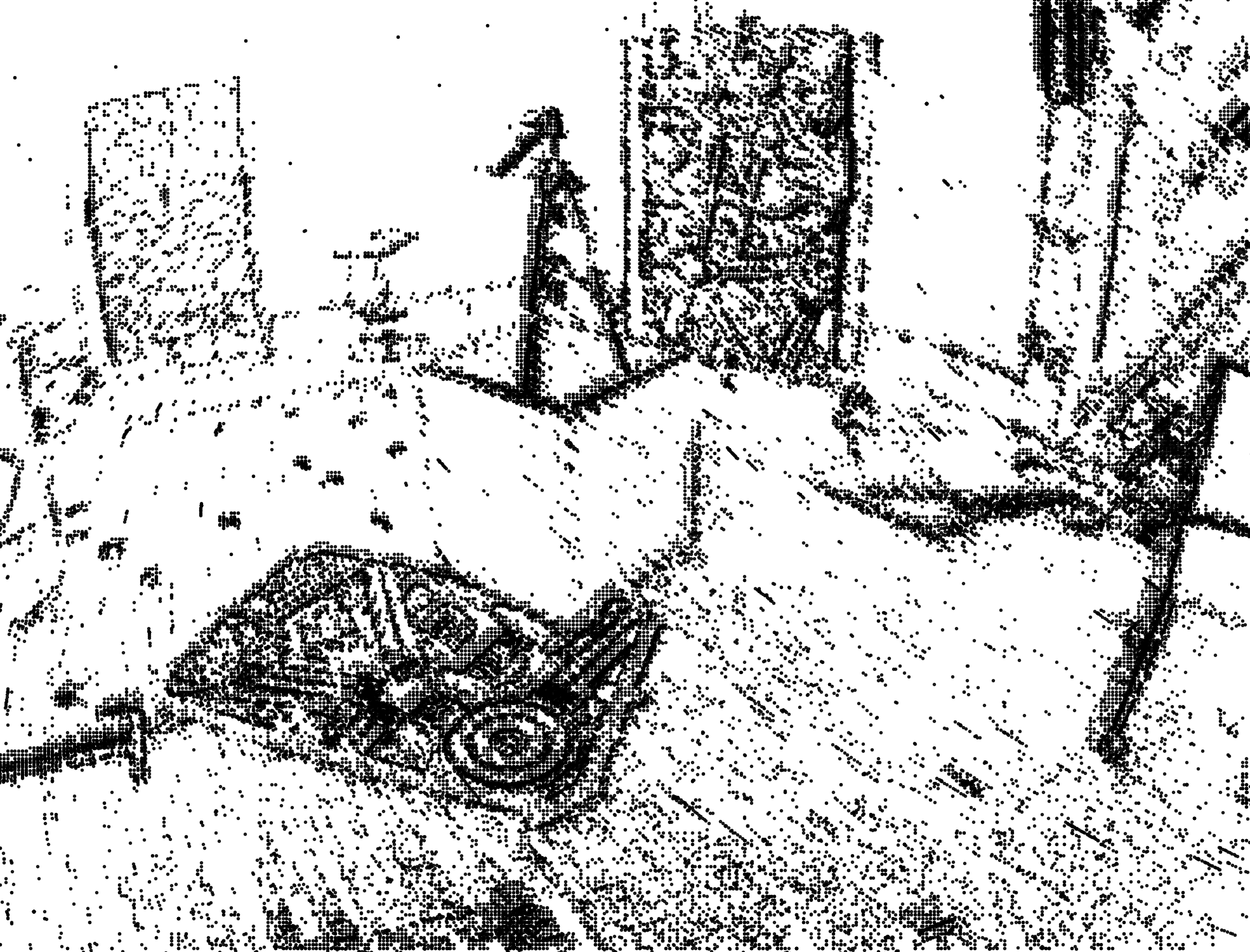} \end{tabular} &       % Original events
        \begin{tabular}{@{}c@{}} \includegraphics[width=0.2\textwidth, cfbox=gray 0.1pt 0pt]{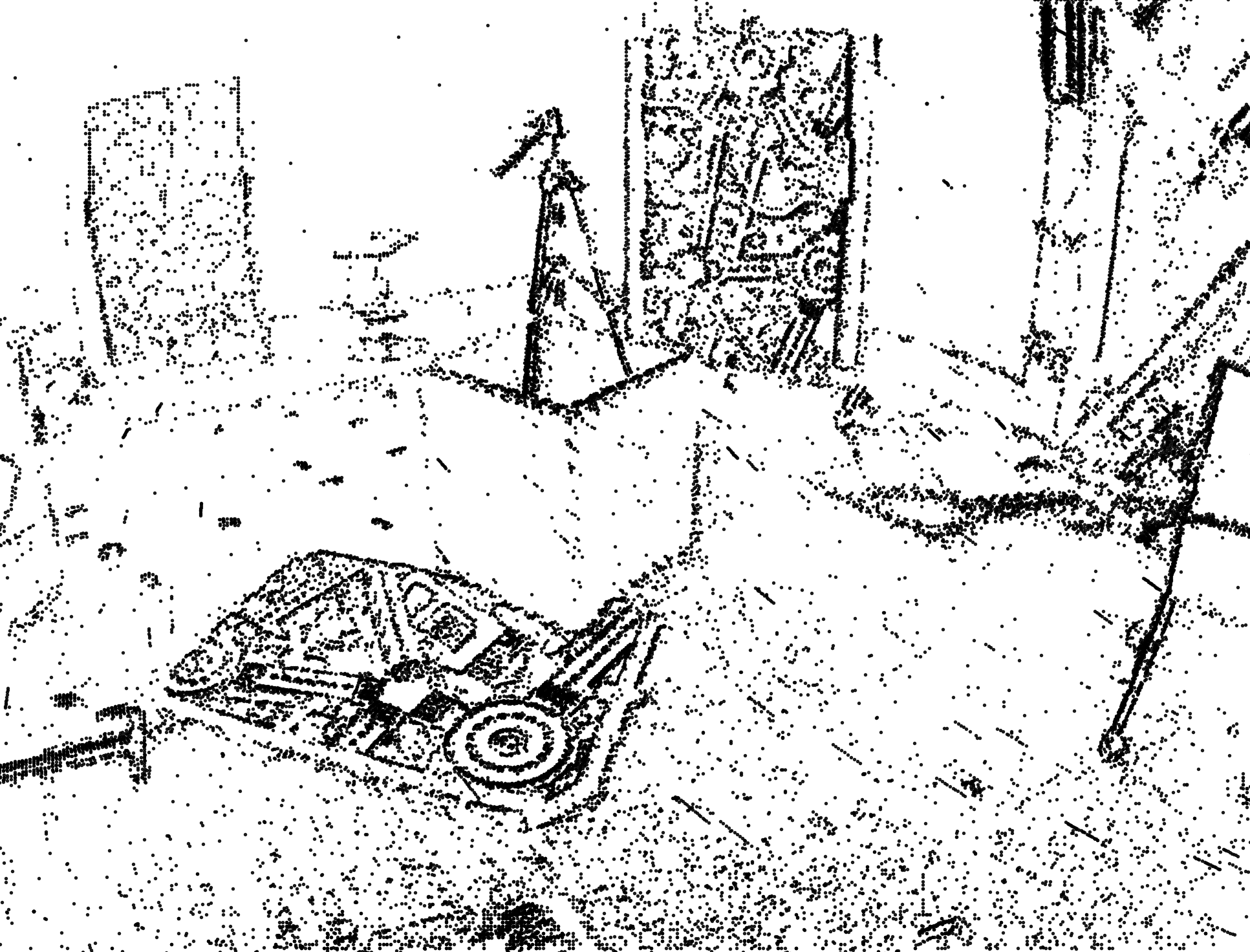} \end{tabular} &    % GT IWE
        \begin{tabular}{@{}c@{}} \includegraphics[width=0.2\textwidth, cfbox=gray 0.1pt 0pt]{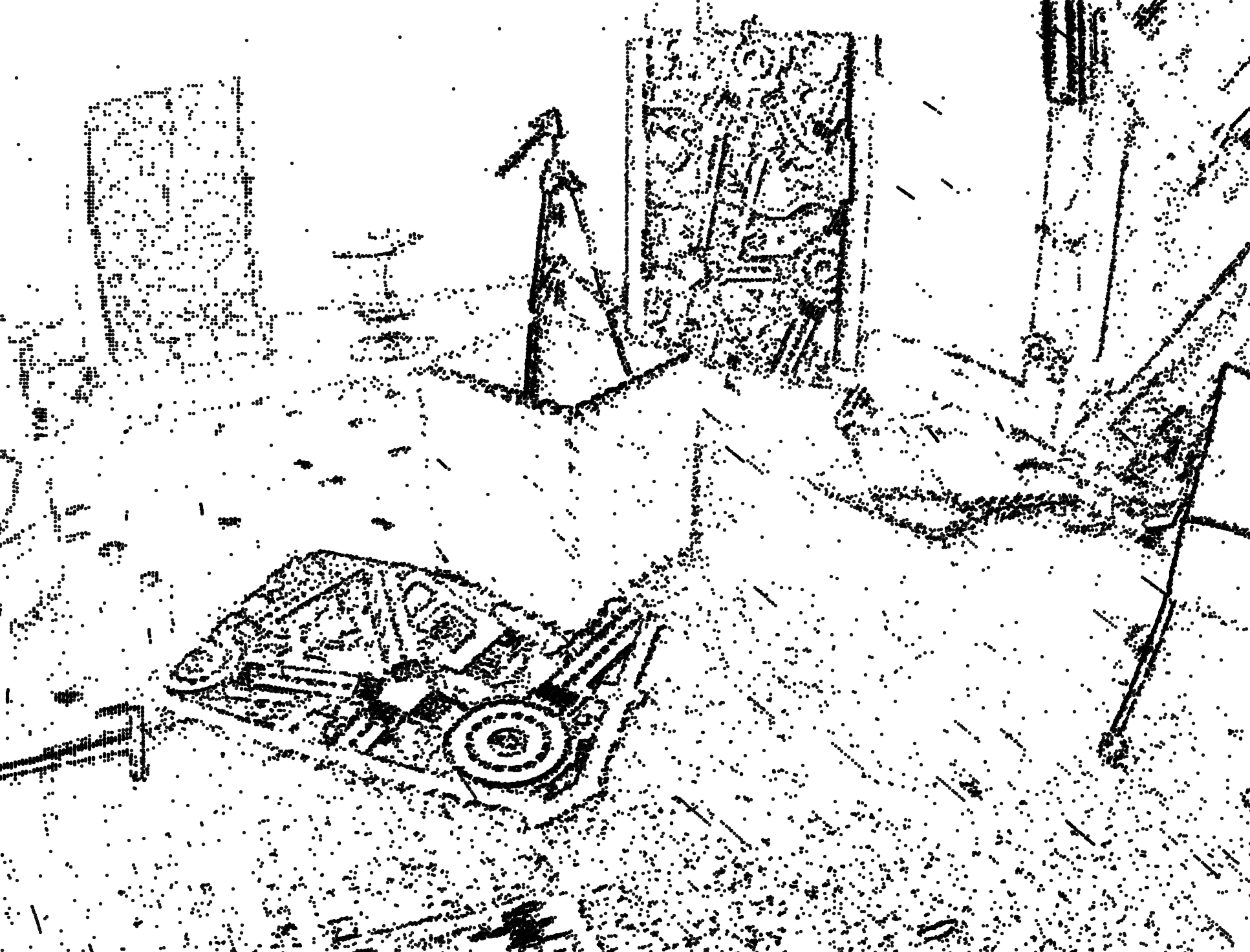} \end{tabular} & % Ours IWE
        \begin{tabular}{@{}c@{}} \includegraphics[width=0.2\textwidth, cfbox=gray 0.1pt 0pt]{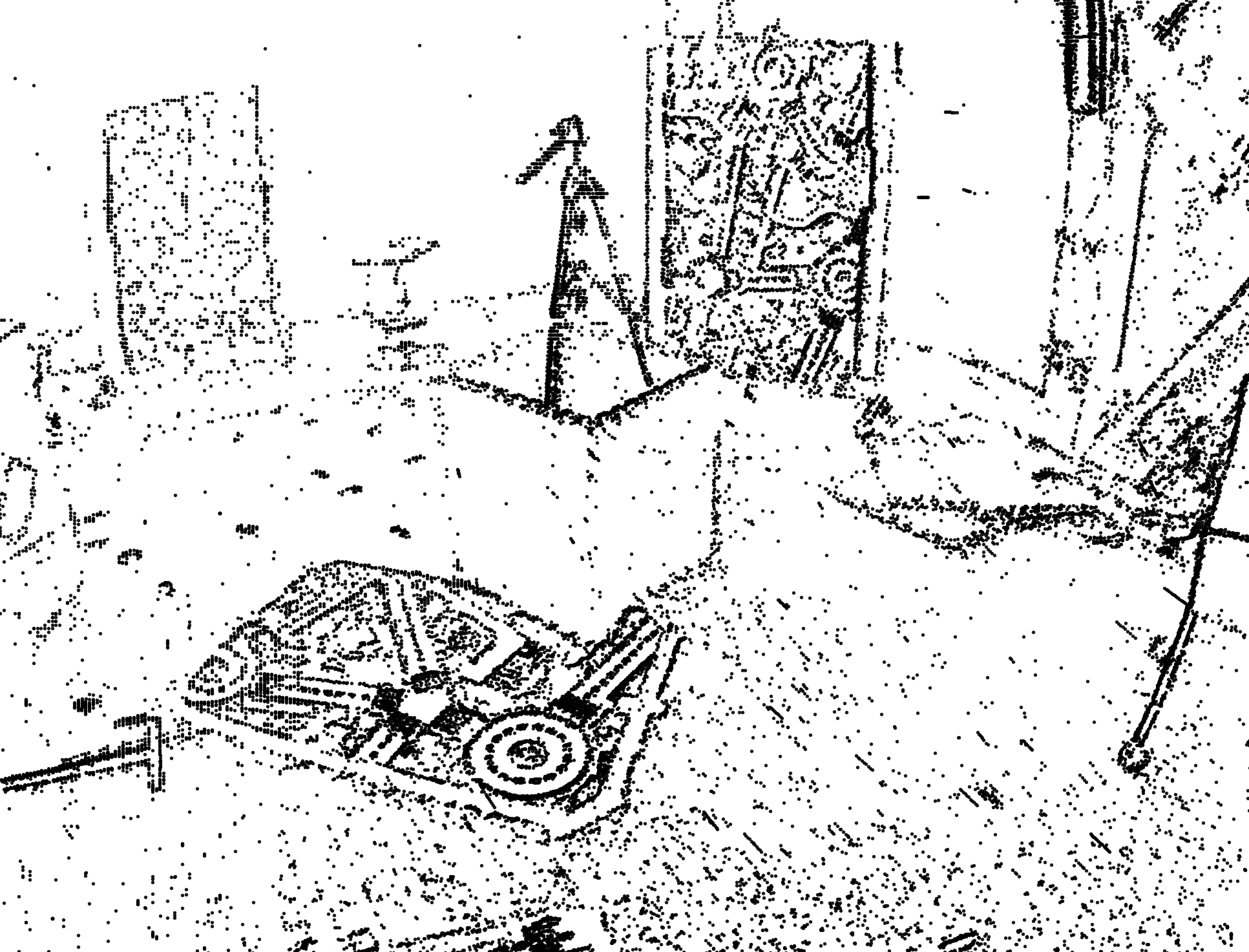} \end{tabular} &                         % MultiCM IWE
        \begin{tabular}{@{}c@{}} \includegraphics[width=0.152\textwidth, cfbox=gray 0.1pt 0pt]{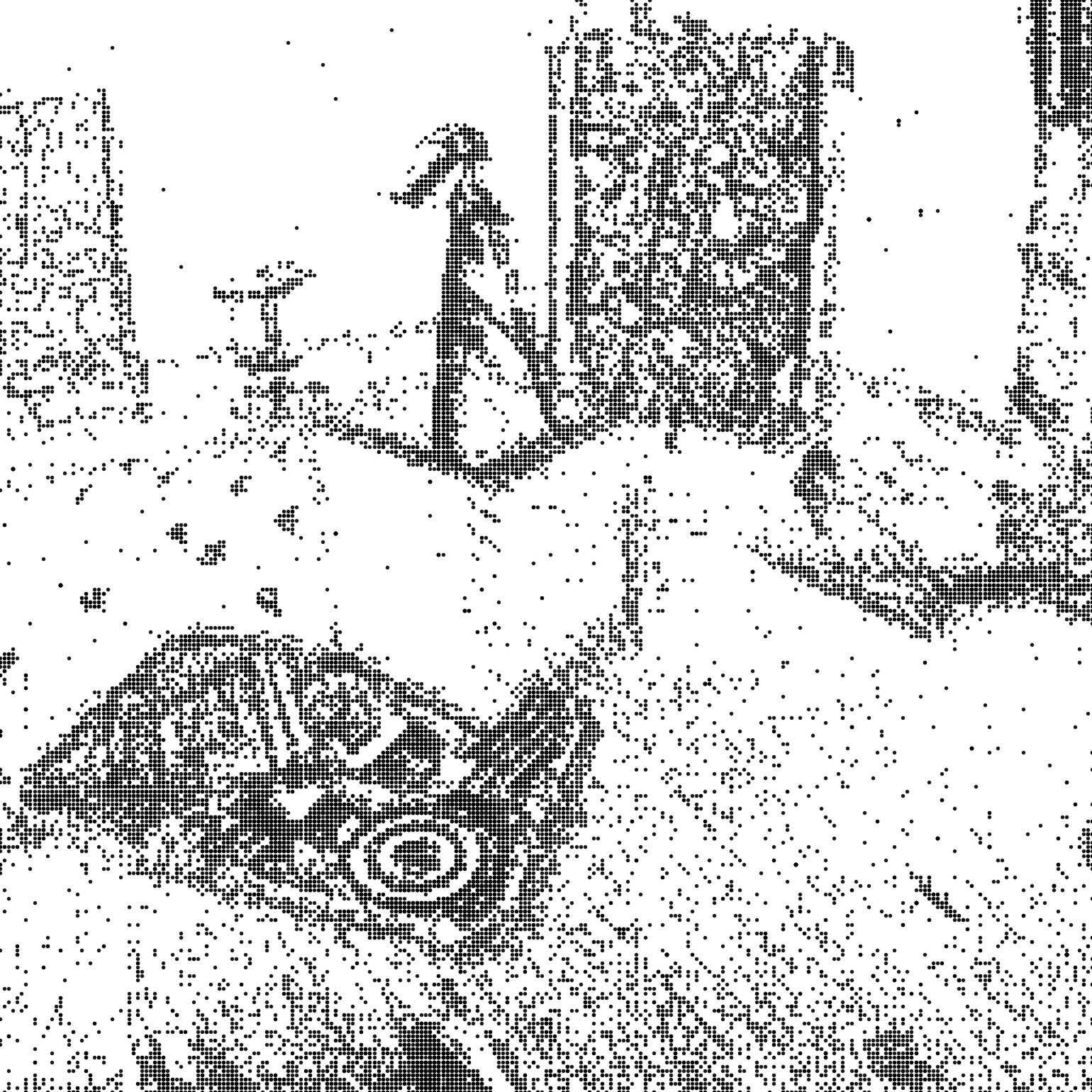} \end{tabular} \\ \vspace{-2pt}          % EV-FlowNet IWE

        % --------------------------------------------------------------------------------
        % indoor_flying2
        \multirow{2}{*}{\rotatebox[origin=c]{90}{\begin{adjustbox}{max width=\labelscaler\textwidth} \texttt{indoor\_flying2} \end{adjustbox}}} &
        \begin{tabular}{@{}c@{}} \includegraphics[width=0.2\textwidth, cfbox=gray 0.1pt 0pt]{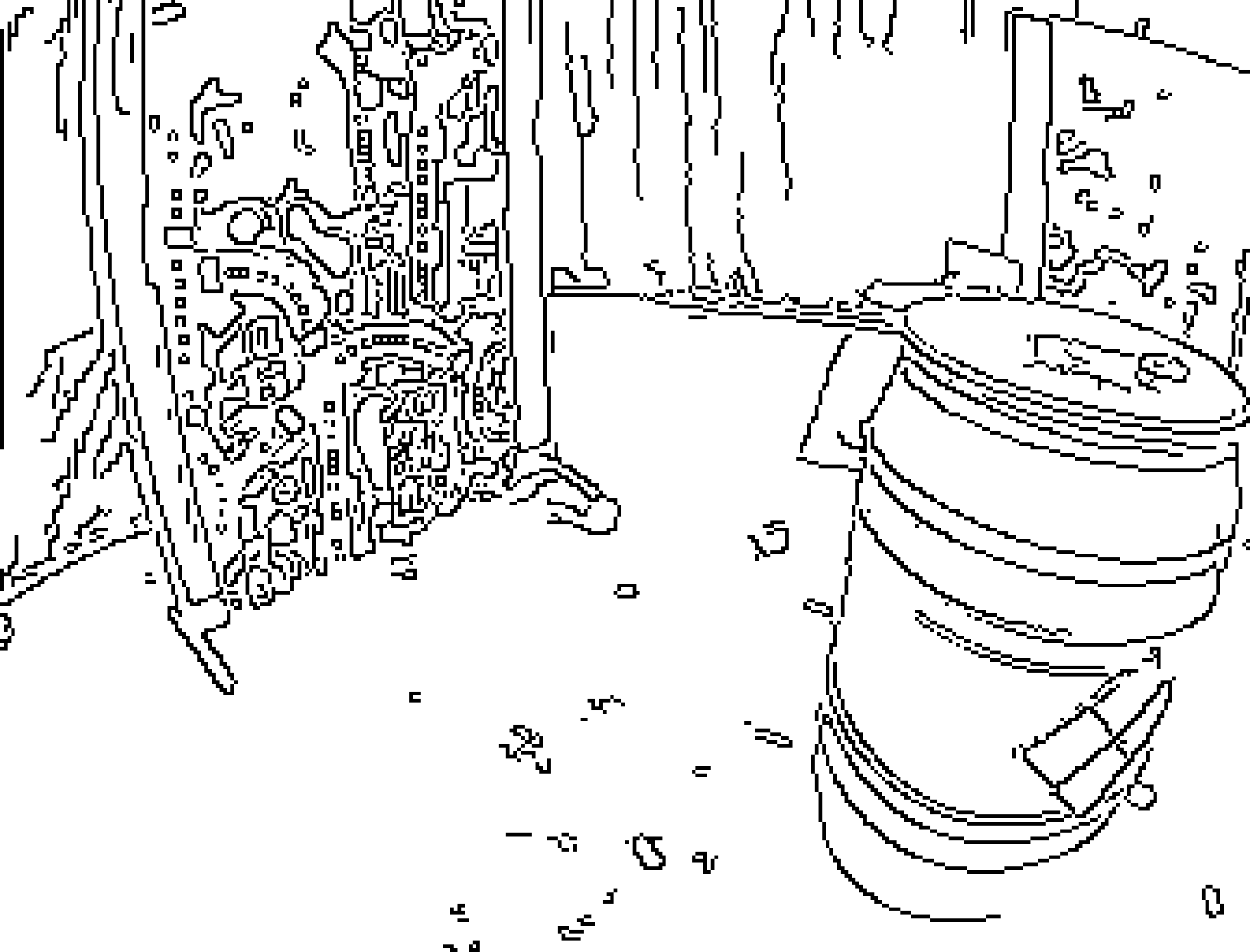} \end{tabular} &      % Edges 
        \begin{tabular}{@{}c@{}} \includegraphics[width=0.2\textwidth, cfbox=gray 0.1pt 0pt]{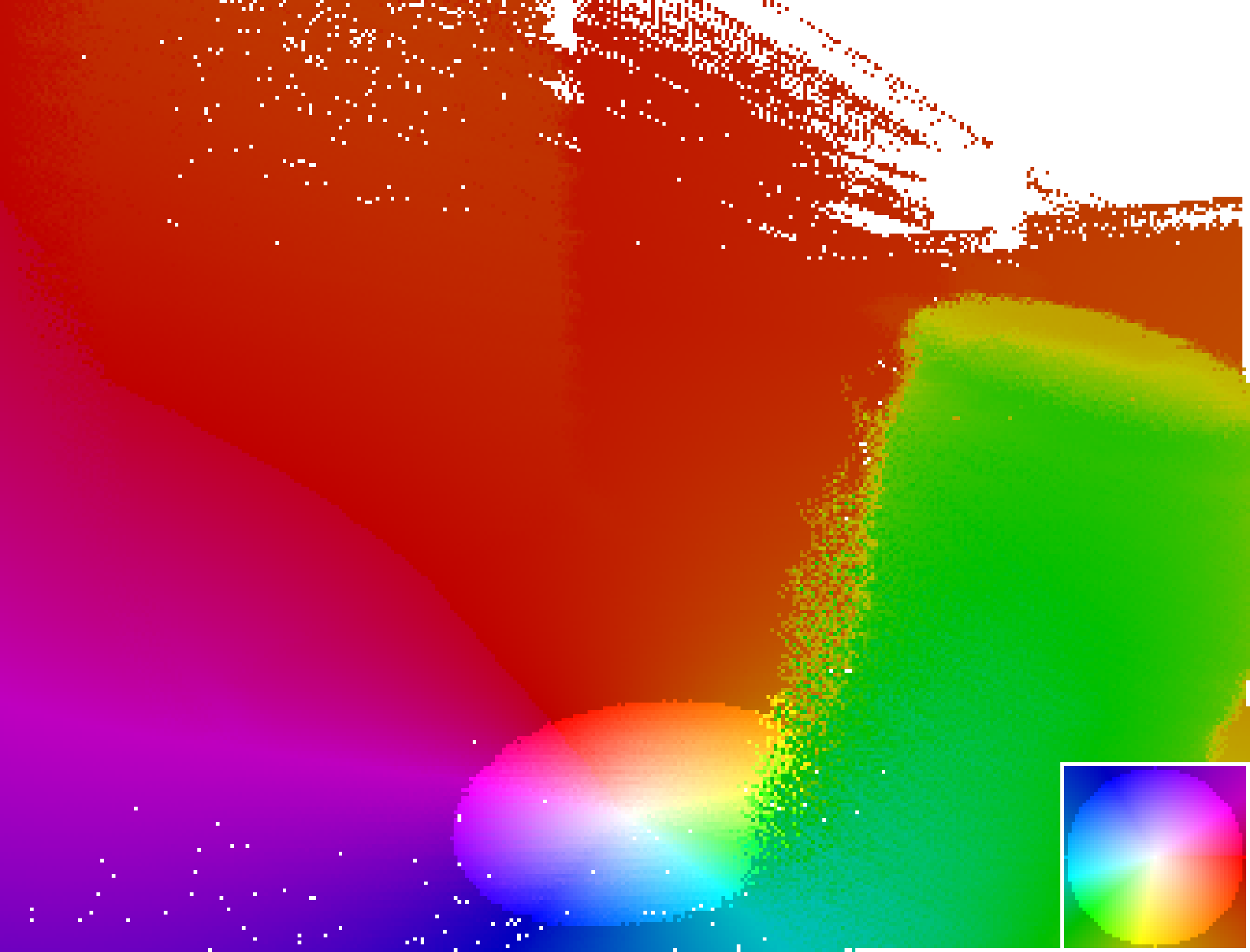} \end{tabular} &        % GT
        \begin{tabular}{@{}c@{}} \includegraphics[width=0.2\textwidth, cfbox=gray 0.1pt 0pt]{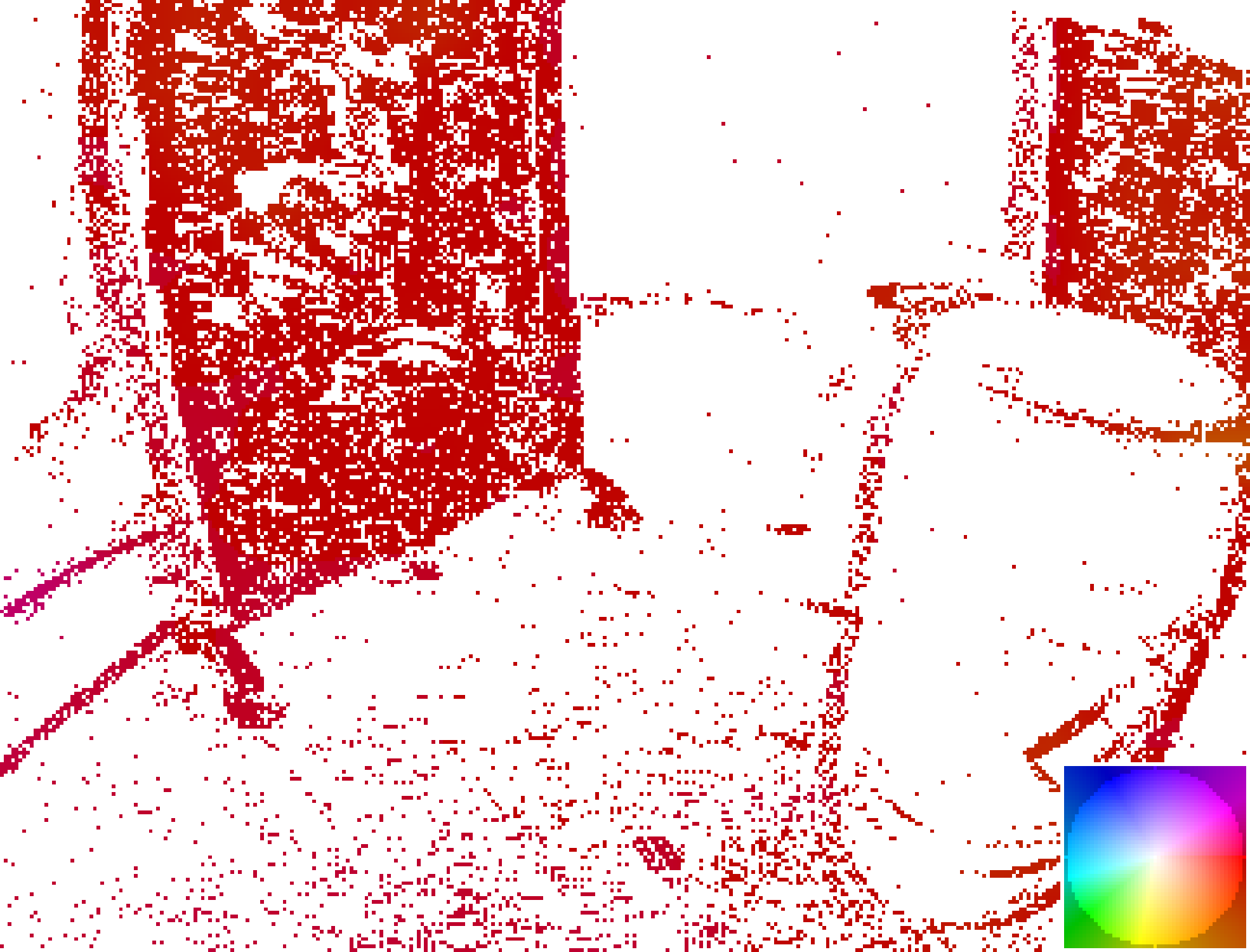} \end{tabular} &  % Ours (predicted flow masked by input events)
        \begin{tabular}{@{}c@{}} \includegraphics[width=0.203\textwidth, cfbox=gray 0.1pt 0pt]{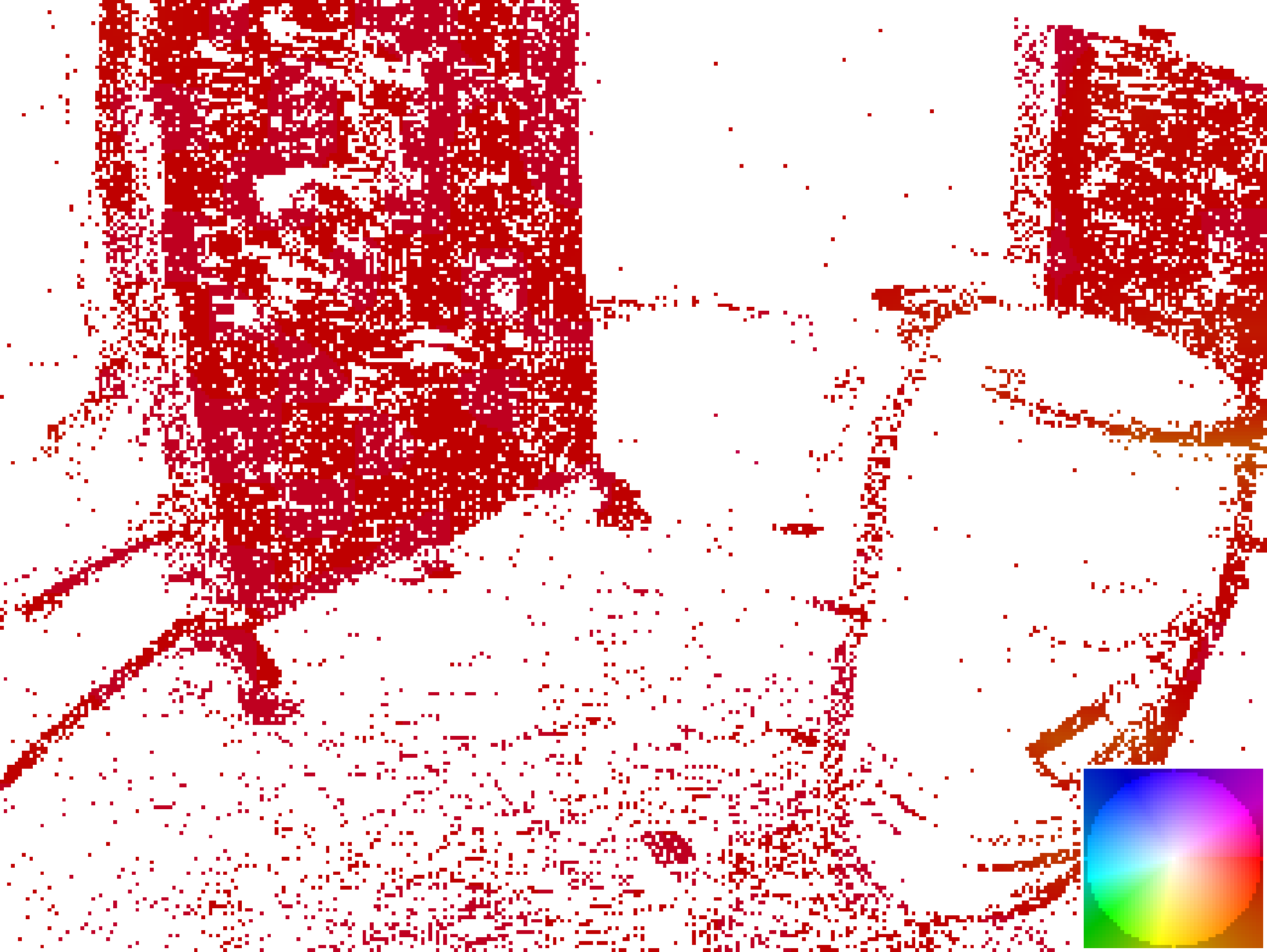} \end{tabular} &                         % MultiCM (predicted flow masked by input events)
        \begin{tabular}{@{}c@{}} \includegraphics[width=0.152\textwidth, cfbox=gray 0.1pt 0pt]{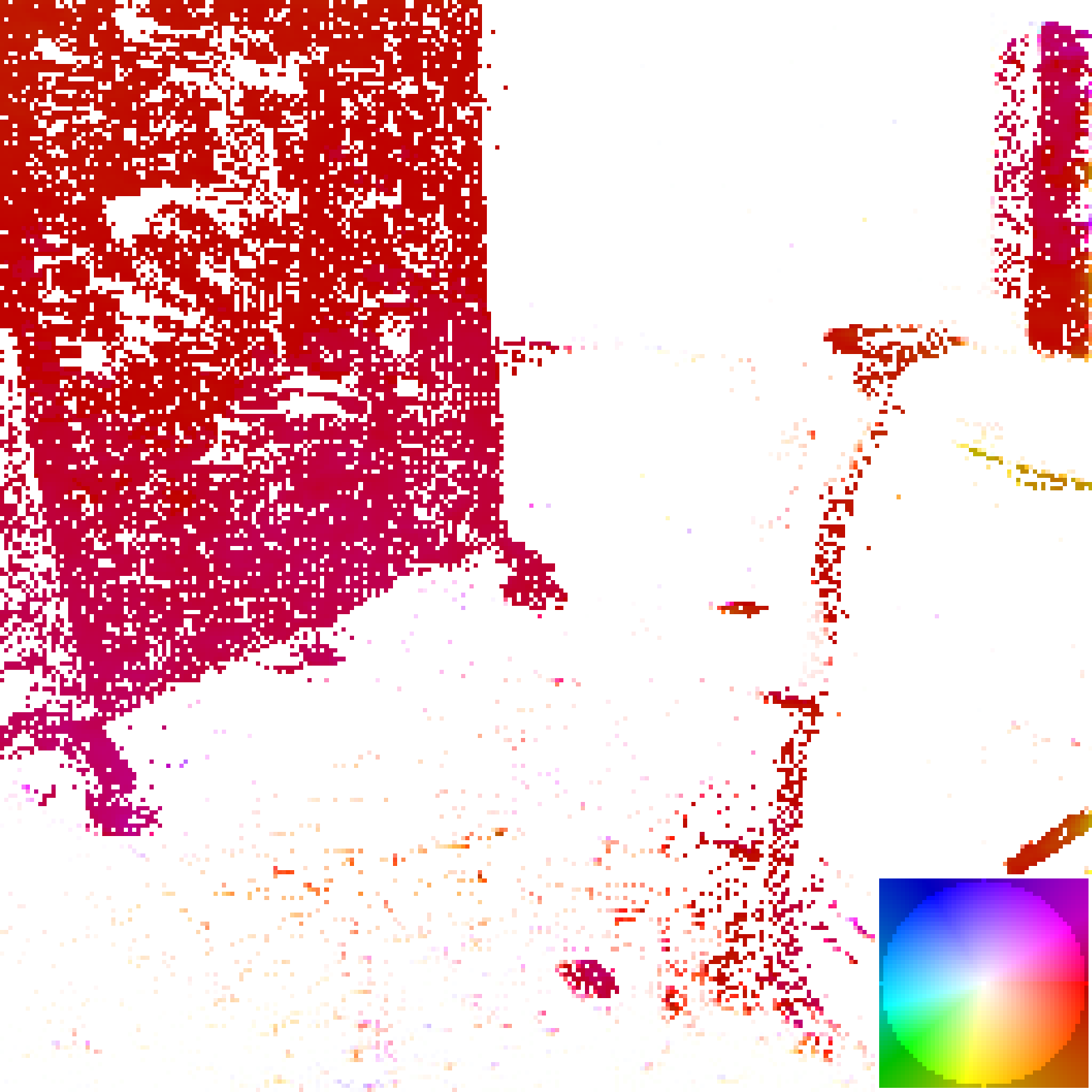} \end{tabular} \\ \vspace{-2pt}          % EV-FlowNet (predicted flow masked by input events)

        &
        \begin{tabular}{@{}c@{}} \includegraphics[width=0.2\textwidth, cfbox=gray 0.1pt 0pt]{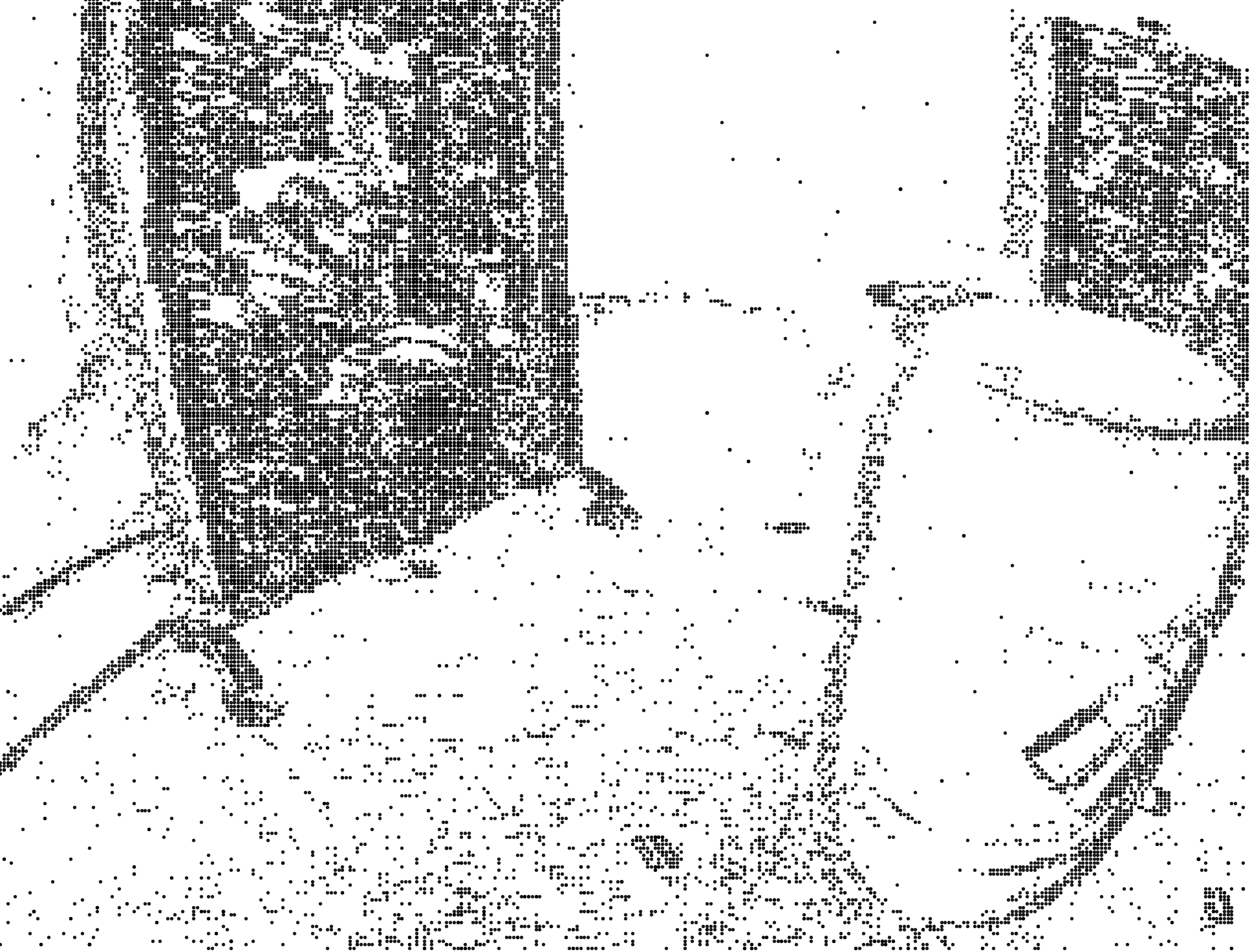} \end{tabular} &       % Original events
        \begin{tabular}{@{}c@{}} \includegraphics[width=0.2\textwidth, cfbox=gray 0.1pt 0pt]{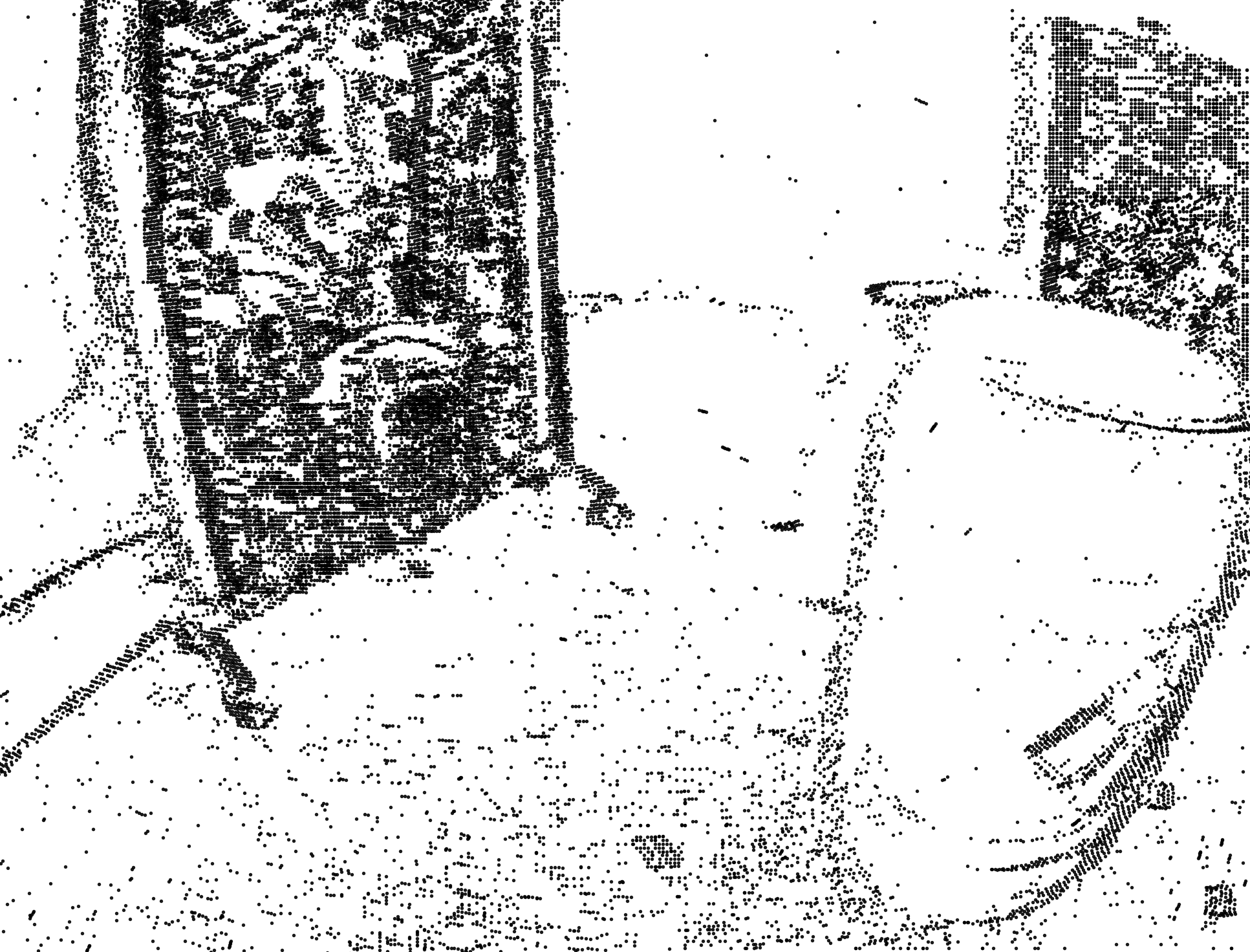} \end{tabular} &    % GT IWE
        \begin{tabular}{@{}c@{}} \includegraphics[width=0.2\textwidth, cfbox=gray 0.1pt 0pt]{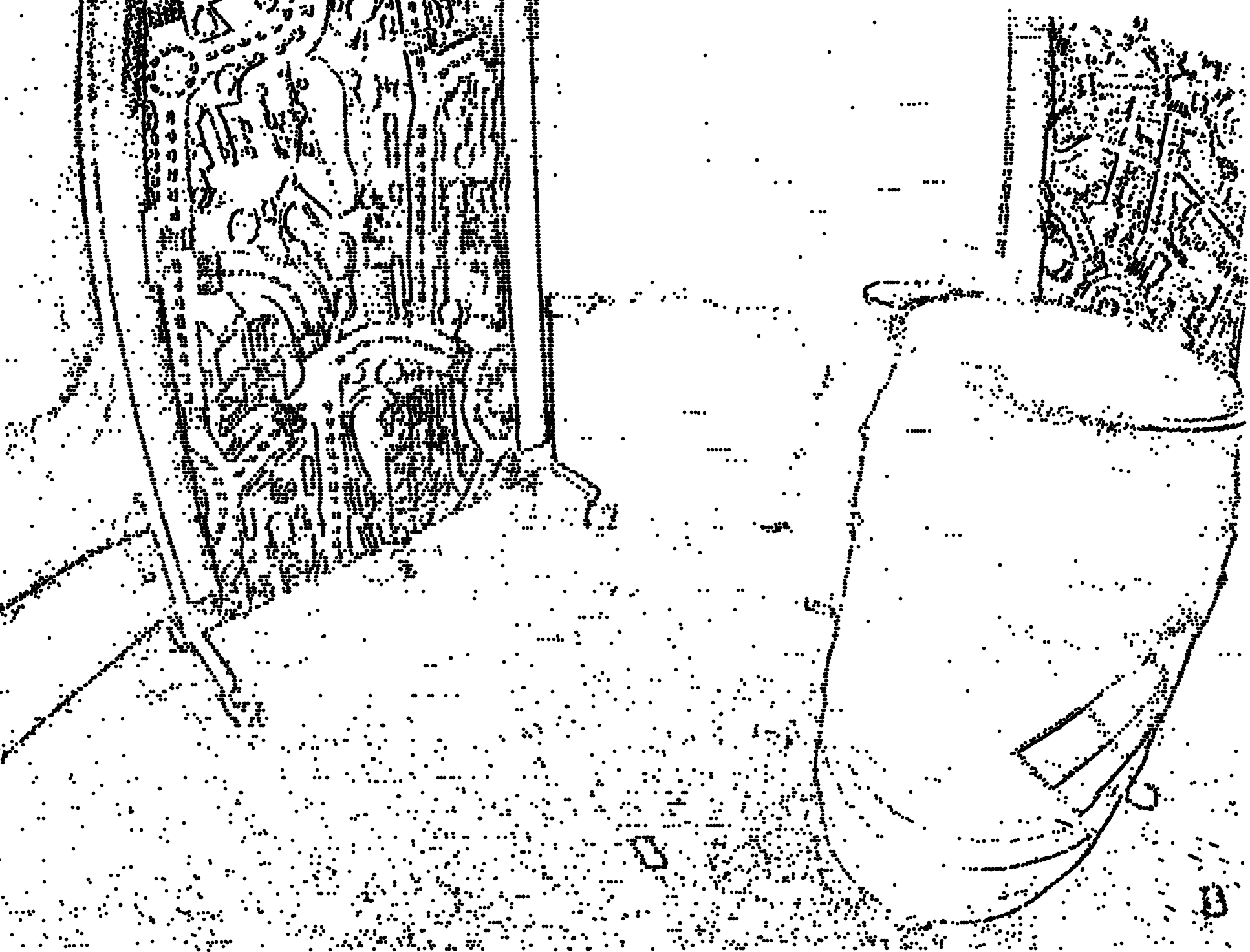} \end{tabular} & % Ours IWE
        \begin{tabular}{@{}c@{}} \includegraphics[width=0.2\textwidth, cfbox=gray 0.1pt 0pt]{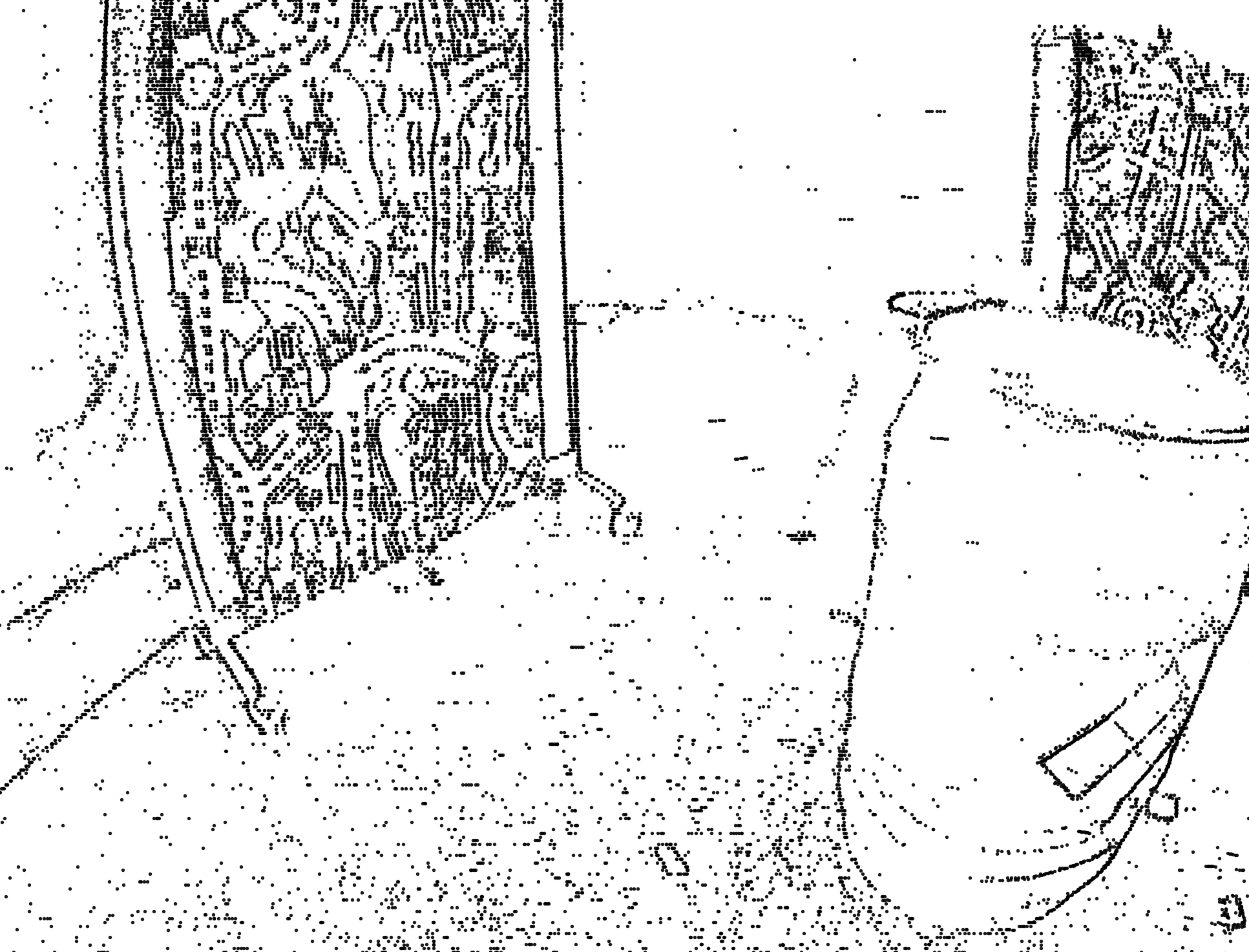} \end{tabular} &                         % MultiCM IWE
        \begin{tabular}{@{}c@{}} \includegraphics[width=0.152\textwidth, cfbox=gray 0.1pt 0pt]{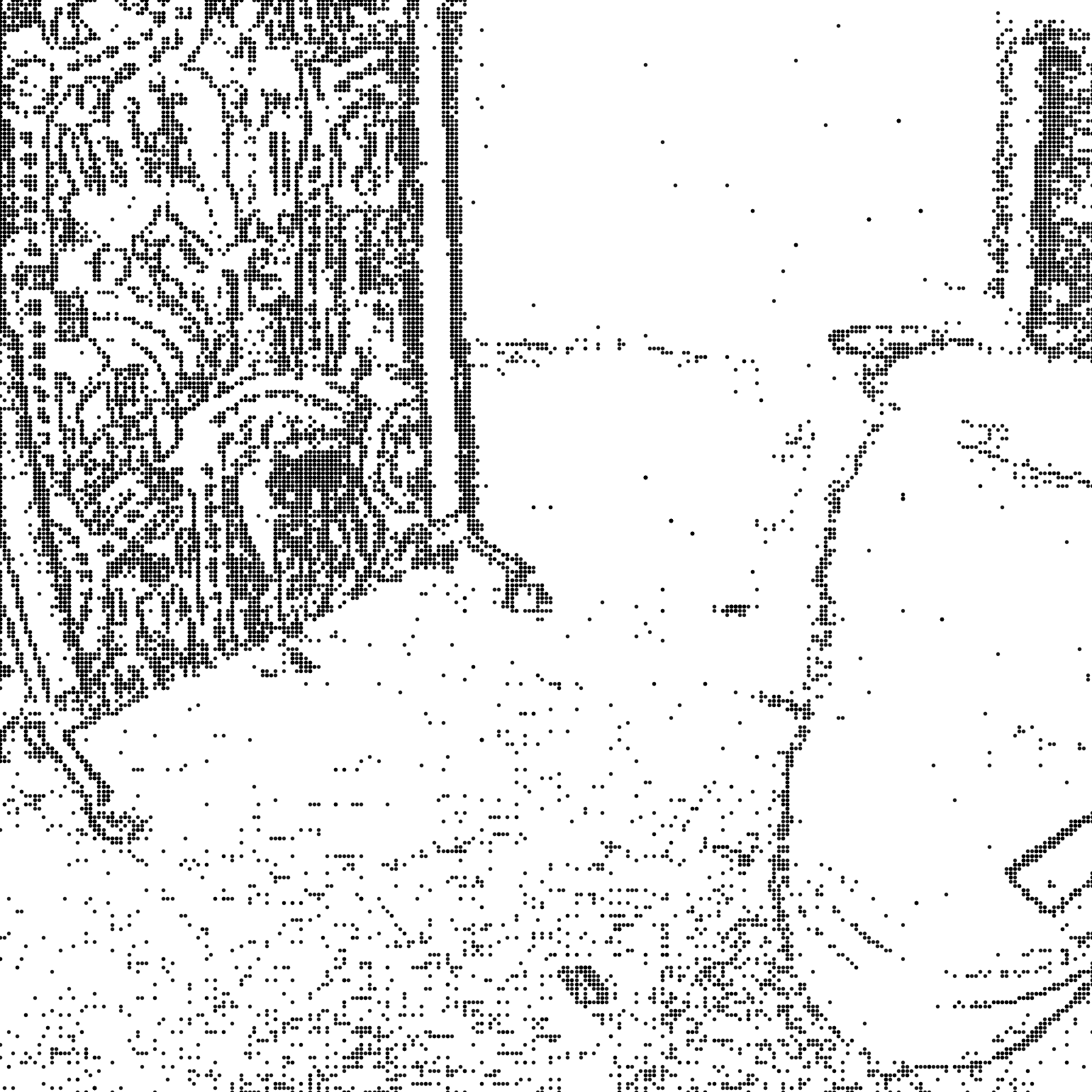} \end{tabular} \\ \vspace{-2pt}          % EV-FlowNet IWE

        % --------------------------------------------------------------------------------
        % indoor_flying3
        \multirow{2}{*}{\rotatebox[origin=c]{90}{\begin{adjustbox}{max width=\labelscaler\textwidth} \texttt{indoor\_flying3} \end{adjustbox}}} &
        \begin{tabular}{@{}c@{}} \includegraphics[width=0.2\textwidth, cfbox=gray 0.1pt 0pt]{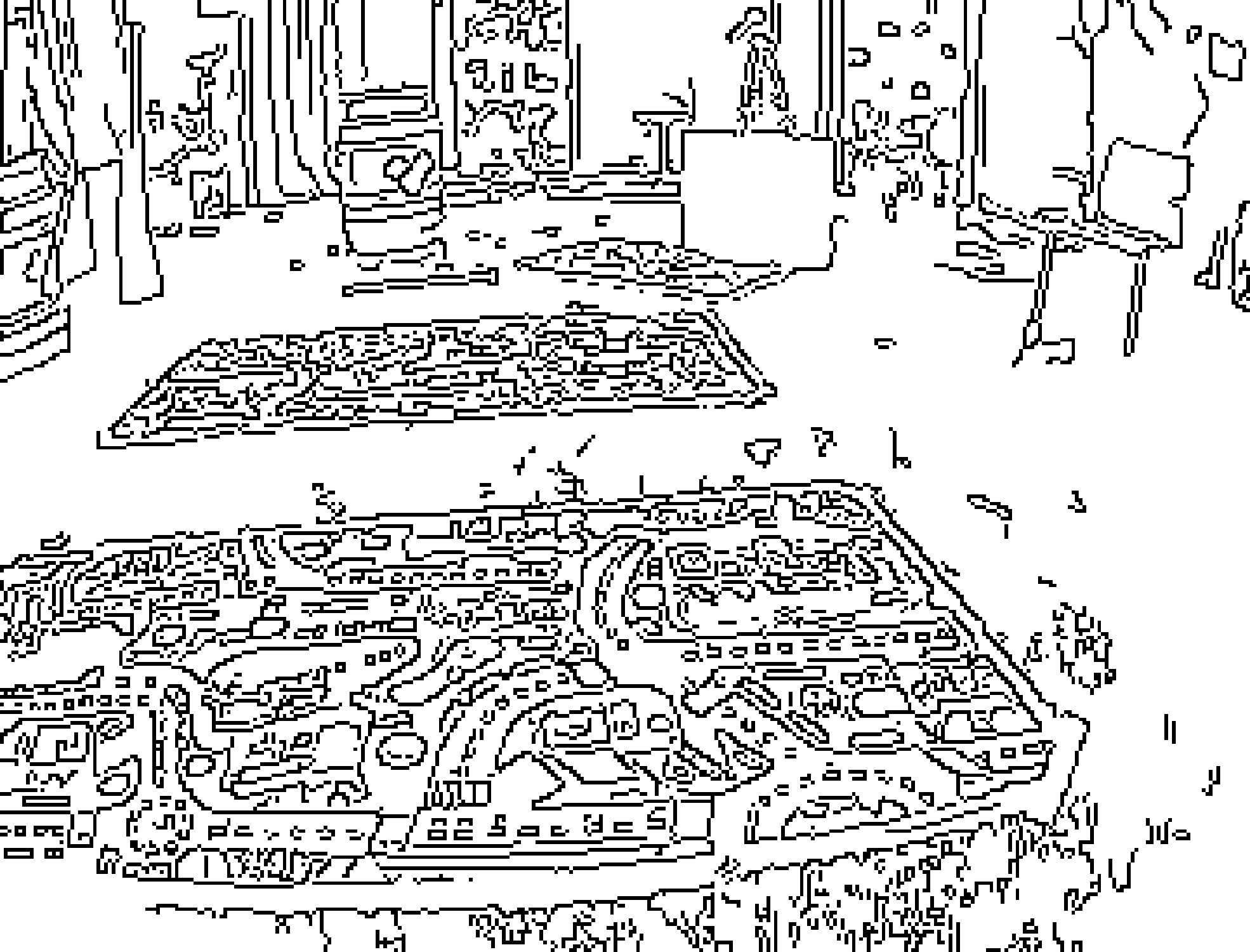} \end{tabular} &      % Edges
        \begin{tabular}{@{}c@{}} \includegraphics[width=0.2\textwidth, cfbox=gray 0.1pt 0pt]{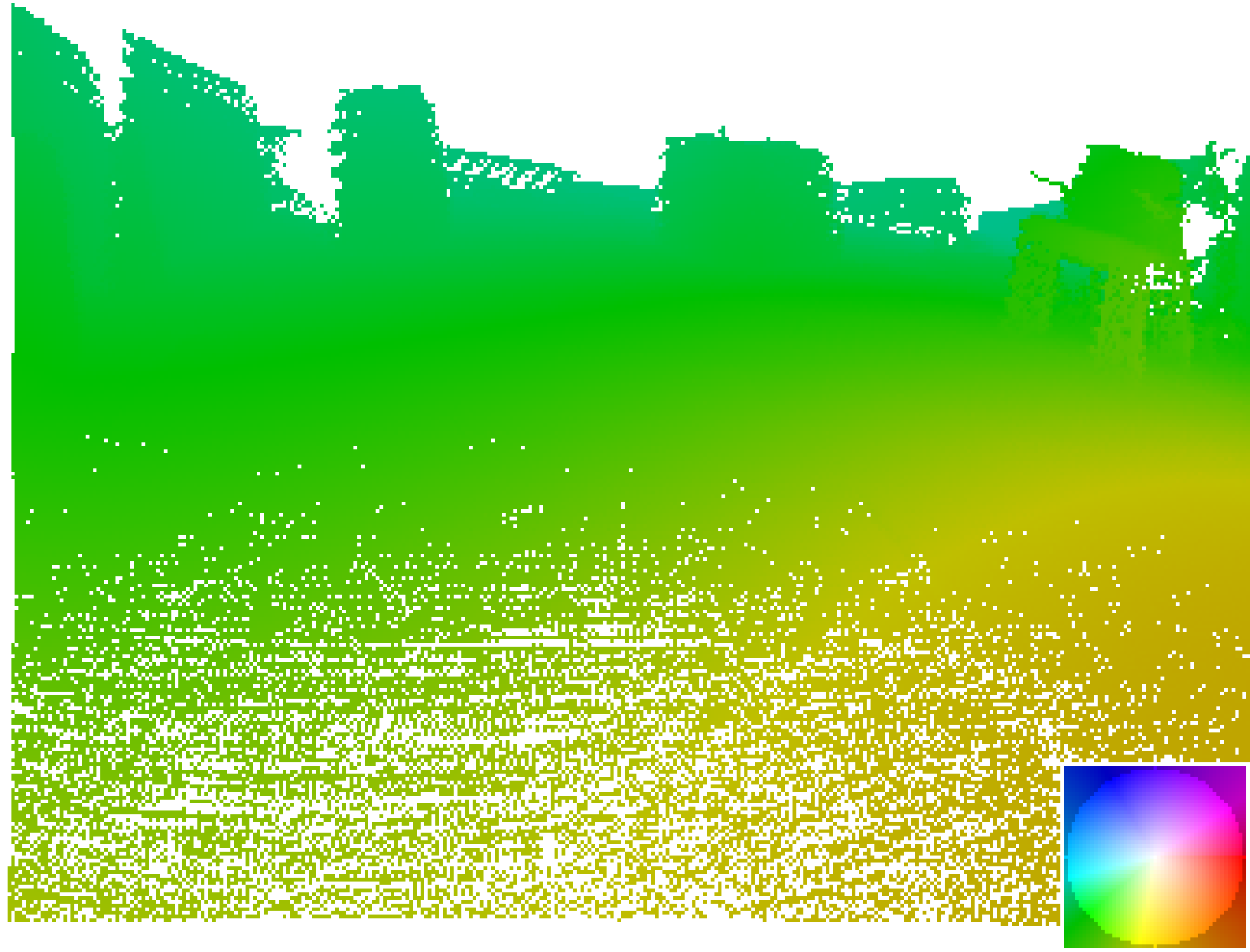} \end{tabular} &        % GT
        \begin{tabular}{@{}c@{}} \includegraphics[width=0.2\textwidth, cfbox=gray 0.1pt 0pt]{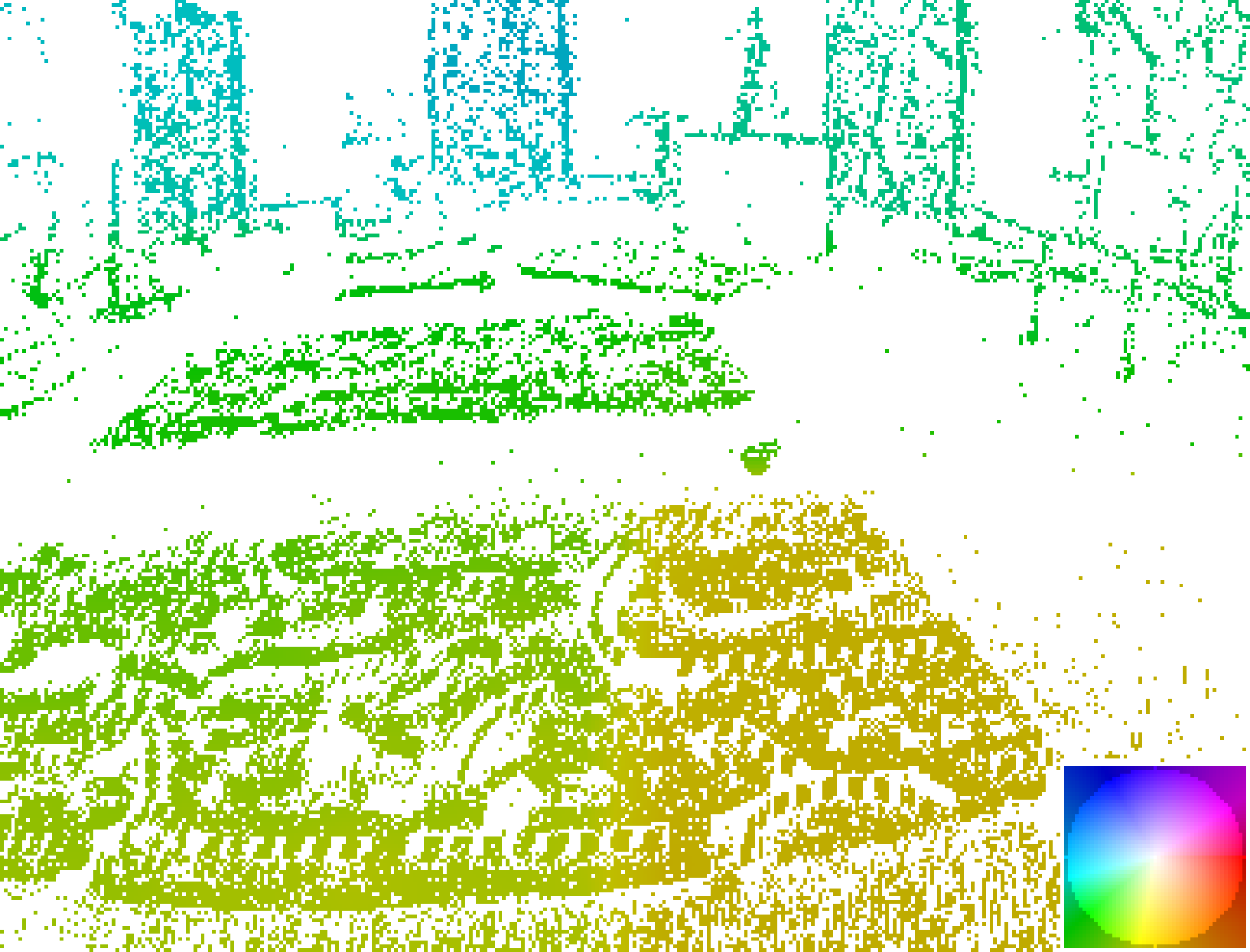} \end{tabular} &  % Ours (predicted flow masked by input events)
        \begin{tabular}{@{}c@{}} \includegraphics[width=0.203\textwidth, cfbox=gray 0.1pt 0pt]{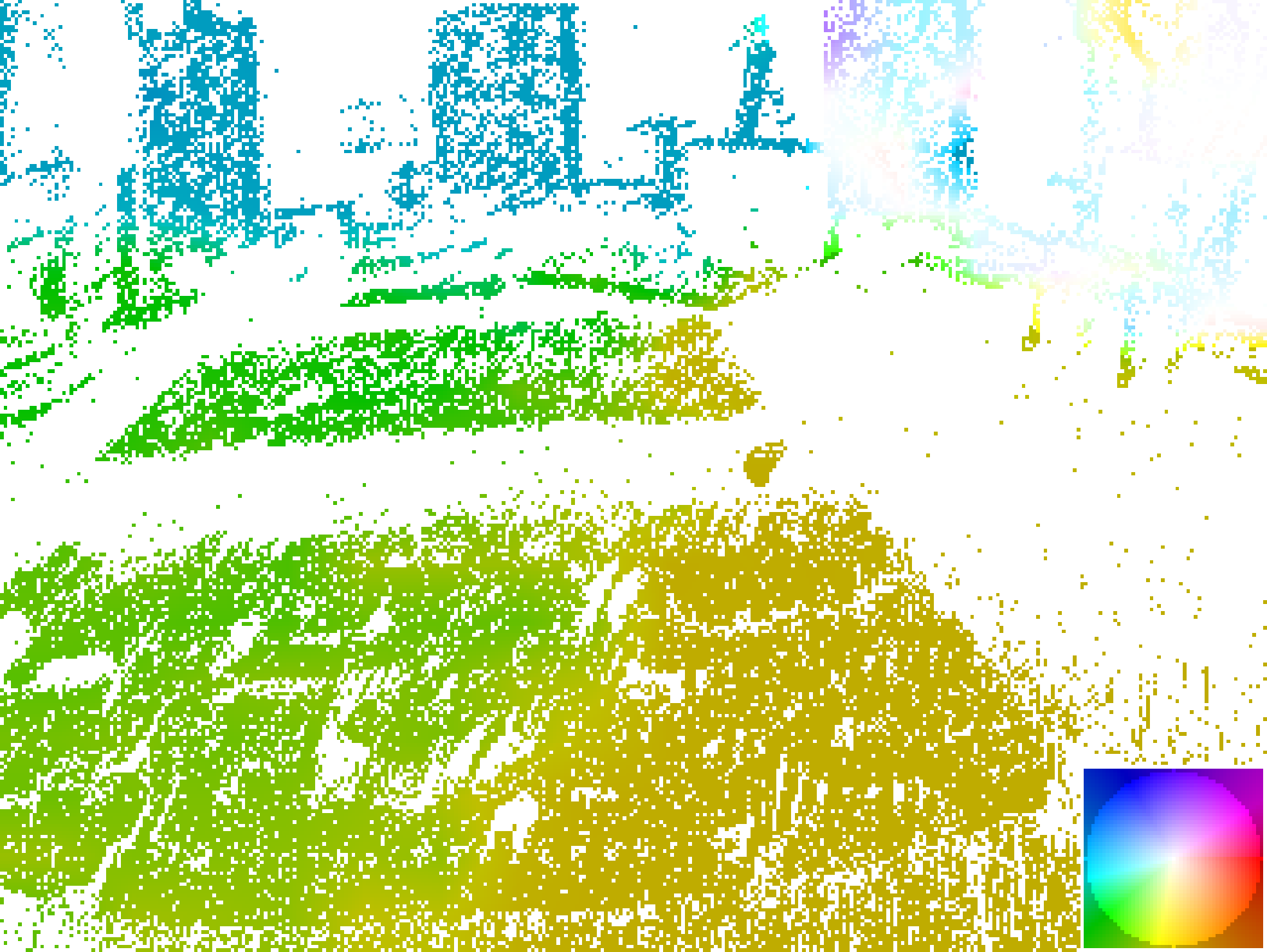} \end{tabular} &                         % MultiCM (predicted flow masked by input events)
        \begin{tabular}{@{}c@{}} \includegraphics[width=0.152\textwidth, cfbox=gray 0.1pt 0pt]{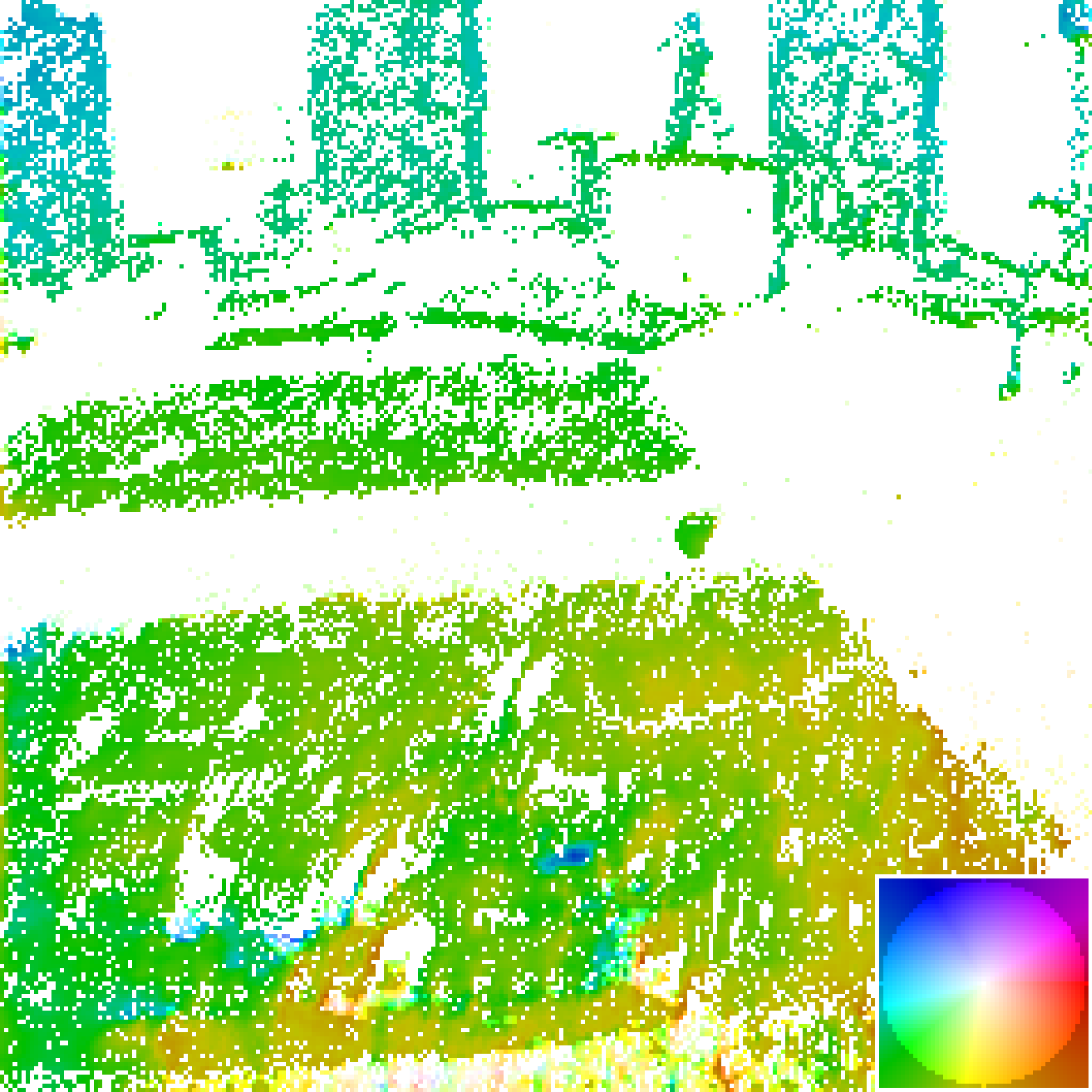} \end{tabular} \\ \vspace{-2pt}          % EV-FlowNet (predicted flow masked by input events)

        &
        \begin{tabular}{@{}c@{}} \includegraphics[width=0.2\textwidth, cfbox=gray 0.1pt 0pt]{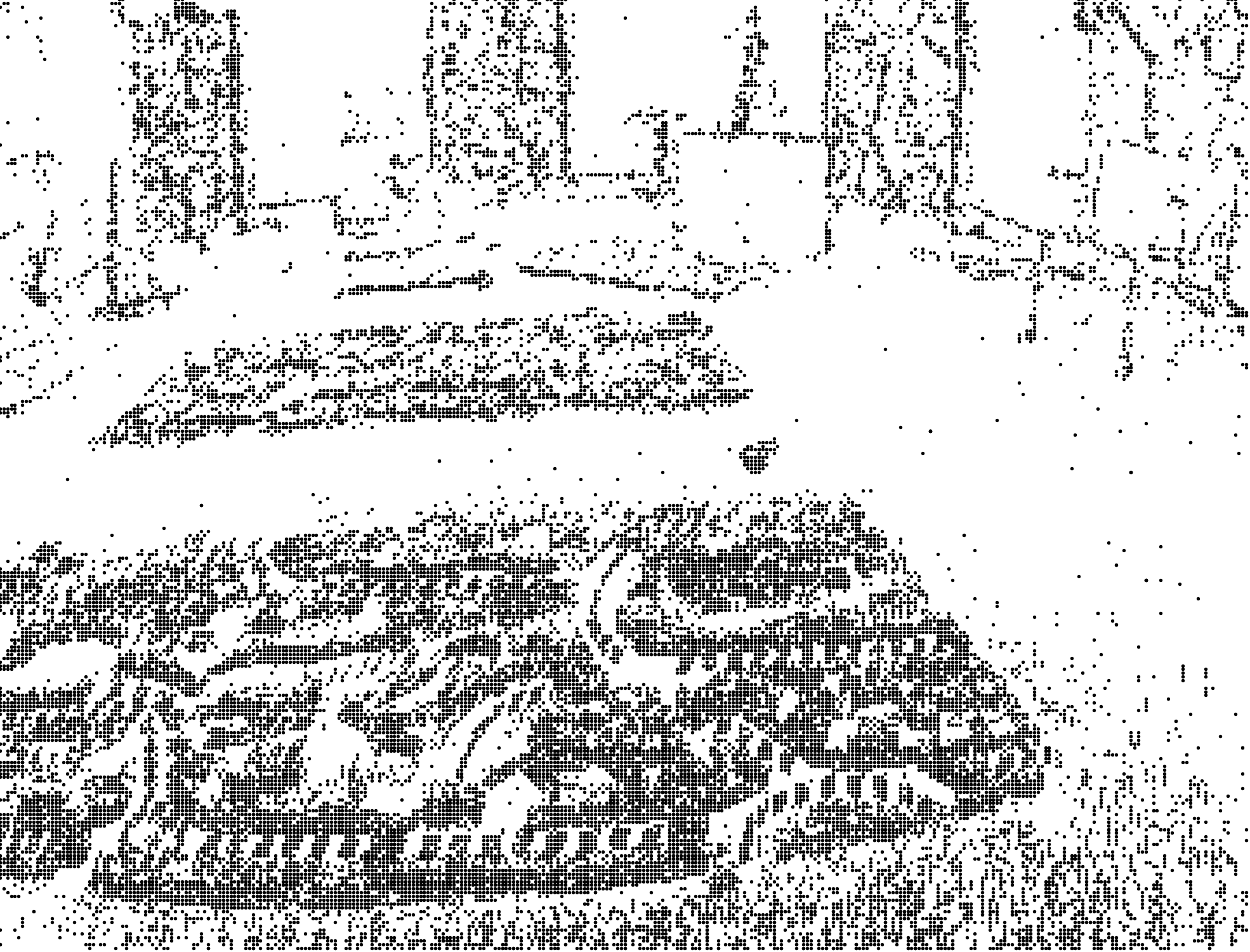} \end{tabular} &       % Original events
        \begin{tabular}{@{}c@{}} \includegraphics[width=0.2\textwidth, cfbox=gray 0.1pt 0pt]{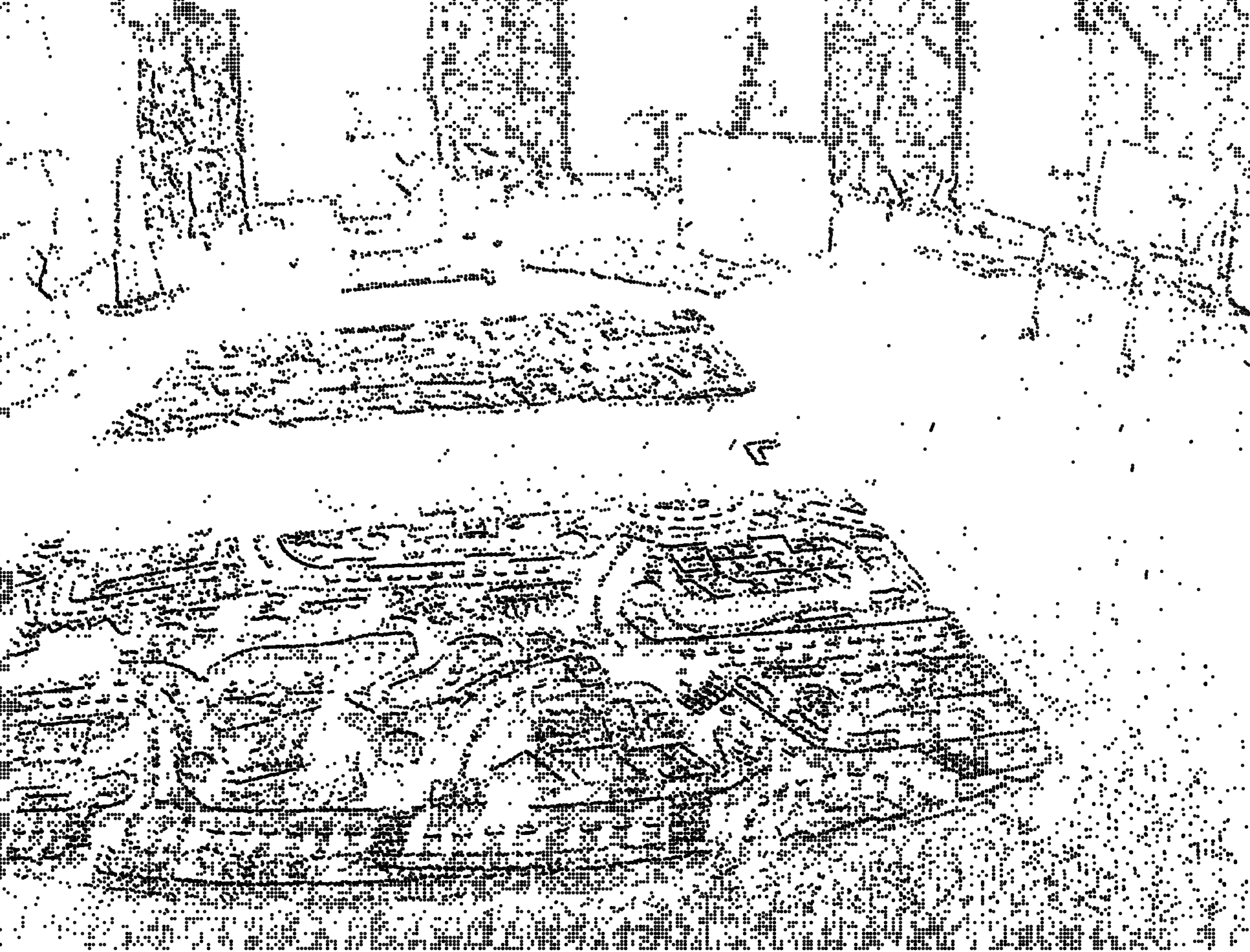} \end{tabular} &    % GT IWE
        \begin{tabular}{@{}c@{}} \includegraphics[width=0.2\textwidth, cfbox=gray 0.1pt 0pt]{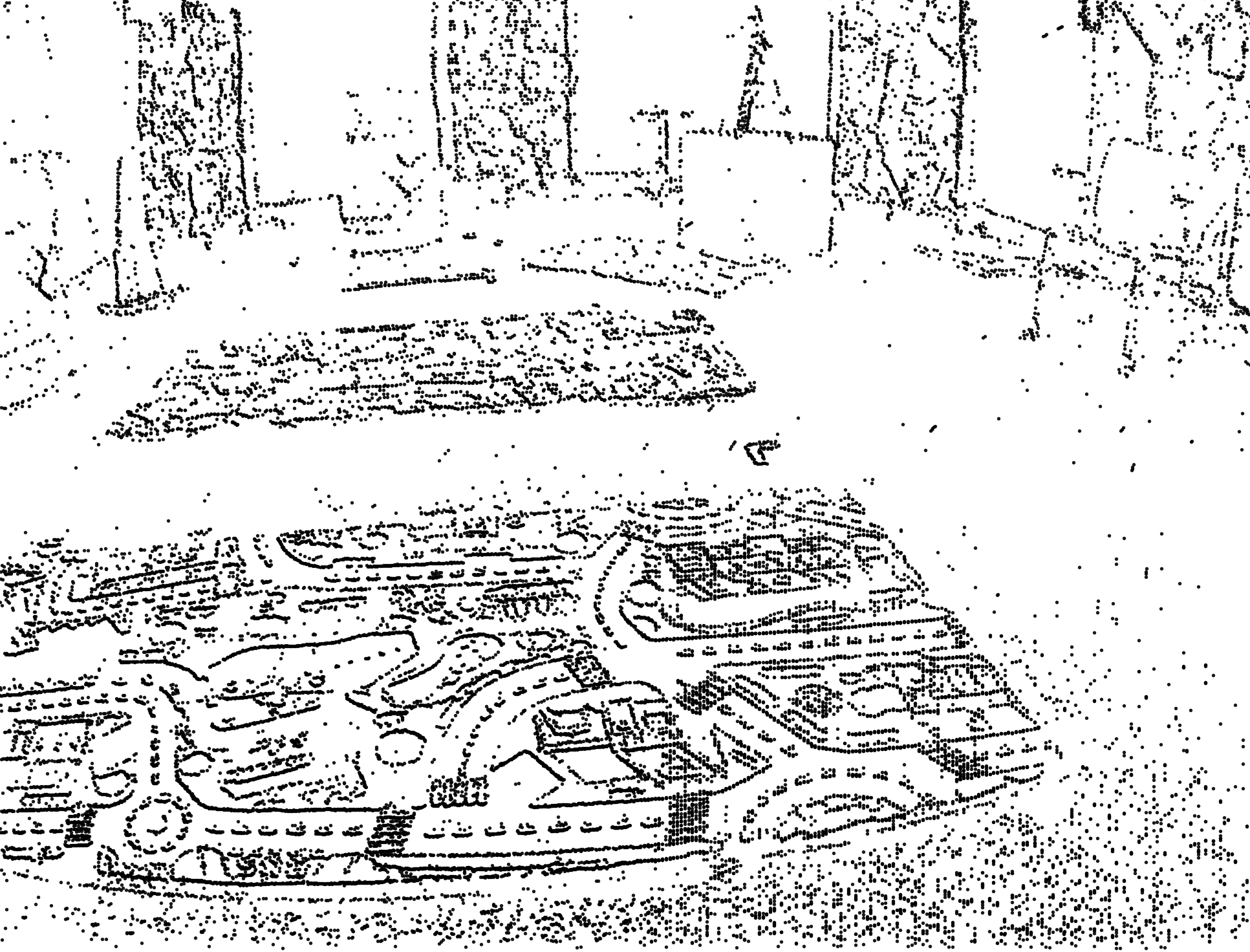} \end{tabular} & % Ours IWE
        \begin{tabular}{@{}c@{}} \includegraphics[width=0.2\textwidth, cfbox=gray 0.1pt 0pt]{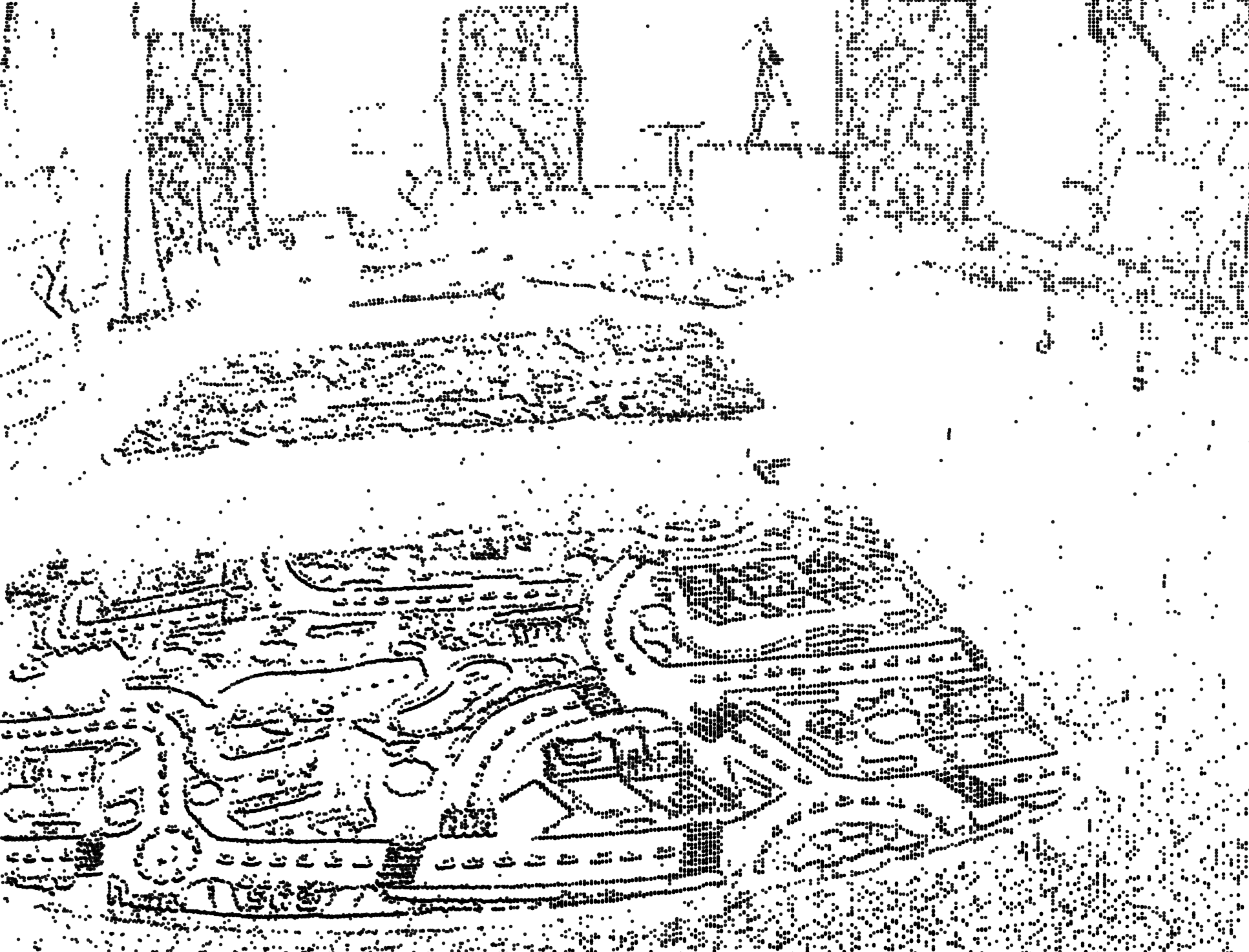} \end{tabular} &                         % MultiCM IWE
        \begin{tabular}{@{}c@{}} \includegraphics[width=0.152\textwidth, cfbox=gray 0.1pt 0pt]{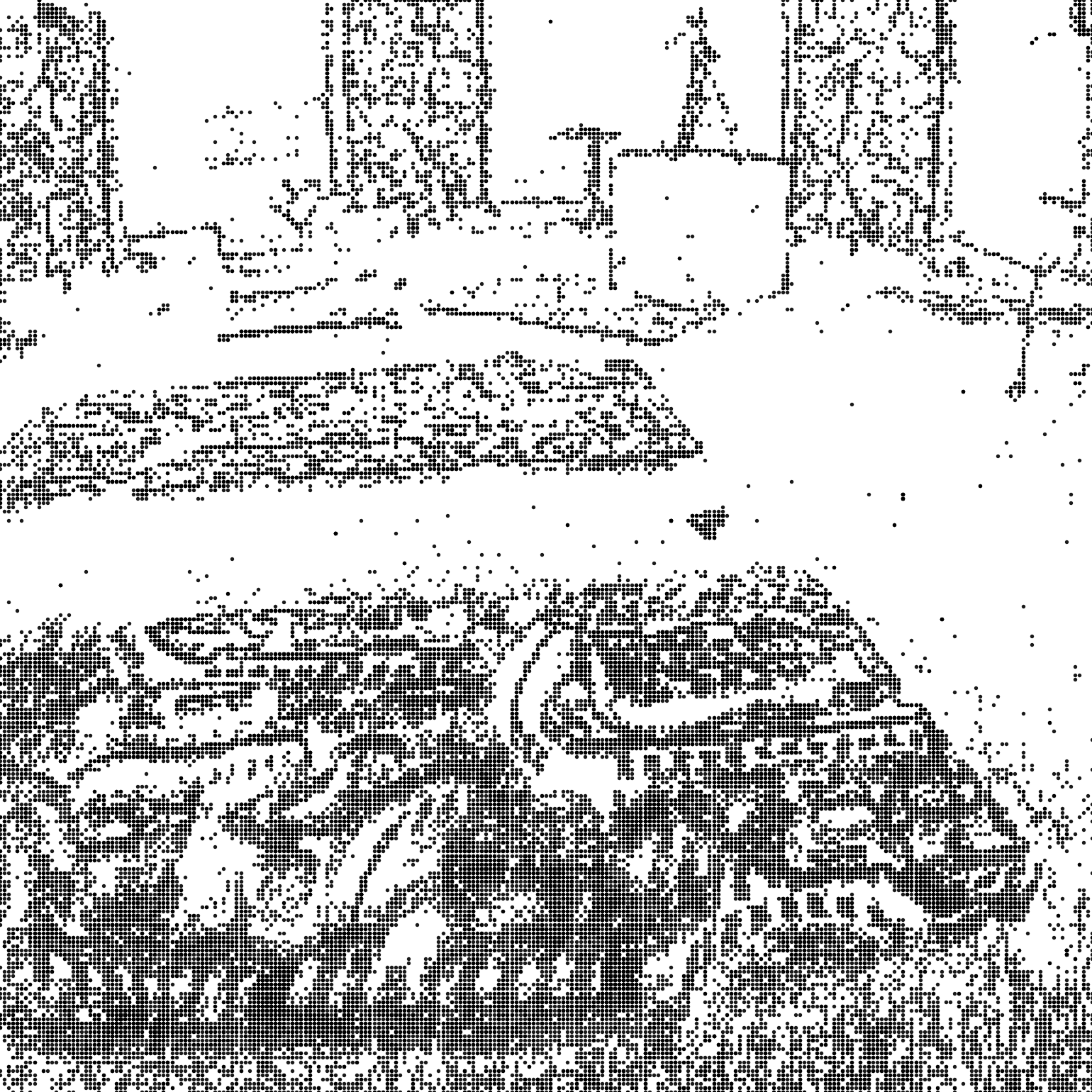} \end{tabular} \\ \vspace{-2pt}          % EV-FlowNet IWE

        % --------------------------------------------------------------------------------
        % outdoor_day1
        \multirow{2}{*}{\rotatebox[origin=c]{90}{\begin{adjustbox}{max width=\labelscaler\textwidth} \texttt{outdoor\_day1} \end{adjustbox}}} &
        \begin{tabular}{@{}c@{}} \includegraphics[width=0.2\textwidth, cfbox=gray 0.1pt 0pt]{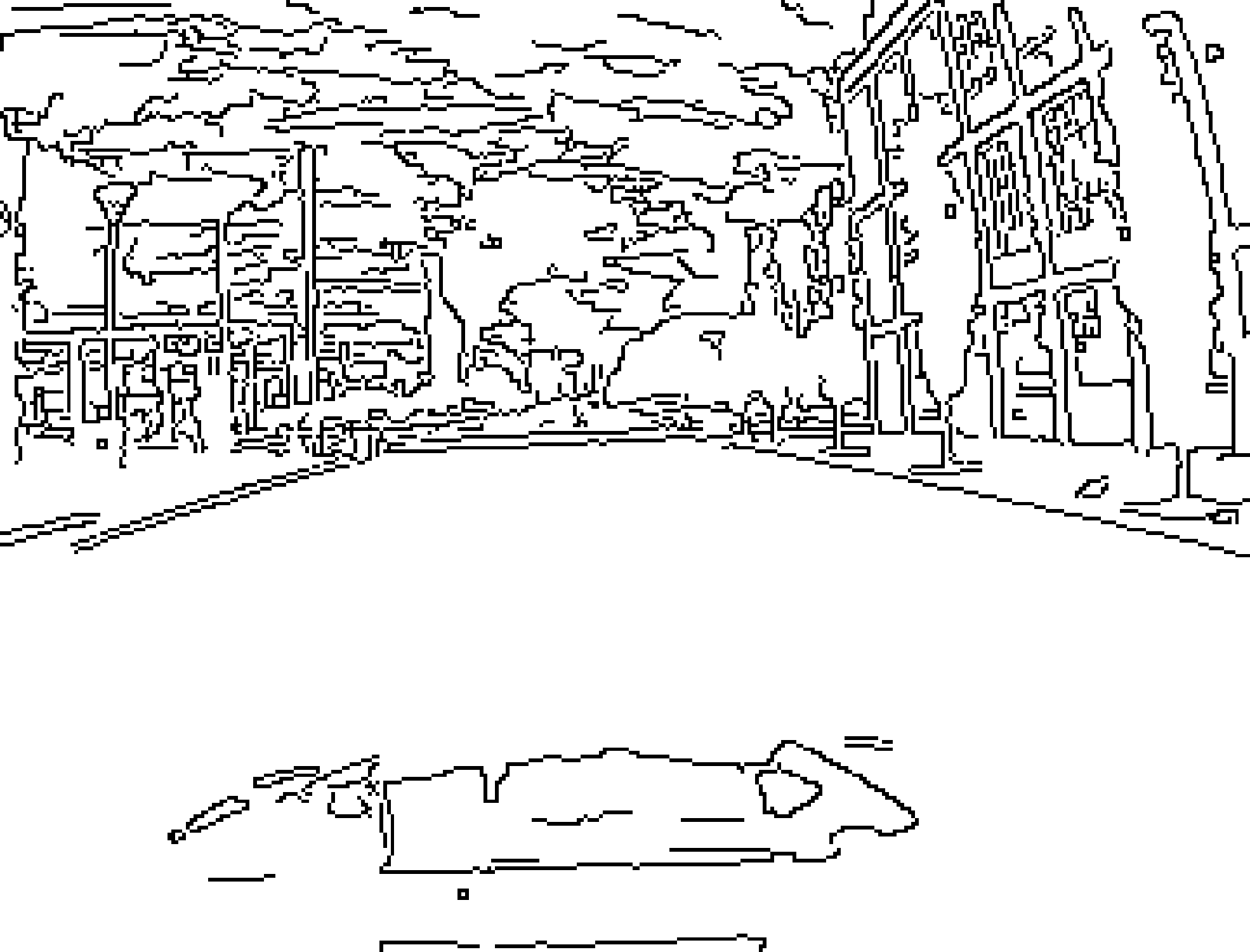} \end{tabular} &      % Edges 
        \begin{tabular}{@{}c@{}} \includegraphics[width=0.2\textwidth, cfbox=gray 0.1pt 0pt]{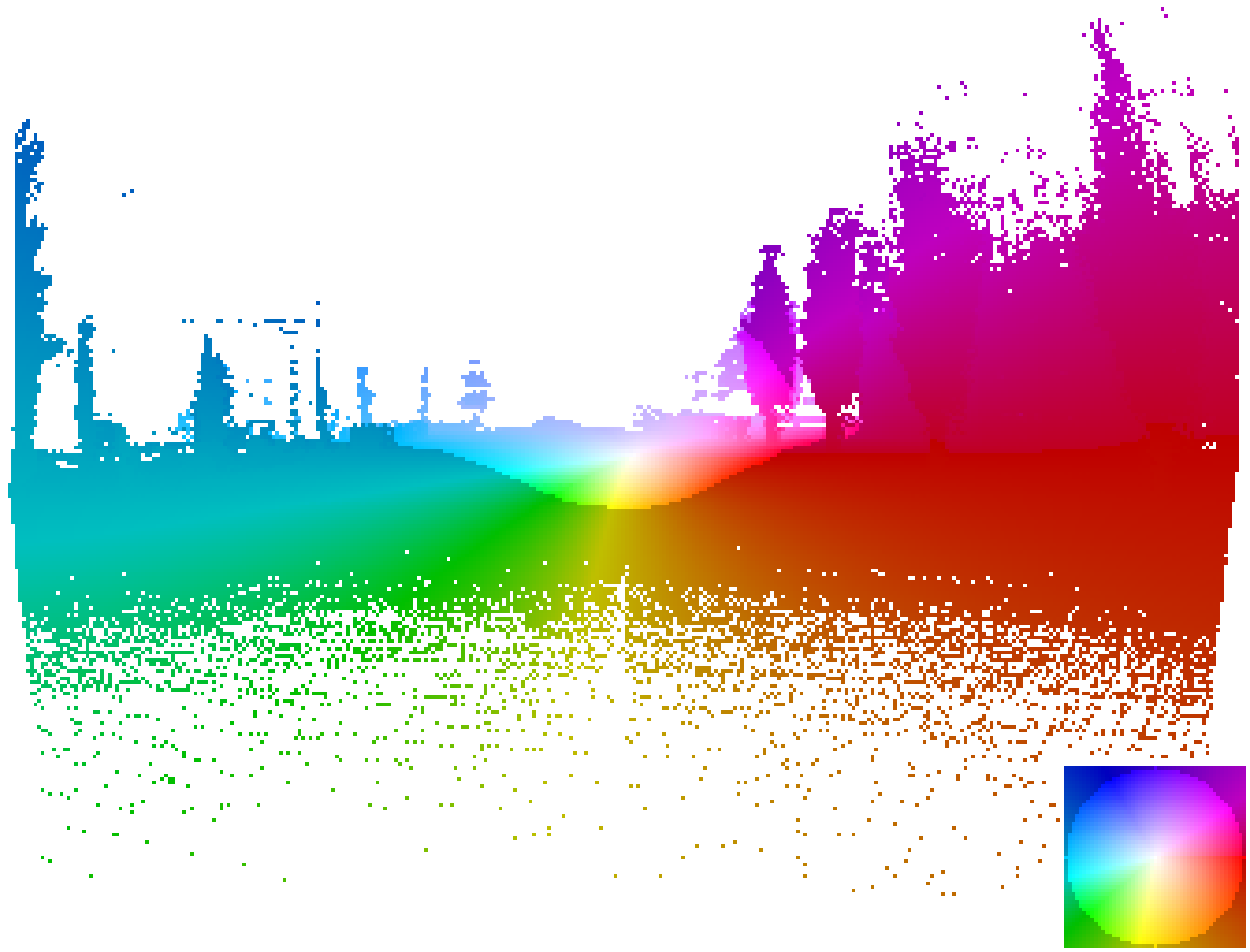} \end{tabular} &        % GT
        \begin{tabular}{@{}c@{}} \includegraphics[width=0.2\textwidth, cfbox=gray 0.1pt 0pt]{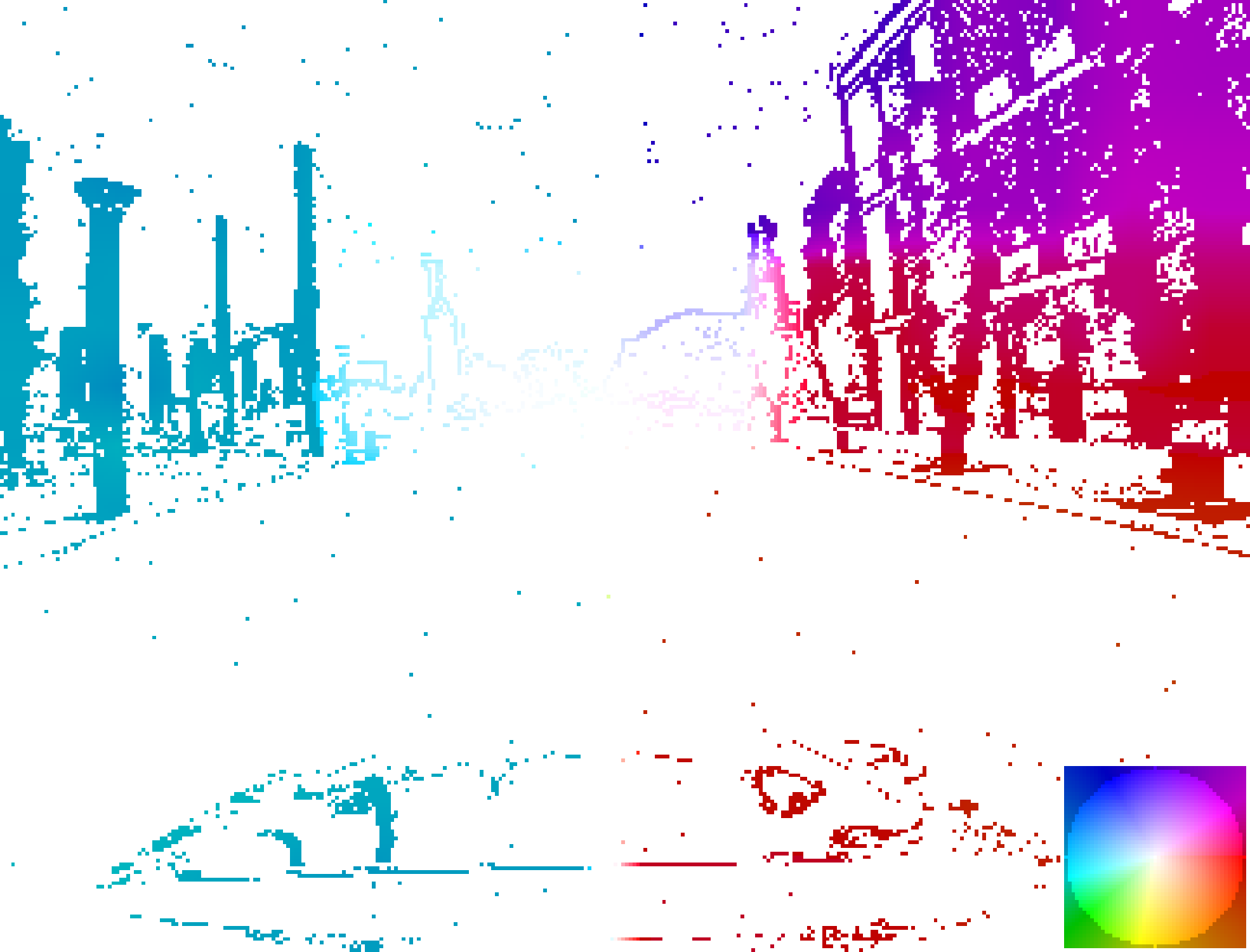} \end{tabular} &  % Ours (predicted flow masked by input events)
        \begin{tabular}{@{}c@{}} \includegraphics[width=0.203\textwidth, cfbox=gray 0.1pt 0pt]{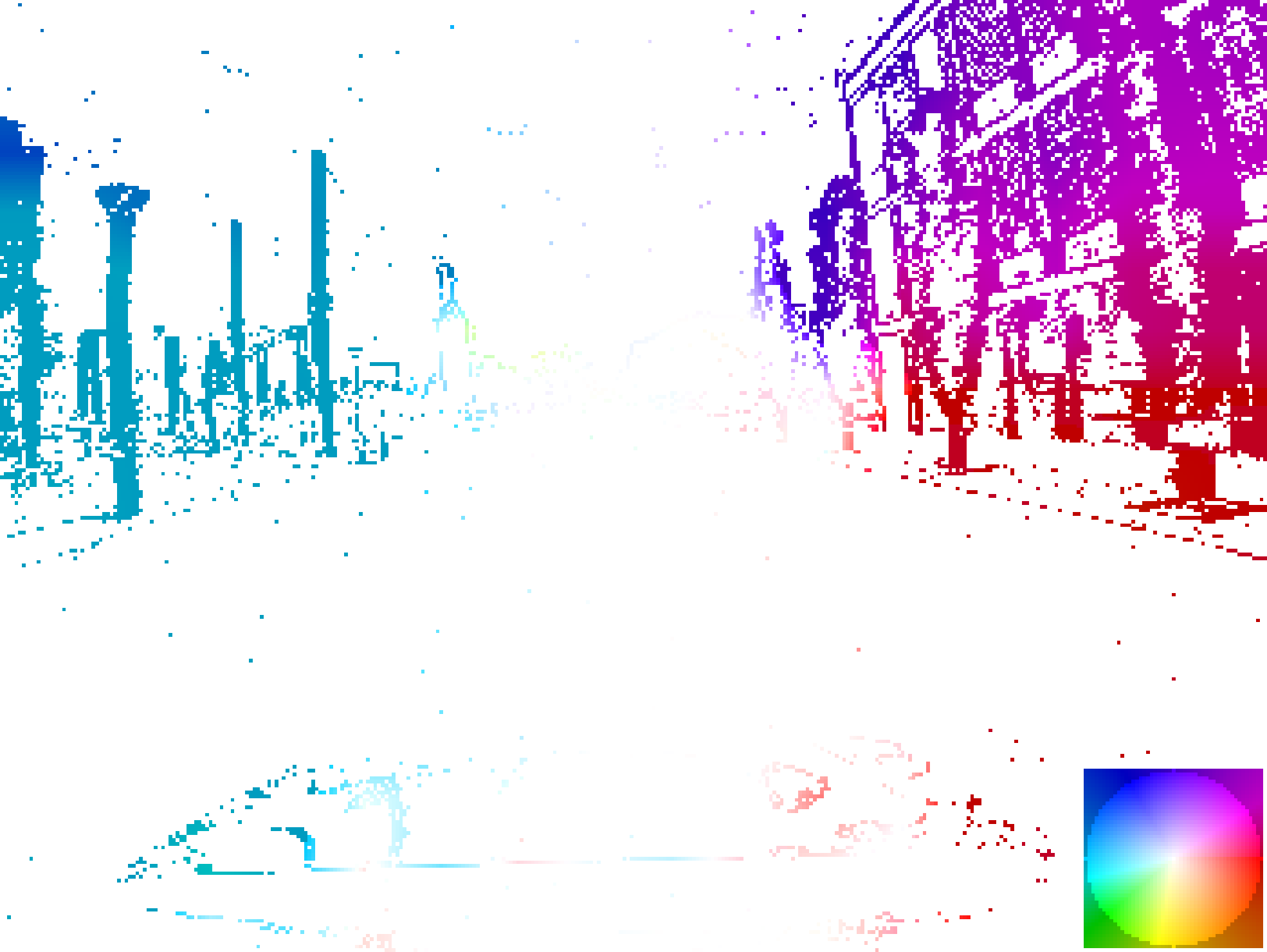} \end{tabular} &                         % MultiCM (predicted flow masked by input events)
        \begin{tabular}{@{}c@{}} \includegraphics[width=0.152\textwidth, cfbox=gray 0.1pt 0pt]{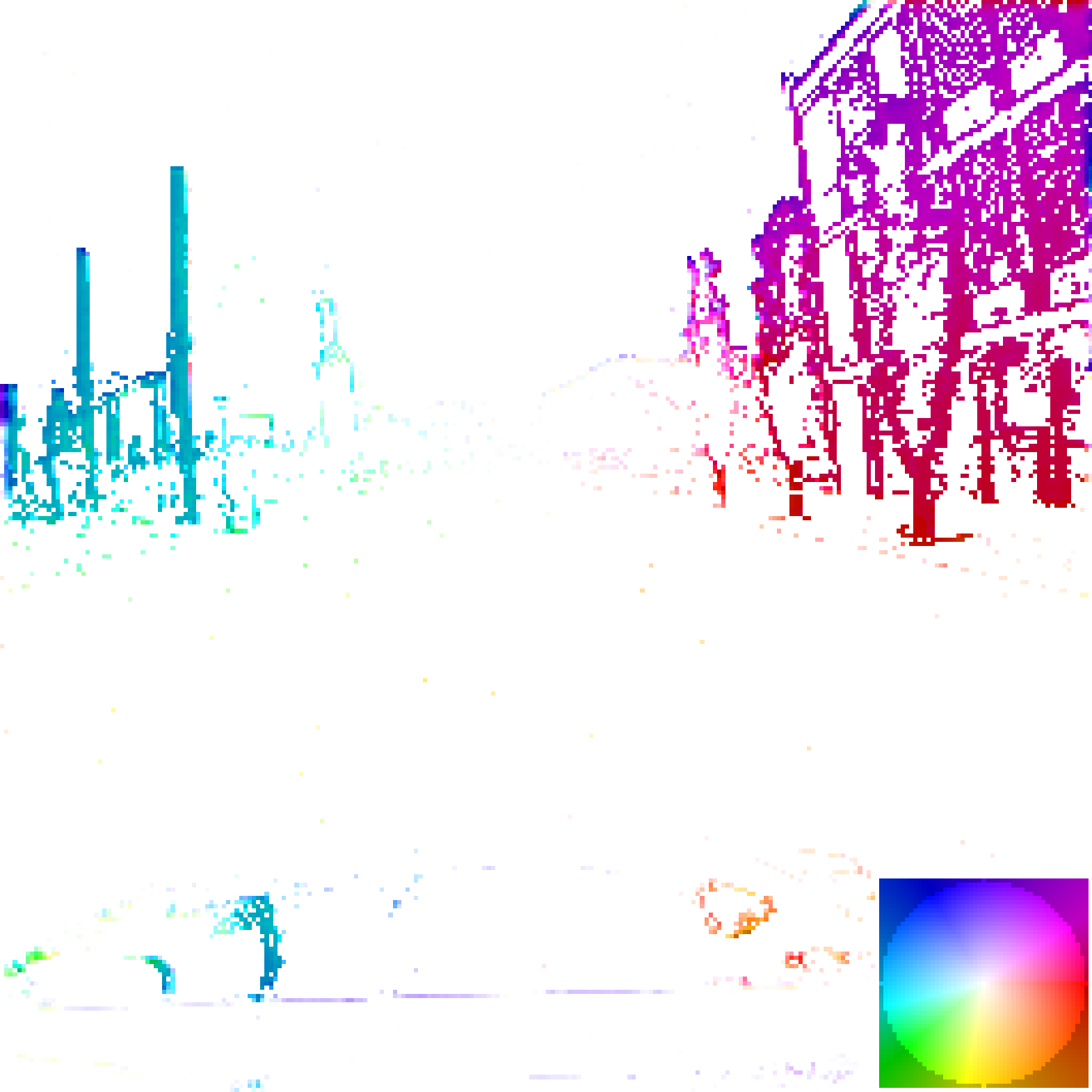} \end{tabular} \\                        % EV-FlowNet (predicted flow masked by input events)

        &
        \begin{tabular}{@{}c@{}} \includegraphics[width=0.2\textwidth, cfbox=gray 0.1pt 0pt]{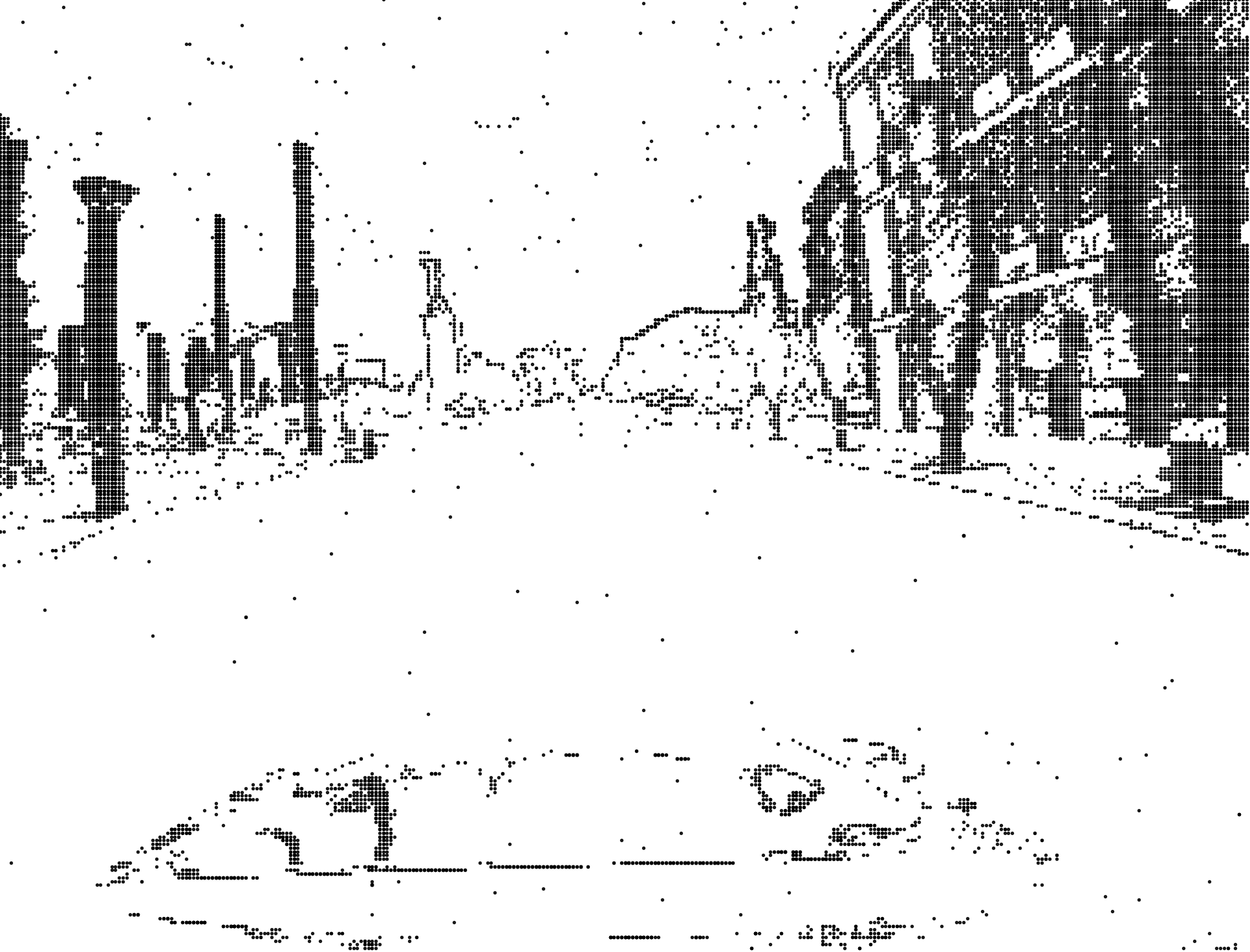} \end{tabular} &       % Original events
        \begin{tabular}{@{}c@{}} \includegraphics[width=0.2\textwidth, cfbox=gray 0.1pt 0pt]{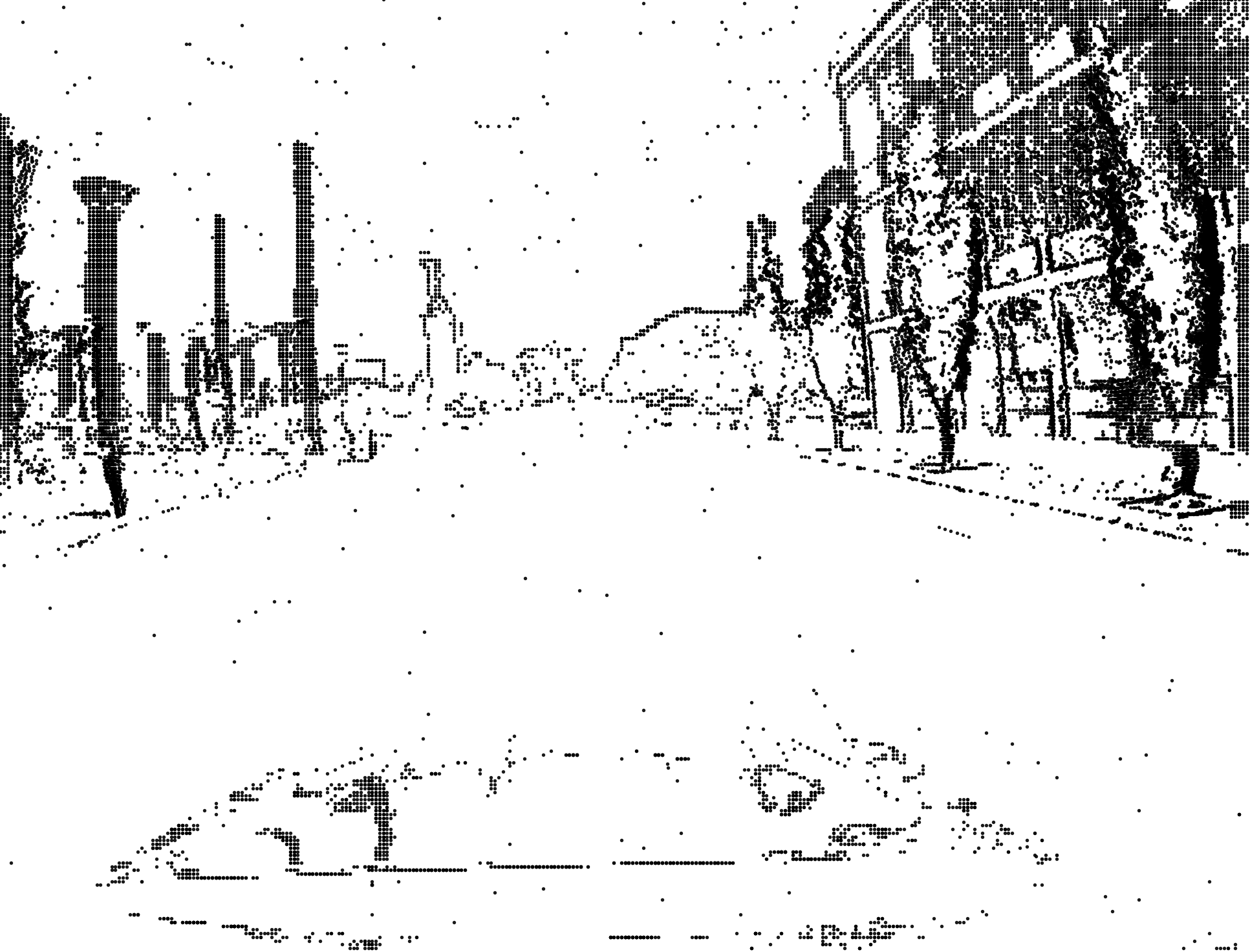} \end{tabular} &    % GT IWE
        \begin{tabular}{@{}c@{}} \includegraphics[width=0.2\textwidth, cfbox=gray 0.1pt 0pt]{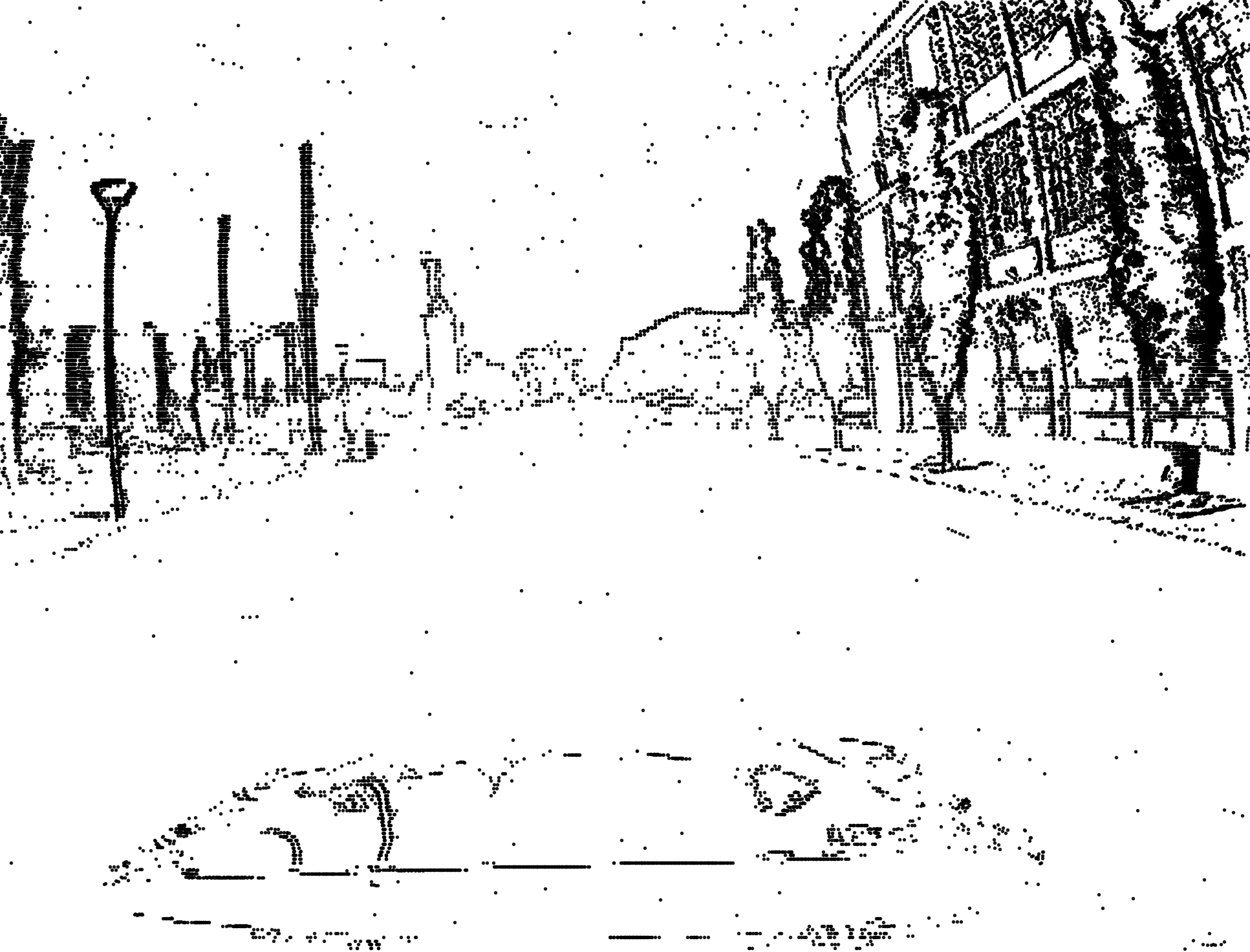} \end{tabular} & % Ours IWE
        \begin{tabular}{@{}c@{}} \includegraphics[width=0.2\textwidth, cfbox=gray 0.1pt 0pt]{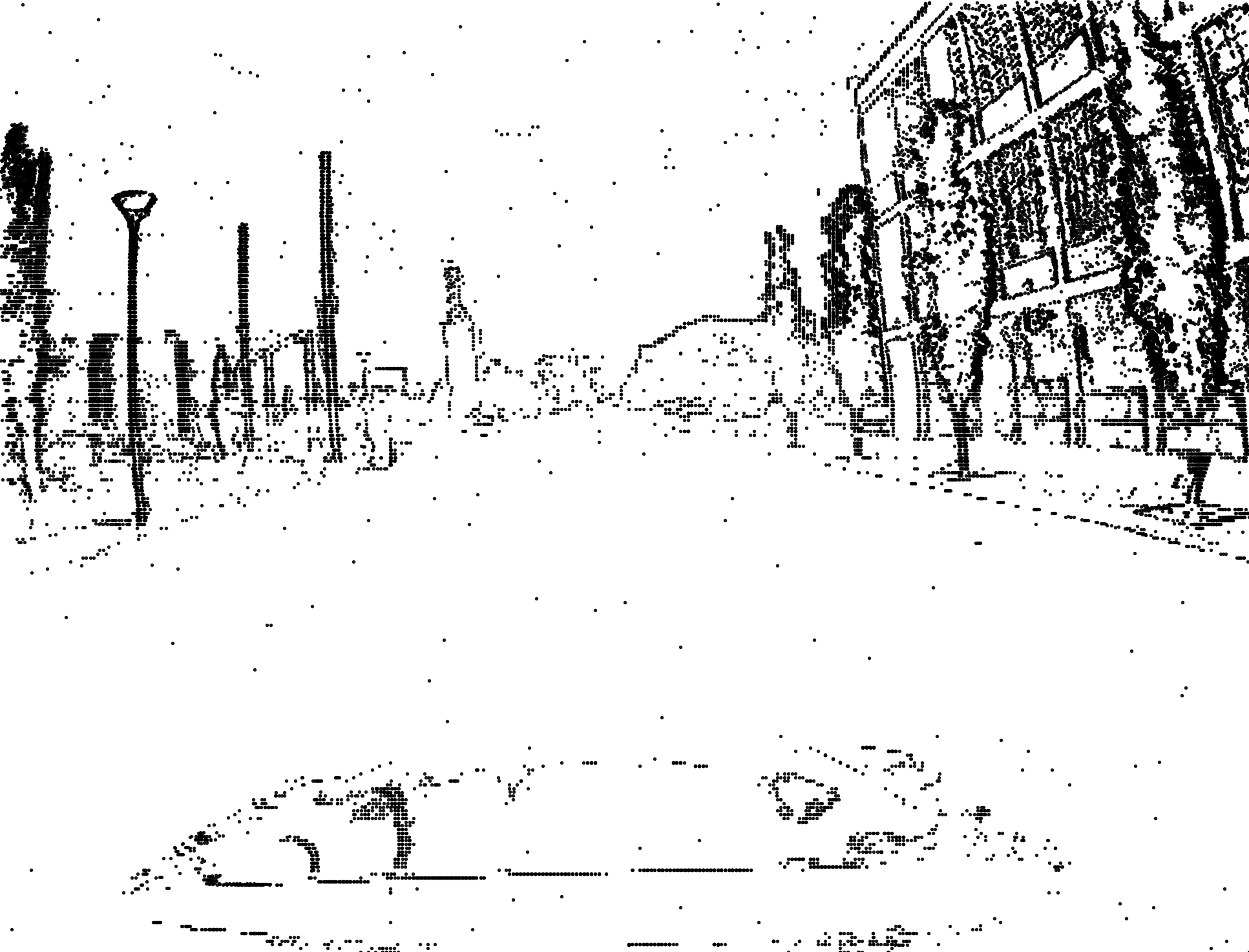} \end{tabular} &                         % MultiCM IWE
        \begin{tabular}{@{}c@{}} \includegraphics[width=0.152\textwidth, cfbox=gray 0.1pt 0pt]{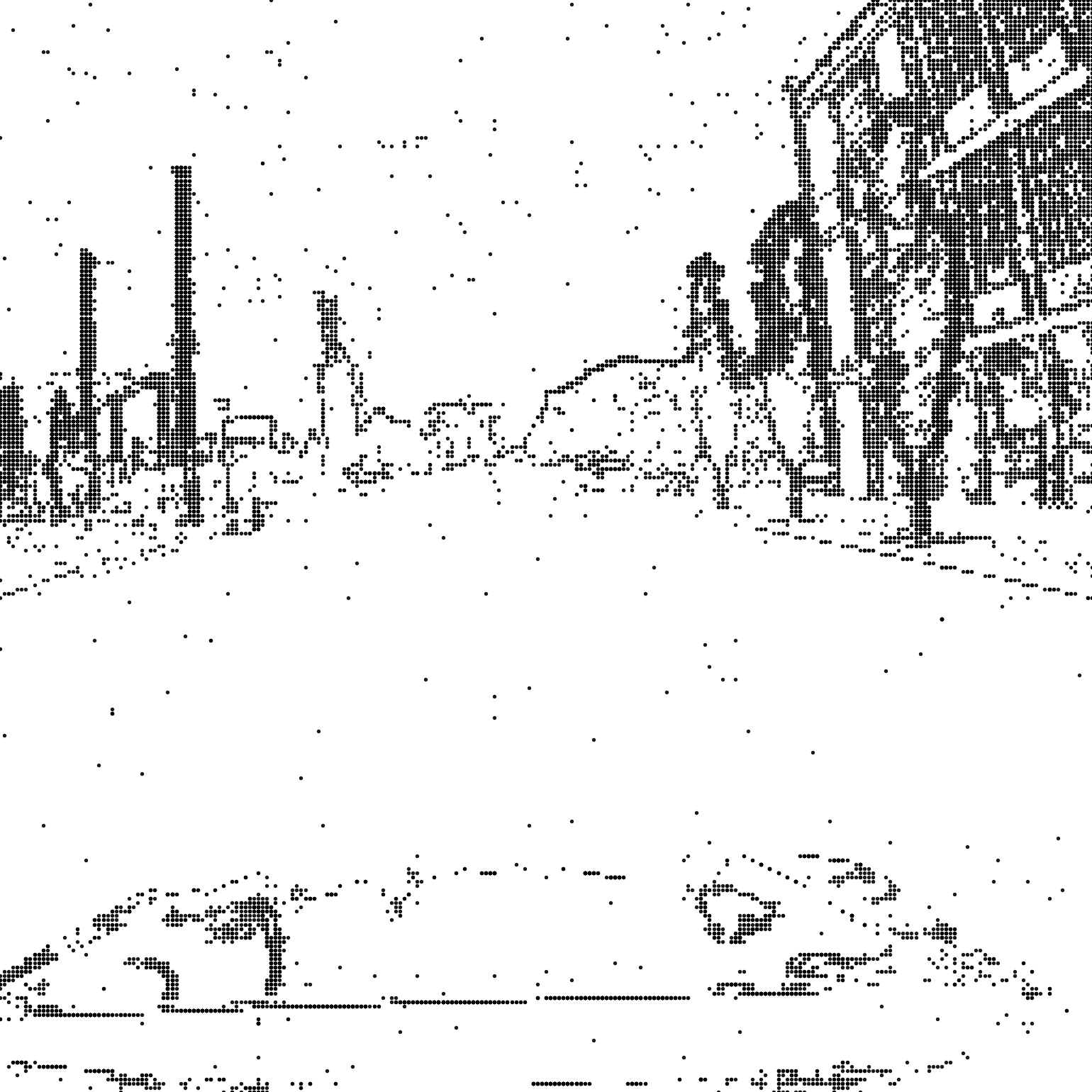} \end{tabular} \\                        % EV-FlowNet IWE

        % --------------------------------------------------------------------------------
        % last row
        &
        \begin{tabular}{@{}c@{}} \begin{adjustbox}{max width=\labelscaler\textwidth}(a) $E$, $I_{\text{events}}$       \end{adjustbox}\end{tabular} & 
        \begin{tabular}{@{}c@{}} \begin{adjustbox}{max width=\labelscaler\textwidth}(b) GT                             \end{adjustbox}\end{tabular} &
        \begin{tabular}{@{}c@{}} \begin{adjustbox}{max width=\labelscaler\textwidth}(c) Ours (MB)                      \end{adjustbox}\end{tabular} &
        \begin{tabular}{@{}c@{}} \begin{adjustbox}{max width=\labelscaler\textwidth}(d) \cite{shiba2022secrets} (MB)   \end{adjustbox}\end{tabular} &
        \begin{tabular}{@{}c@{}} \begin{adjustbox}{max width=\labelscaler\textwidth}(e) \cite{zhu2018evflownet} (SSL)  \end{adjustbox}\end{tabular} \\
    \end{tabular}
    \end{adjustbox}
    \caption{Qualitative comparisons ($dt=4$) of our approach against two
    prominent methods \cite{zhu2018evflownet,shiba2022secrets} on MVSEC. For
    each sequence, the two subsequent rows highlight the results. Column (a)
    shows our preprocessed edge images and the images of (original) events.
    Column (b) displays the available ground-truth (GT) flows and the
    corresponding IWEs. Columns (c-e) display the predicted flows masked by the
    original events and the constructed IWEs for each method.}
    \label{fig:mvsec_comparisons}
\end{figure*}

%%%%%%%%%%%%%%%%%%%%%%%%%%%%%%%%%%%%%%%%%%%%%%%%%%%%%%%%%%%%%%%%%%%%%%%%%%%%%%%%

\begin{figure}
    \centering
    \begin{adjustbox}{max width=\columnwidth}
    \begin{tabular}{@{}c@{\thinspace}c@{\thinspace}c@{\thinspace}|@{\thinspace}c@{\thinspace}c@{}}
        \vspace{-2pt}
        % --------------------------------------------------------------------------------
        % mvsec if2 and dsec thun_01_b
        
        \begin{tabular}{@{}c@{}} \includegraphics[width=0.27\columnwidth, cfbox=gray 0.1pt 0pt]{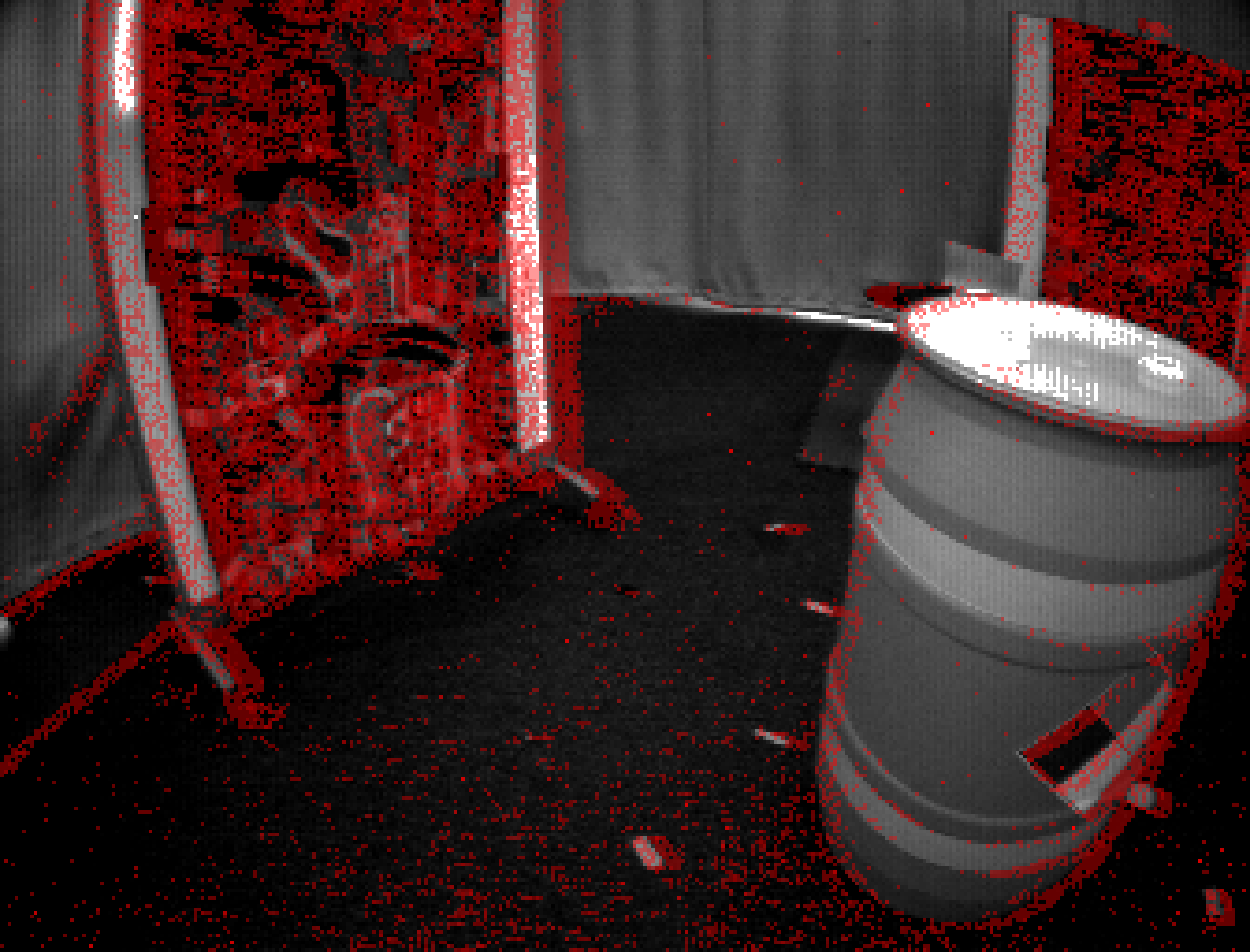} \end{tabular} &     
        \begin{tabular}{@{}c@{}} \includegraphics[width=0.27\columnwidth, cfbox=gray 0.1pt 0pt]{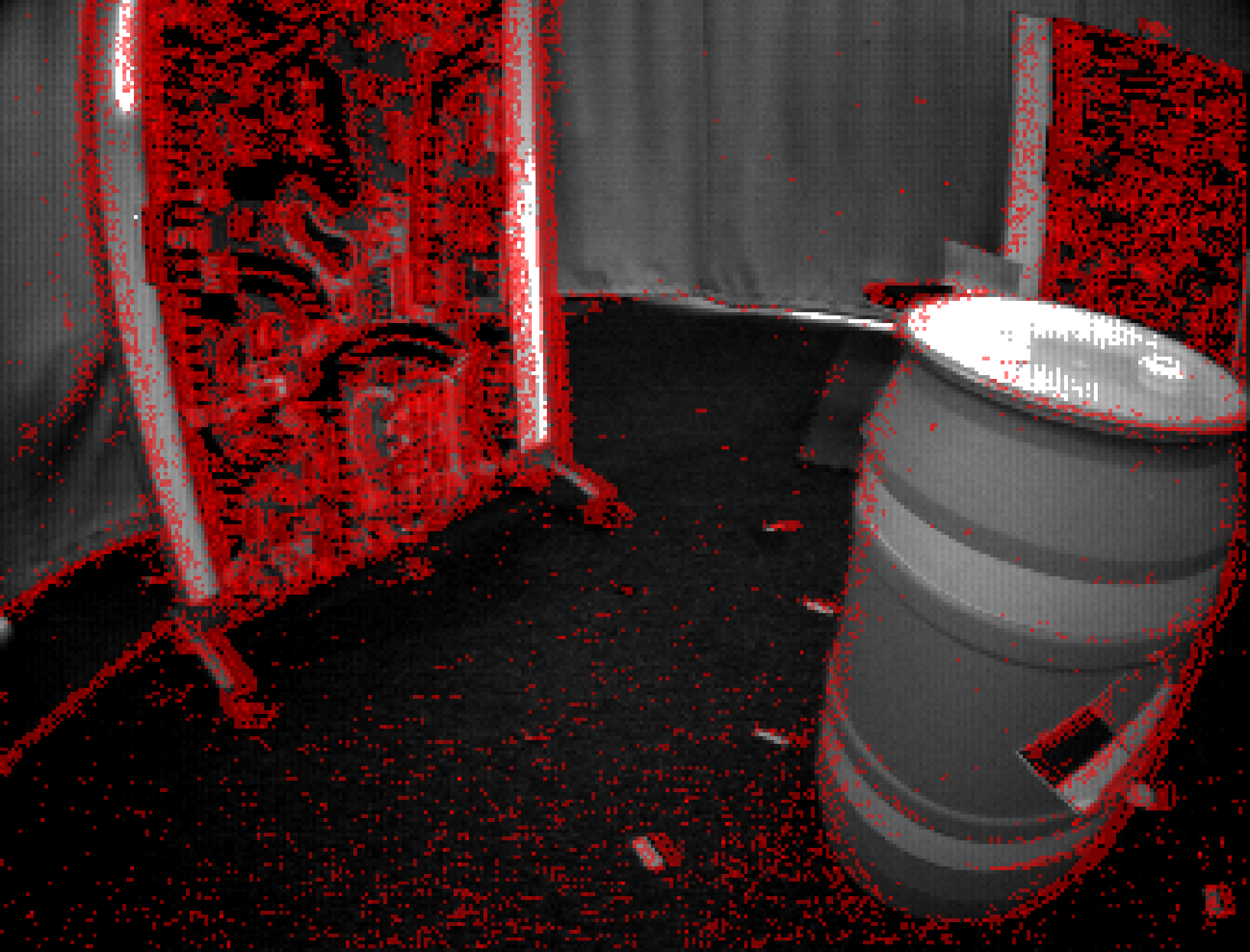} \end{tabular} &       
        \begin{tabular}{@{}c@{}} \includegraphics[width=0.27\columnwidth, cfbox=gray 0.1pt 0pt]{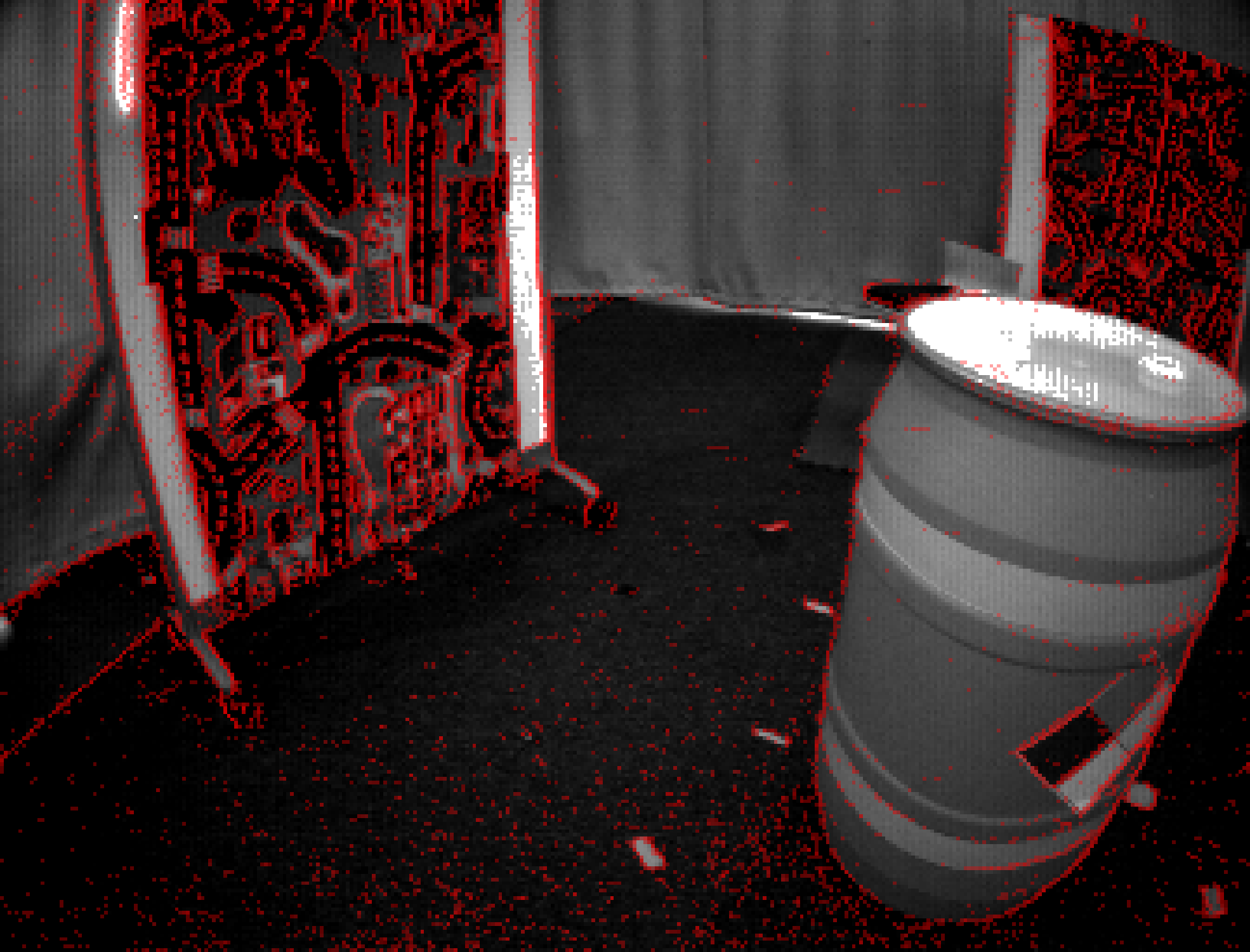}\end{tabular} &
        \begin{tabular}{@{}c@{}} \includegraphics[width=0.27\columnwidth, cfbox=gray 0.1pt 0pt]{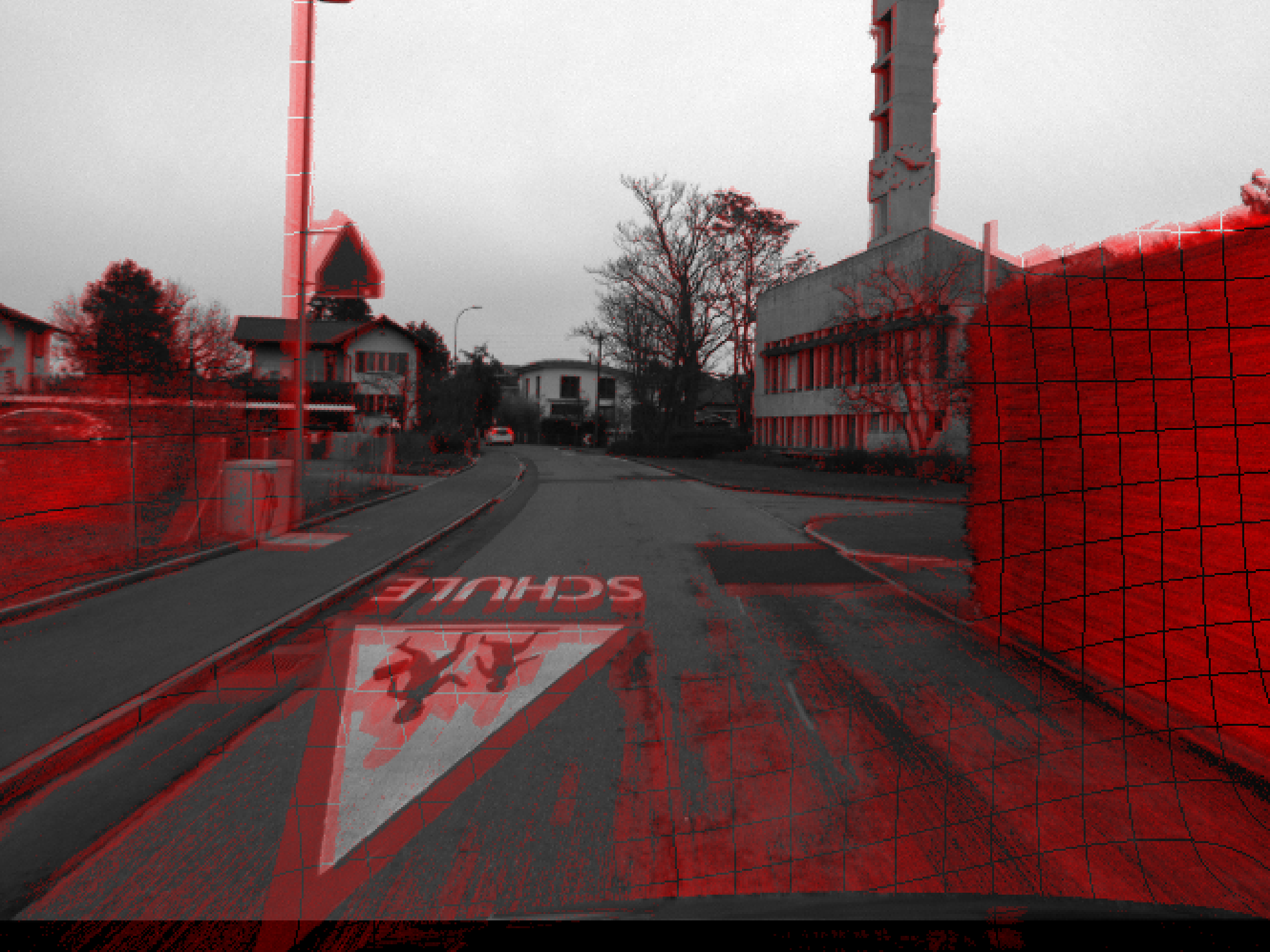} \end{tabular} &
        \begin{tabular}{@{}c@{}} \includegraphics[width=0.27\columnwidth, cfbox=gray 0.1pt 0pt]{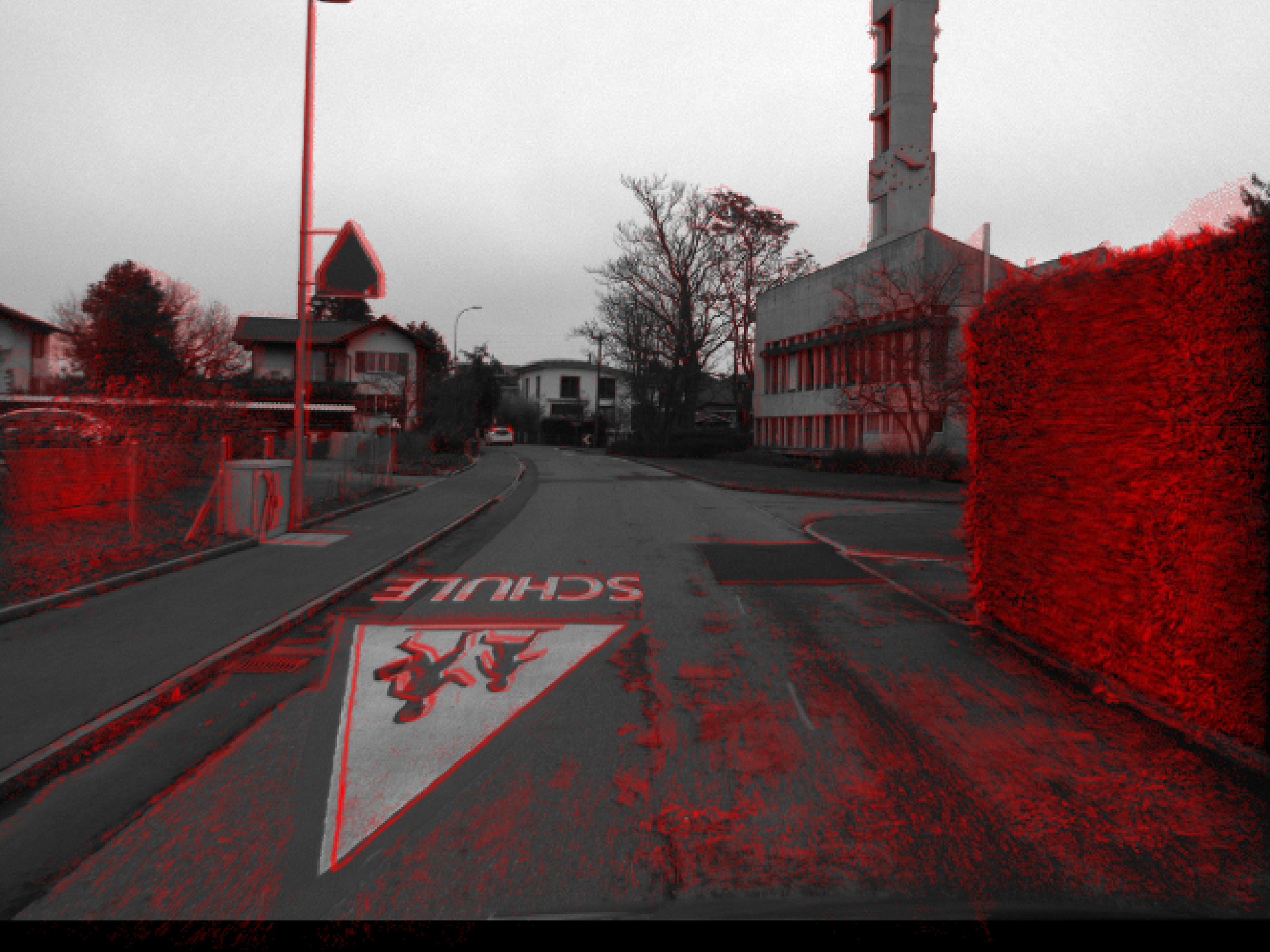} \end{tabular} \\

        % --------------------------------------------------------------------------------
        % last row        
        \begin{tabular}{@{}c@{}} \begin{adjustbox}{max width=0.27\columnwidth}(a) $I_{\text{events}}$  \end{adjustbox}\end{tabular} & 
        \begin{tabular}{@{}c@{}} \begin{adjustbox}{max width=0.27\columnwidth}(b)  IWE (GT) \end{adjustbox}\end{tabular} &
        \begin{tabular}{@{}c@{}} \begin{adjustbox}{max width=0.27\columnwidth}(c) IWE (Ours)  \end{adjustbox}\end{tabular} &
        \begin{tabular}{@{}c@{}} \begin{adjustbox}{max width=0.27\columnwidth}(d) $I_{\text{events}}$  \end{adjustbox}\end{tabular} &
        \begin{tabular}{@{}c@{}} \begin{adjustbox}{max width=0.27\columnwidth}(e) IWE (Ours)  \end{adjustbox}\end{tabular} \\
    \end{tabular}
    
    \begin{tikzpicture}[overlay, remember picture, baseline=0pt,]
        \draw [white!50!cyan,thick,opacity=1] ($(-9.36,1.19)+(1.4,-0.5)$) rectangle ++(0.4,-1.1);
        \draw [white!50!cyan,thick,opacity=1] ($(-7.05,1.19)+(1.4,-0.5)$) rectangle ++(0.4,-1.1);
        
        \draw [lightgray!25!lime,thick,opacity=1] ($(-4.66,1.19)+(1.69,-0.39)$) rectangle ++(0.54,-1.2);
        
        \draw [lightgray!25!yellow,thick,opacity=1] ($(-4.66,1.19)+(0.5,-1.1)$) rectangle ++(0.25,-0.5);
        \draw [lightgray!25!yellow,thick,opacity=1] ($(-2.35,1.19)+(0.5,-1.1)$) rectangle ++(0.25,-0.5);
    \end{tikzpicture}
    
    \end{adjustbox}
    
    \caption{MVSEC and DSEC ground-truth (GT) diagnosis. Events are overlaid
    over corresponding image frames. (a-c) shows the original events, GT warped
    events, and our warped events, respectively, on the MVSEC sequence
    \texttt{indoor\_flying\_2}. Compared to the GT, our method yields sharper
    warped events (\protect\tikz[baseline] \protect\draw[color=white!50!cyan,
    thick, opacity=1] (0ex,0ex) rectangle (1.4ex,1.4ex);) that display better
    alignment with the image edges (also refer to \cref{fig:mvsec_comparisons}
    (c) rows 3 and 4). (d) and (e) show the original and our warped events,
    respectively, on the DSEC sequence \texttt{thun\_01\_b}, which was captured
    using different sensors. Note the grid-like rectification artifacts
    (\protect\tikz[baseline] \protect\draw[color=lightgray!25!lime, thick,
    opacity=1] (0ex,0ex) rectangle (1.4ex,1.4ex);) in (d). Also, observe in (e)
    that the warped events are sharp, however the image alignment is limited.
    Misalignment artifacts from imperfect frame registration become prominent
    (\eg, road markings \protect\tikz[baseline]
    \protect\draw[color=lightgray!25!yellow, thick, opacity=1] (0ex,0ex)
    rectangle (1.4ex,1.4ex);) at points near the camera (see
    \cref{fig:dsec_ecd_evaluations} row 3 for further reference). These
    artifacts can render the problem ill-posed. In such scenarios, assigning a
    higher value to the coefficient of the correlation objective, $\beta$, may
    hinder overall convergence.}
    \label{fig:gt_challenging}
\end{figure}

Our method, EINCM, significantly improves upon the baseline (especially in the
$\emph{dt} = 4$ setting) and establishes a new benchmark among MB methods.
Furthermore, our results are comparable to learning-based techniques, minus the
need for ground-truth optical flow for training. In terms of FWL
(\cref{tab:fwl_scores}), which indicates the sharpness of the constructed IWEs,
EINCM achieves superior results compared to the previous state of the art
\cite{shiba2022secrets}. \cref{fig:mvsec_comparisons} shows qualitative results
against other methods that do not require ground-truth optical flow. Note that
all three methods provide dense flow for every pixel and we only visualize the
flows at pixels where events exist. Visually, EV-FlowNet produces the worst
results in terms of flow consistency (with respect to ground-truth flow and
neighboring objects) and the sharpness of the constructed IWEs. Compared to
MultiCM, our method provides more consistent flow predictions, especially for
objects that are farther away from the camera (\eg, rows 1 and 5 in
\cref{fig:mvsec_comparisons}). Moreover, our IWEs also contain fewer artifacts
(\eg, rows 4, 6, 8 in \cref{fig:mvsec_comparisons}) and achieve better
alignment with the image edges (\cref{fig:gt_challenging}).

%%%%%%%%%%%%%%%%%%%%%%%%%%%%%%%%%%%%%%%%%%%%%%%%%%%%%%%%%%%%%%%%%%%%%%%%%%%%%%%%
% Table: MVSEC, ECD, DSEC FWLS
\begin{table}
    \centering
    \begin{adjustbox}{max width=\columnwidth}
        \begin{tabular}{@{\thinspace}l@{\thickspace}c@{\thickspace}
        c@{\thickspace}c@{\thickspace}c@{\thickspace}c@{\thickspace}c@{\thickspace}c@{\thickspace}c@{\thickspace}c@{\thickspace}
        c@{\thickspace}c@{\thickspace}
        c@{\thickspace}c@{\thickspace}c}
    
            % ================================================================================
            \toprule
            & \thickspace &
            \multicolumn{7}{c}{MVSEC ($dt=4$)}              & \thickspace &     % MVSEC
            \multicolumn{1}{c}{ECD}                       & \thickspace &     % ECD
            \multicolumn{3}{c}{DSEC (train)}              \\                  % DSEC (train)    
            % --------------------------------------------------------------------------------
            \cmidrule(){3-9} \cmidrule(){11-11} \cmidrule(){13-15} 

            & \thickspace & 
            \texttt{ind\_fly1}                      & \thickspace &     % indoor_flying1
            \texttt{ind\_fly2}                      & \thickspace &     % indoor_flying2
            \texttt{ind\_fly3}                      & \thickspace &     % indoor_flying2 
            \texttt{out\_day1}                      & \thickspace &     % outdoor_day1 
            \texttt{slider\_depth}                  & \thickspace &     % slider_depth
            \texttt{thu\_00\_a}                     & \thickspace &     % thun_00_a
            \texttt{zur\_07\_a}                     \\                  % zurich_city_07_a 
            % --------------------------------------------------------------------------------
            \midrule 
            
            Ground truth & \thickspace &
            1.09                                          & \thickspace &     % indoor_flying1
            1.20                                          & \thickspace &     % indoor_flying2
            1.12                                          & \thickspace &     % indoor_flying3 
            1.07                                          & \thickspace &     % outdoor_day1 
            --                                            & \thickspace &     % slider_depth
            1.01                                          & \thickspace &     % thun_00_a
            1.04                                          \\                  % zurich_city_07_a 
            Shiba \etal \cite{shiba2022secrets} & \thickspace &
            1.17                                          & \thickspace &     % indoor_flying1
            1.30                                          & \thickspace &     % indoor_flying2
            1.23                                          & \thickspace &     % indoor_flying3 
            1.11                                          & \thickspace &     % outdoor_day1 
            1.93                                          & \thickspace &     % slider_depth
            1.42                                          & \thickspace &     % thun_00_a
            1.63                                          \\                  % zurich_city_07_a 
            Ours (EINCM) & \thickspace &
            \textbf{1.33\tiny{3}}                         & \thickspace &     % indoor_flying1
            \textbf{1.45\tiny{5}}                         & \thickspace &     % indoor_flying2
            \textbf{1.39\tiny{2}}                         & \thickspace &     % indoor_flying3 
            \textbf{1.23\tiny{0}}                         & \thickspace &     % outdoor_day1 
            \textbf{1.97}                                 & \thickspace &     % slider_depth
            \textbf{1.53\tiny{2}}                         & \thickspace &     % thun_00_a
            \textbf{1.63\tiny{6}}                         \\                  % zurich_city_07_a 
            % ================================================================================
            \bottomrule
        \end{tabular}
    \end{adjustbox}
    \caption{Flow warp loss (FWL) on MVSEC ($dt=1$ in the supplementary
    material), ECD, and DSEC (train). \emph{Bold} indicates \textbf{best}.}
    \label{tab:fwl_scores}
\end{table}

%%%%%%%%%%%%%%%%%%%%%%%%%%%%%%%%%%%%%%%%%%%%%%%%%%%%%%%%%%%%%%%%%%%%%%%%%%%%%%%%

% ------------------------------------------------------------------------------
\subsection{DSEC Evaluation}
\label{subsec:dsec_evaluation}
\begin{figure}
    \centering
    \begin{adjustbox}{max width=\columnwidth}
    \begin{tabular}{@{}c@{\thinspace}c@{\thinspace}c@{\thinspace}|@{\thinspace}c@{\thinspace}c@{\thinspace}c@{}}
        \vspace{-2pt}
        % --------------------------------------------------------------------------------
        % thun_00_a
        
        \begin{tabular}{@{}c@{}} \includegraphics[width=0.27\columnwidth, cfbox=gray 0.1pt 0pt]{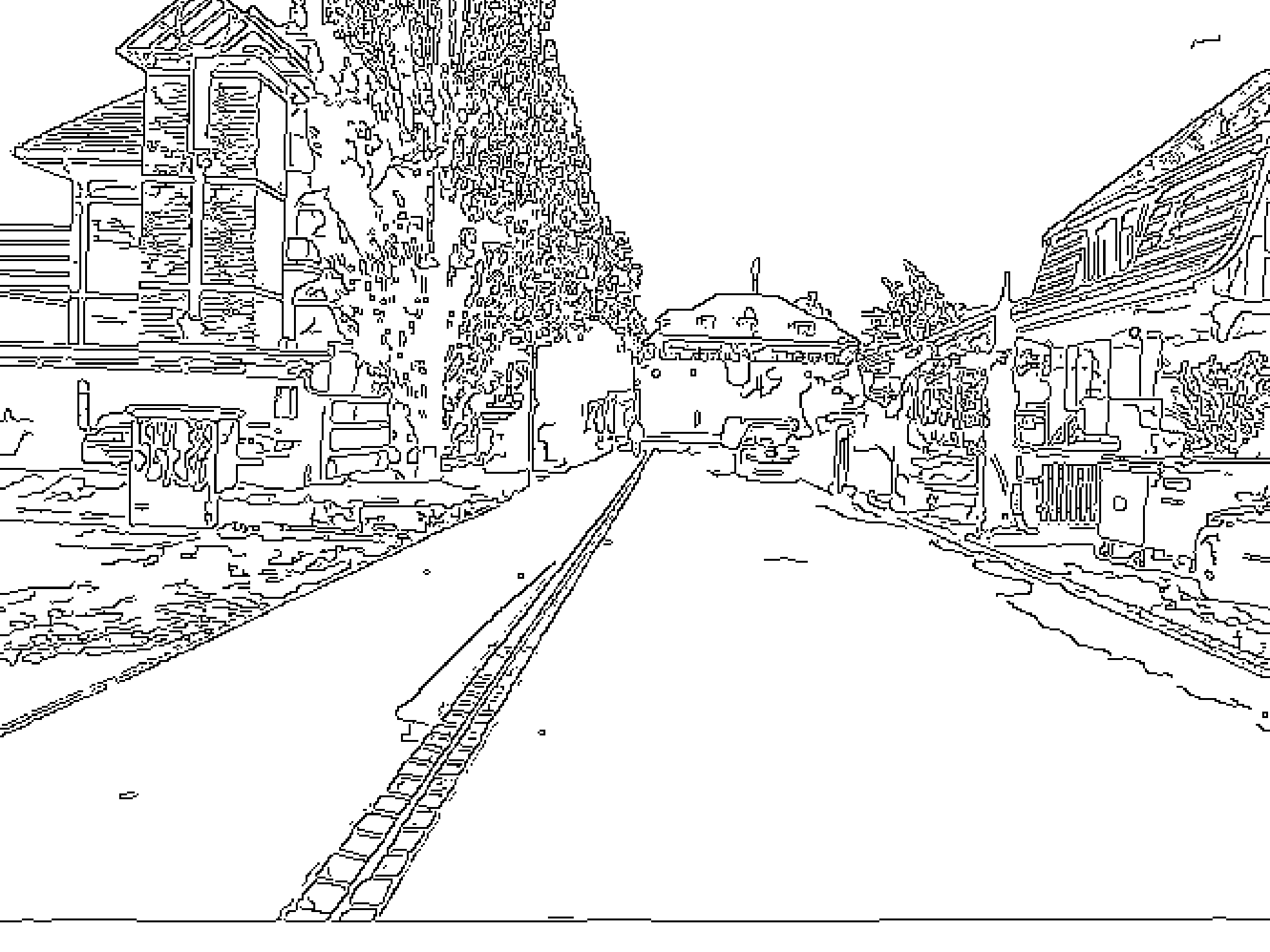} \end{tabular} &     
        \begin{tabular}{@{}c@{}} \includegraphics[width=0.27\columnwidth, cfbox=gray 0.1pt 0pt]{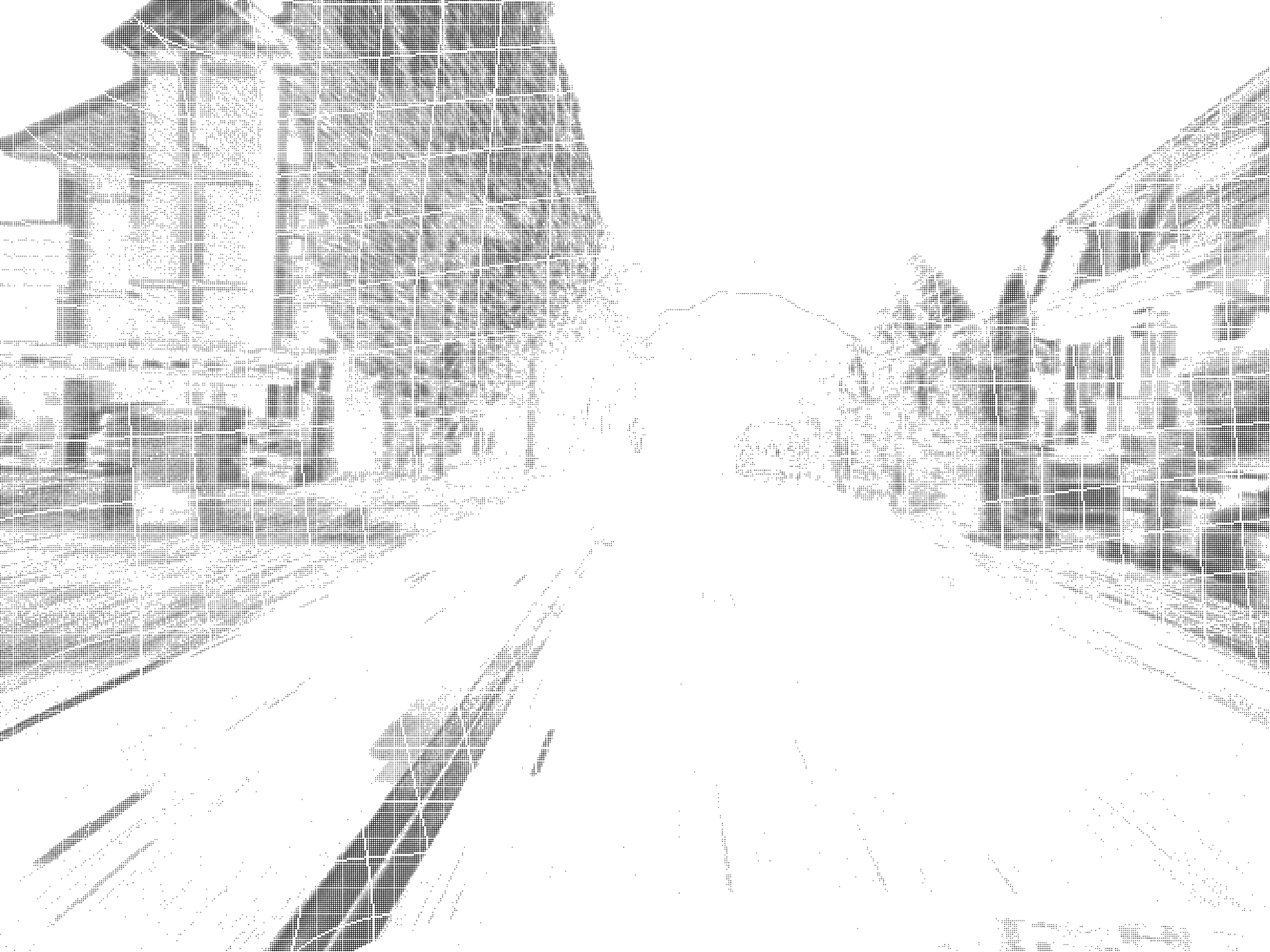} \end{tabular} &       
        \begin{tabular}{@{}c@{}} \includegraphics[width=0.27\columnwidth, cfbox=gray 0.1pt 0pt]{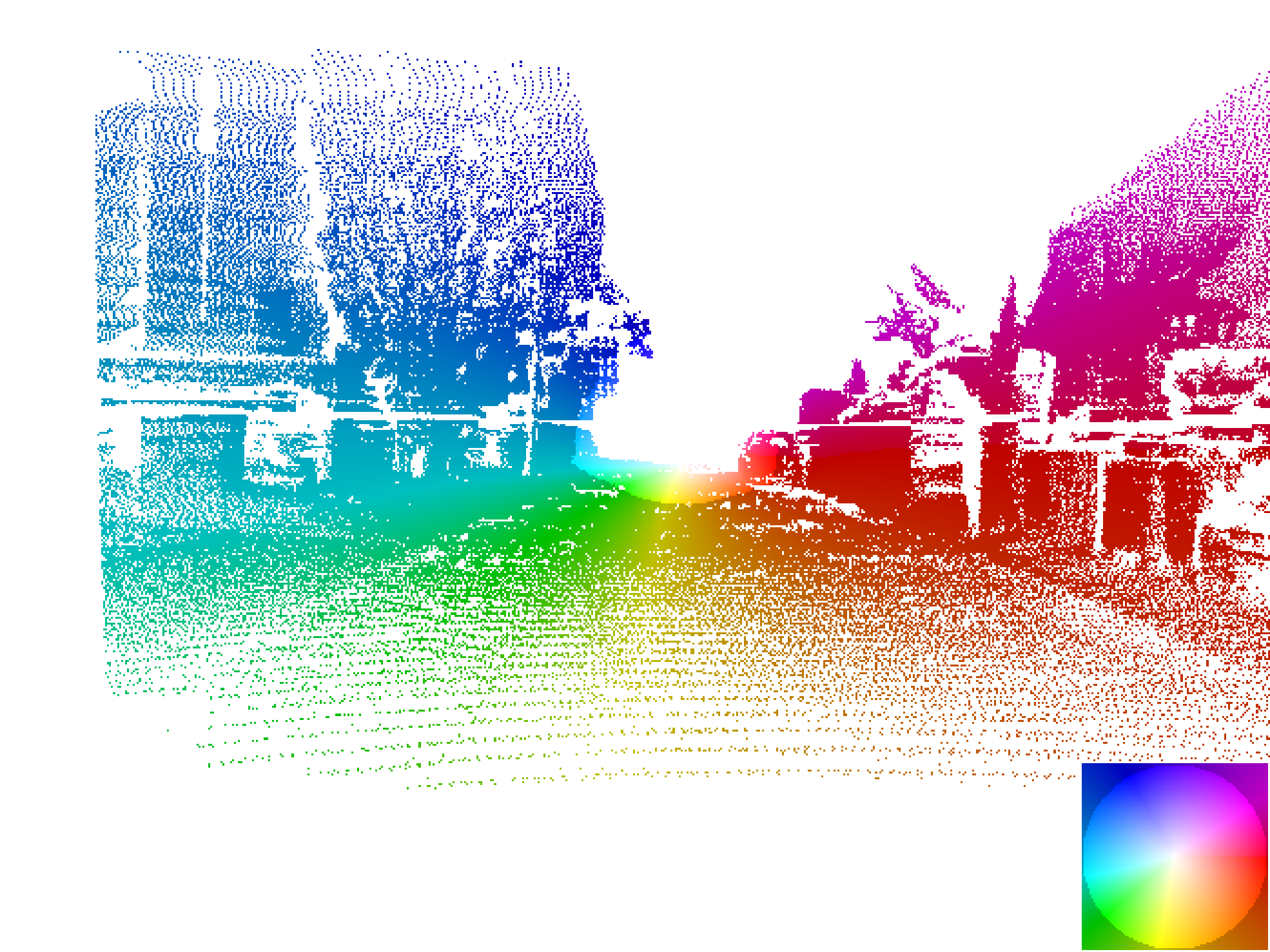} \end{tabular} &
        \begin{tabular}{@{}c@{}} \includegraphics[width=0.27\columnwidth, cfbox=gray 0.1pt 0pt]{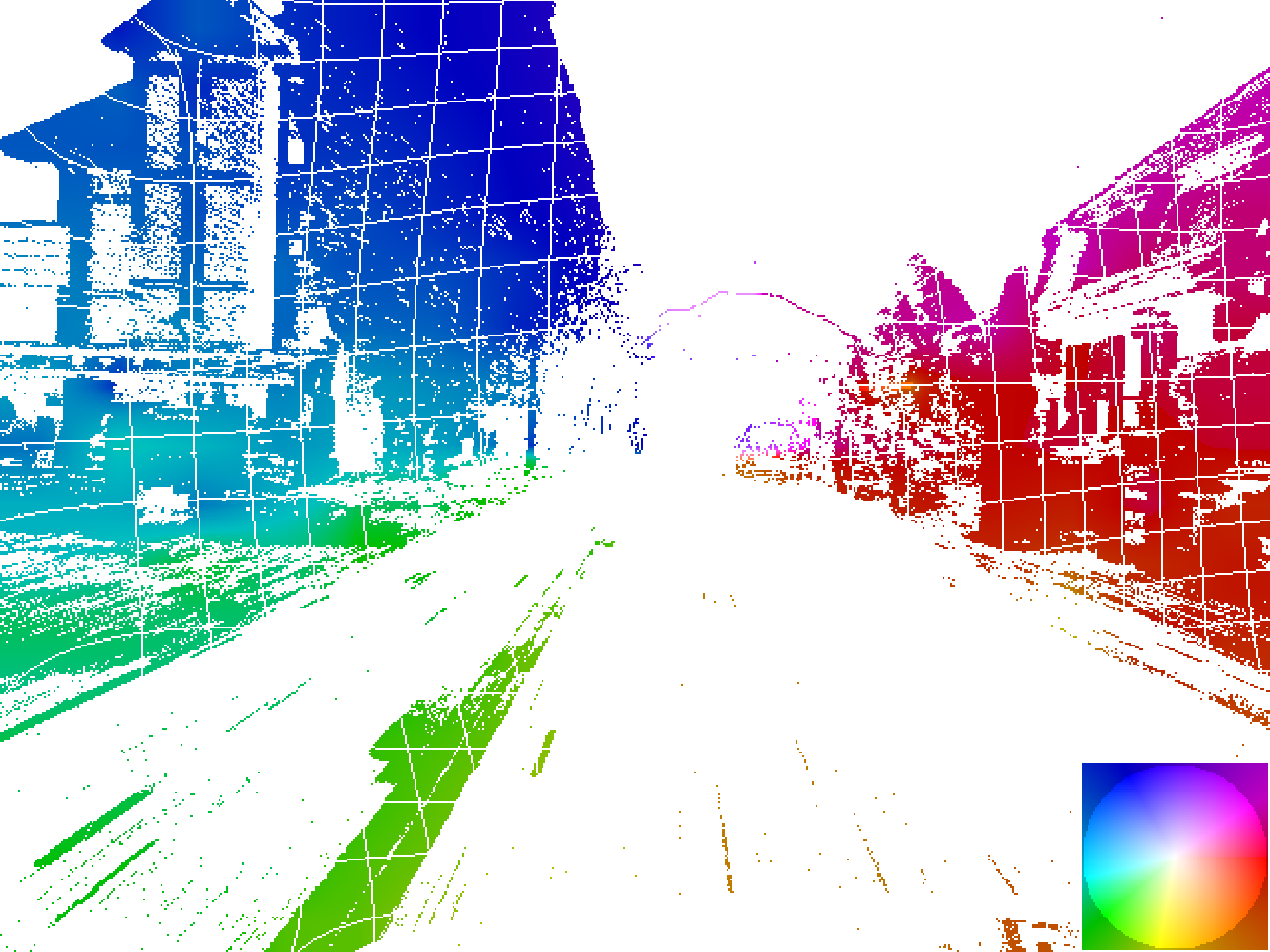} \end{tabular} & 
        \begin{tabular}{@{}c@{}} \includegraphics[width=0.27\columnwidth, cfbox=gray 0.1pt 0pt]{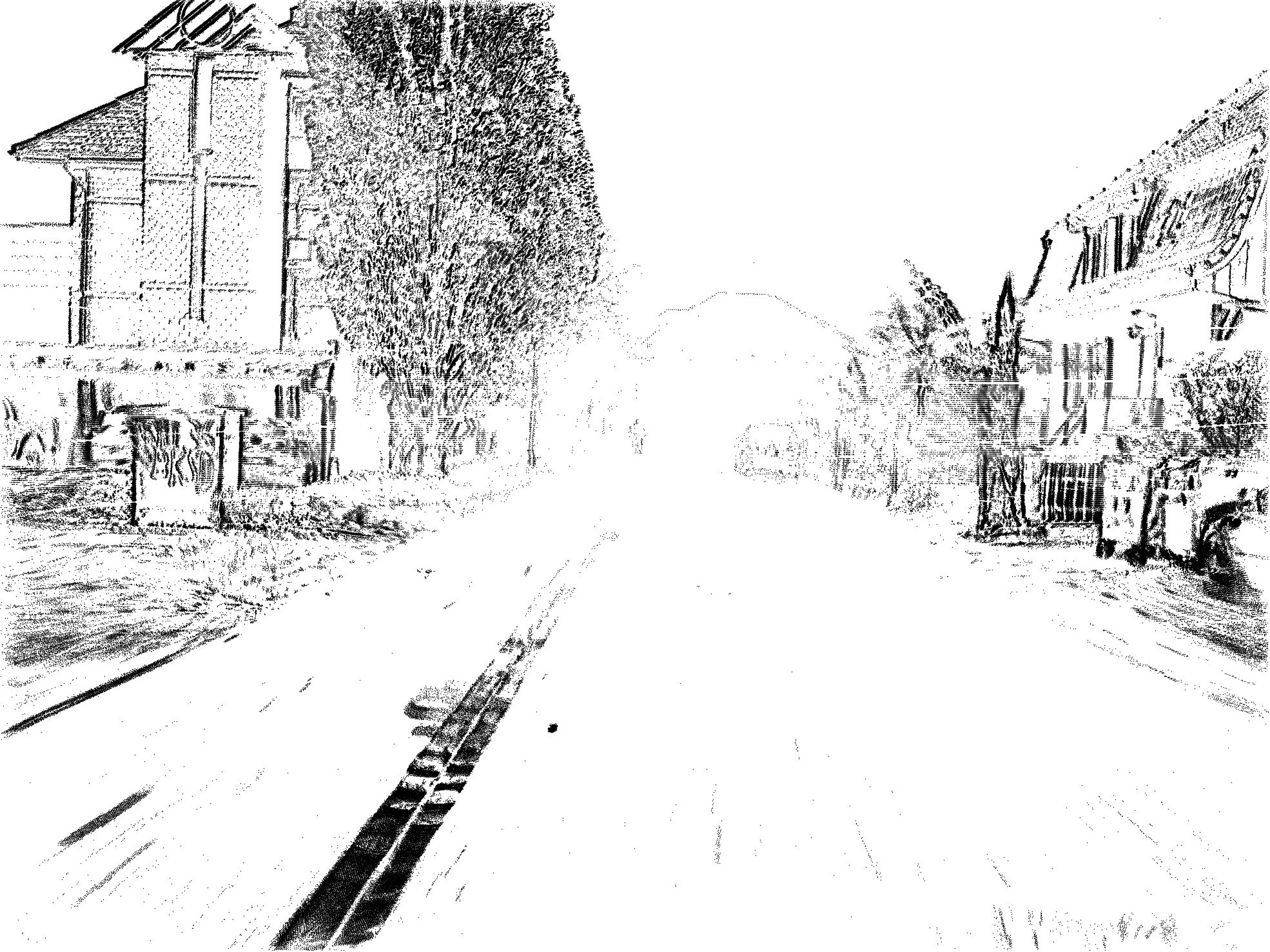} \end{tabular} \\

        % --------------------------------------------------------------------------------
        % last row
        
        \begin{tabular}{@{}c@{}} \begin{adjustbox}{max width=0.27\columnwidth}(a) $E$       \end{adjustbox}\end{tabular} & 
        \begin{tabular}{@{}c@{}} \begin{adjustbox}{max width=0.27\columnwidth}(b) $I_{\text{events}}$  \end{adjustbox}\end{tabular} &
        \begin{tabular}{@{}c@{}} \begin{adjustbox}{max width=0.27\columnwidth}(c) GT  \end{adjustbox}\end{tabular} &
        \begin{tabular}{@{}c@{}} \begin{adjustbox}{max width=0.27\columnwidth}(d) Pred. Flow \end{adjustbox}\end{tabular} &
        \begin{tabular}{@{}c@{}} \begin{adjustbox}{max width=0.27\columnwidth}(e) IWE  \end{adjustbox}\end{tabular} \\
    \end{tabular}

    \begin{tikzpicture}[overlay, remember picture, baseline=0pt,]
        \draw [violet,thick,opacity=1] ($(-9.44,1.19)+(0.45,-0.03)$) rectangle ++(0.6,-0.6);
        \draw [violet,thick,opacity=1] ($(-7.1,1.19)+(0.45,-0.03)$) rectangle ++(0.6,-0.6);
        \draw [violet,thick,opacity=1] ($(-4.7,1.19)+(0.45,-0.03)$) rectangle ++(0.6,-0.6);
        \draw [violet,thick,opacity=1] ($(-2.39,1.19)+(0.45,-0.03)$) rectangle ++(0.6,-0.6);
        
        \draw [orange,thick,opacity=1] ($(-9.44,1.19)+(0.05,-0.05)$) rectangle ++(0.5,-1.05);
        \draw [orange,thick,opacity=1] ($(-7.1,1.19)+(0.05,-0.05)$) rectangle ++(0.5,-1.05);
        \draw [orange,thick,opacity=1] ($(-4.7,1.19)+(0.05,-0.05)$) rectangle ++(0.5,-1.05);
        \draw [orange,thick,opacity=1] ($(-2.39,1.19)+(0.05,-0.05)$) rectangle ++(0.5,-1.05);
        
        \draw [lightgray!25!lime,thick,opacity=1] ($(-9.44,1.19)+(0.45,-1.18)$) rectangle ++(0.55,-0.5);
        \draw [lightgray!25!lime,thick,opacity=1] ($(-7.1,1.19)+(0.45,-1.18)$) rectangle ++(0.55,-0.5);
        \draw [lightgray!25!lime,thick,opacity=1] ($(-4.7,1.19)+(0.45,-1.18)$) rectangle ++(0.55,-0.5);
        \draw [lightgray!25!lime,thick,opacity=1] ($(-2.39,1.19)+(0.45,-1.18)$) rectangle ++(0.55,-0.5);
    \end{tikzpicture}
    
    \end{adjustbox}
    \caption{Qualitative results on additional factors that contribute to lower
    accuracy scores (compared to FWL) for our method on DSEC. The above
    instance is from the training set sequence \texttt{thun\_00\_a}. (a-c)
    depicts the edges, events, and available ground truth (GT). Note that
    although the IWE (e) looks sharp, the predicted flow (d) is only evaluated
    by the benchmark where valid GT exists. Regions in the image, such as the
    top-left (\protect\tikz[baseline] \protect\draw[color=orange, thick,
    opacity=1,] (0ex,0ex) rectangle (1.4ex,1.4ex);, \protect\tikz[baseline]
    \protect\draw[color=violet,thick, opacity=1] (0ex,0ex) rectangle (1.4ex,
    1.4ex);) and the bottom-center (\protect\tikz[baseline]
    \protect\draw[color=lightgray!25!lime,thick, opacity=1] (0ex,0ex) rectangle
    (1.4ex, 1.4ex);) with the building, trees and the lanes, respectively, are
    discounted from contributing to the accuracy due to sparse ground truth. In
    addition, the DSEC benchmark measures dense flow instead of sparse flow.
    Our method is model-based; hence, flow is reliably estimated only where
    events exist.}
    \label{fig:dsec_challenging}
\end{figure}

\begin{figure}
    \centering
    \begin{adjustbox}{max width=\columnwidth}
    \begin{tabular}{@{}c@{\thinspace}c@{\thinspace}c@{\thinspace}|@{\thinspace}c@{\thinspace}c}
    \vspace{-2pt}
    \rotatebox[origin=c]{90}{\begin{adjustbox}{max width=0.10\textwidth} \texttt{slider\_depth} \end{adjustbox}} &
    \begin{tabular}{@{}c@{}} \includegraphics[width=0.15\textwidth, cfbox=gray 0.1pt 0pt]{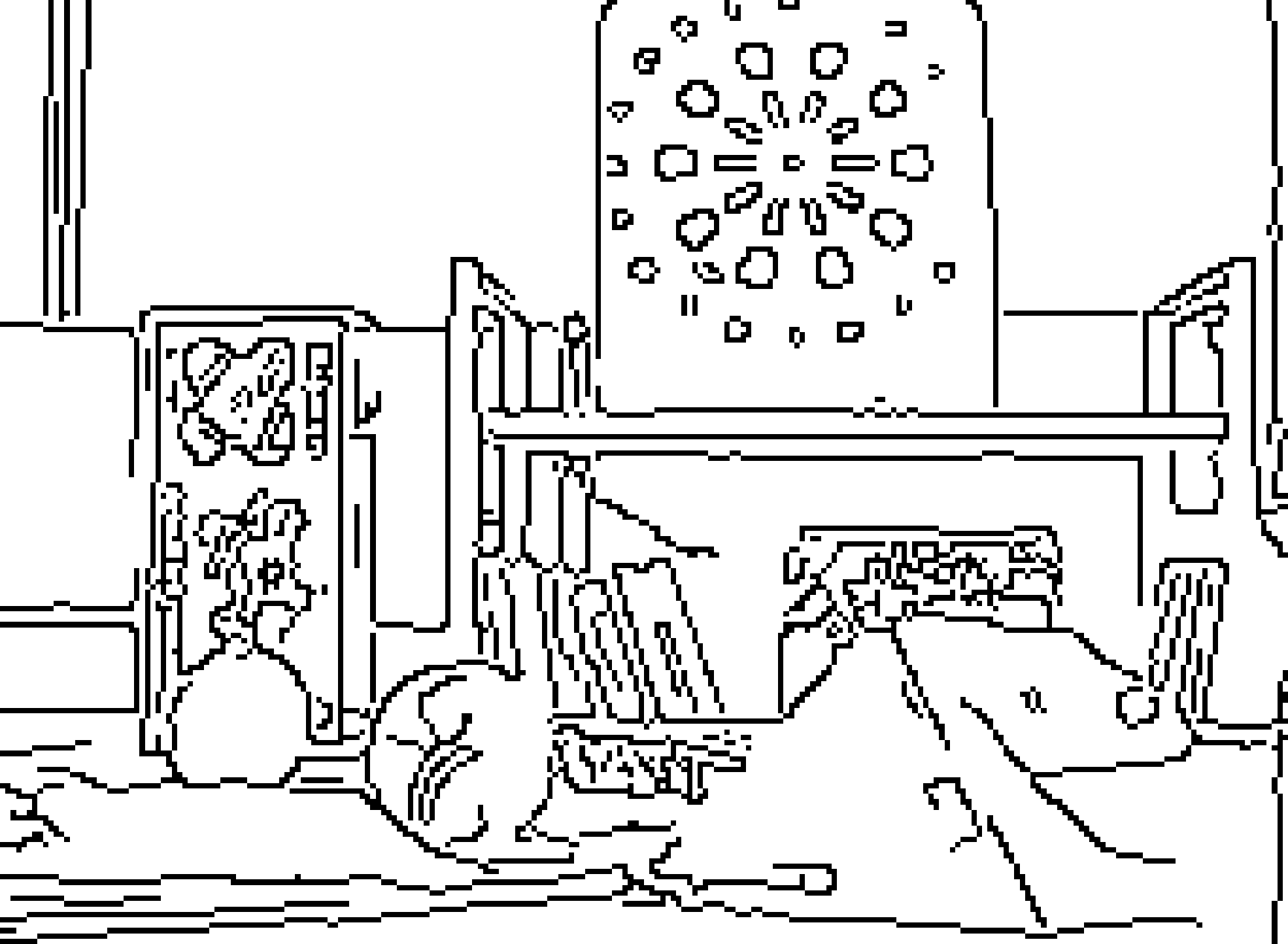} \end{tabular} &
    \begin{tabular}{@{}c@{}} \includegraphics[width=0.15\textwidth, cfbox=gray 0.1pt 0pt]{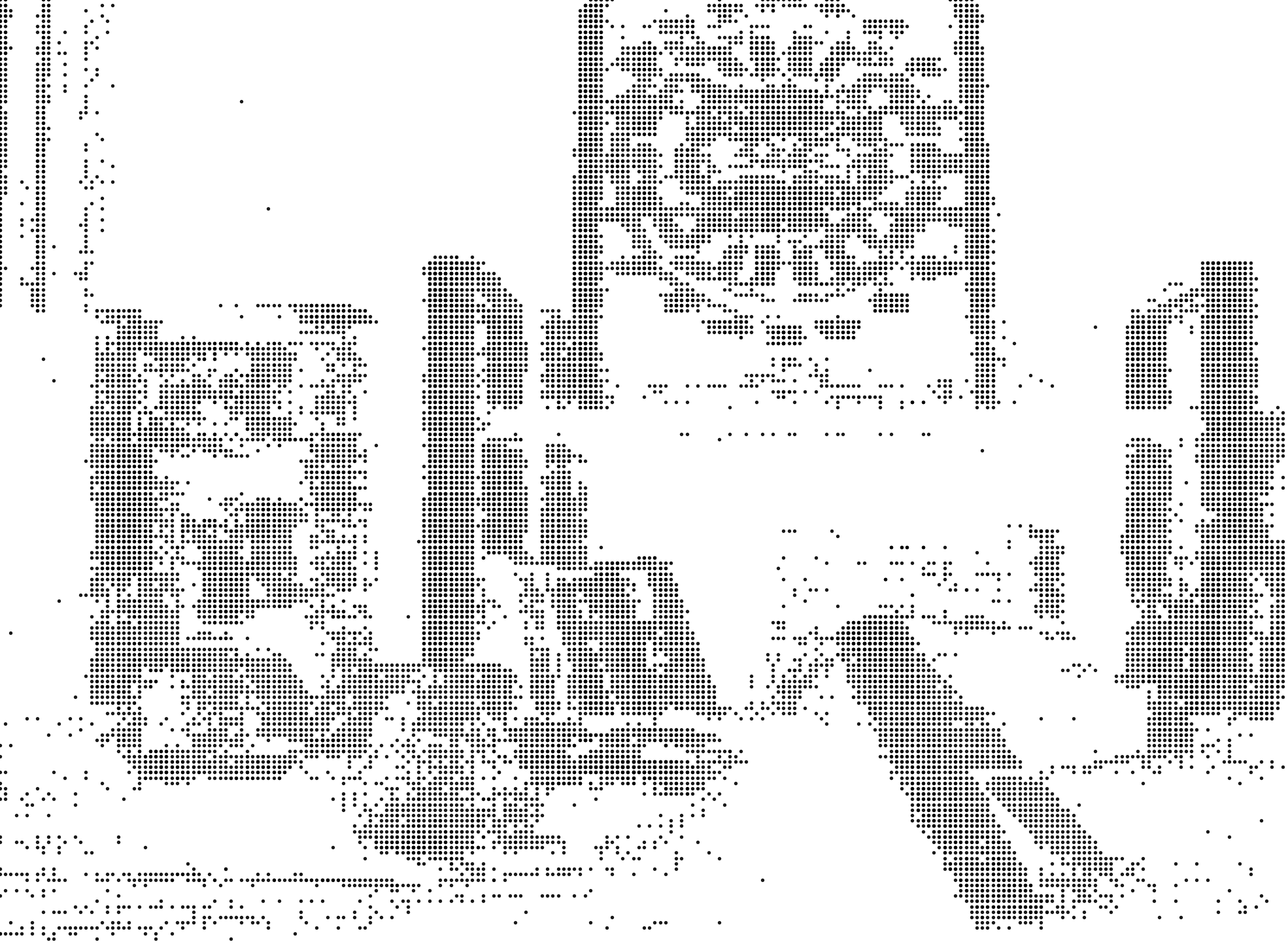} \end{tabular} &
    \begin{tabular}{@{}c@{}} \includegraphics[width=0.15\textwidth, cfbox=gray 0.1pt 0pt]{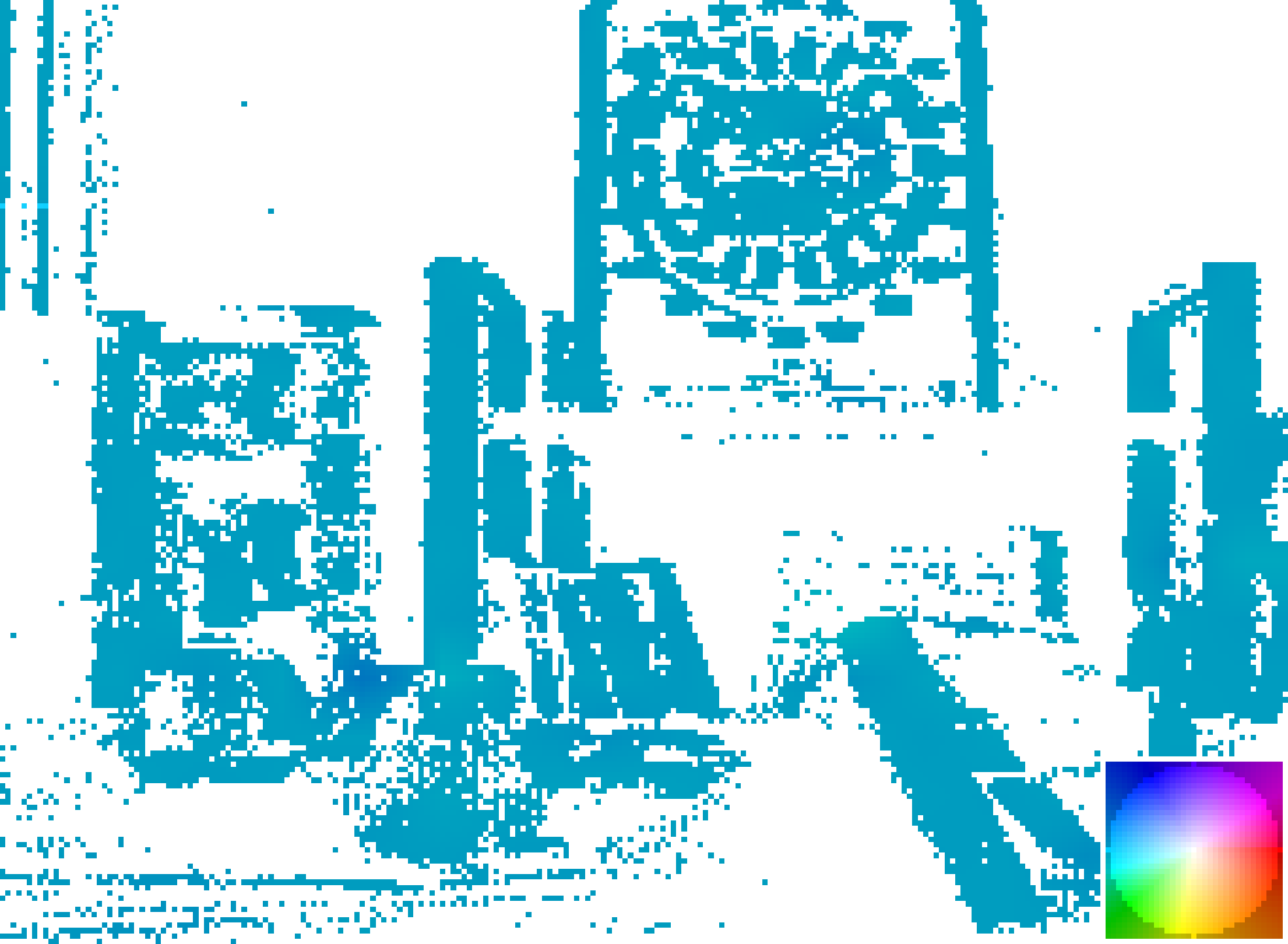} \end{tabular} &
    \begin{tabular}{@{}c@{}} \includegraphics[width=0.15\textwidth, cfbox=gray 0.1pt 0pt]{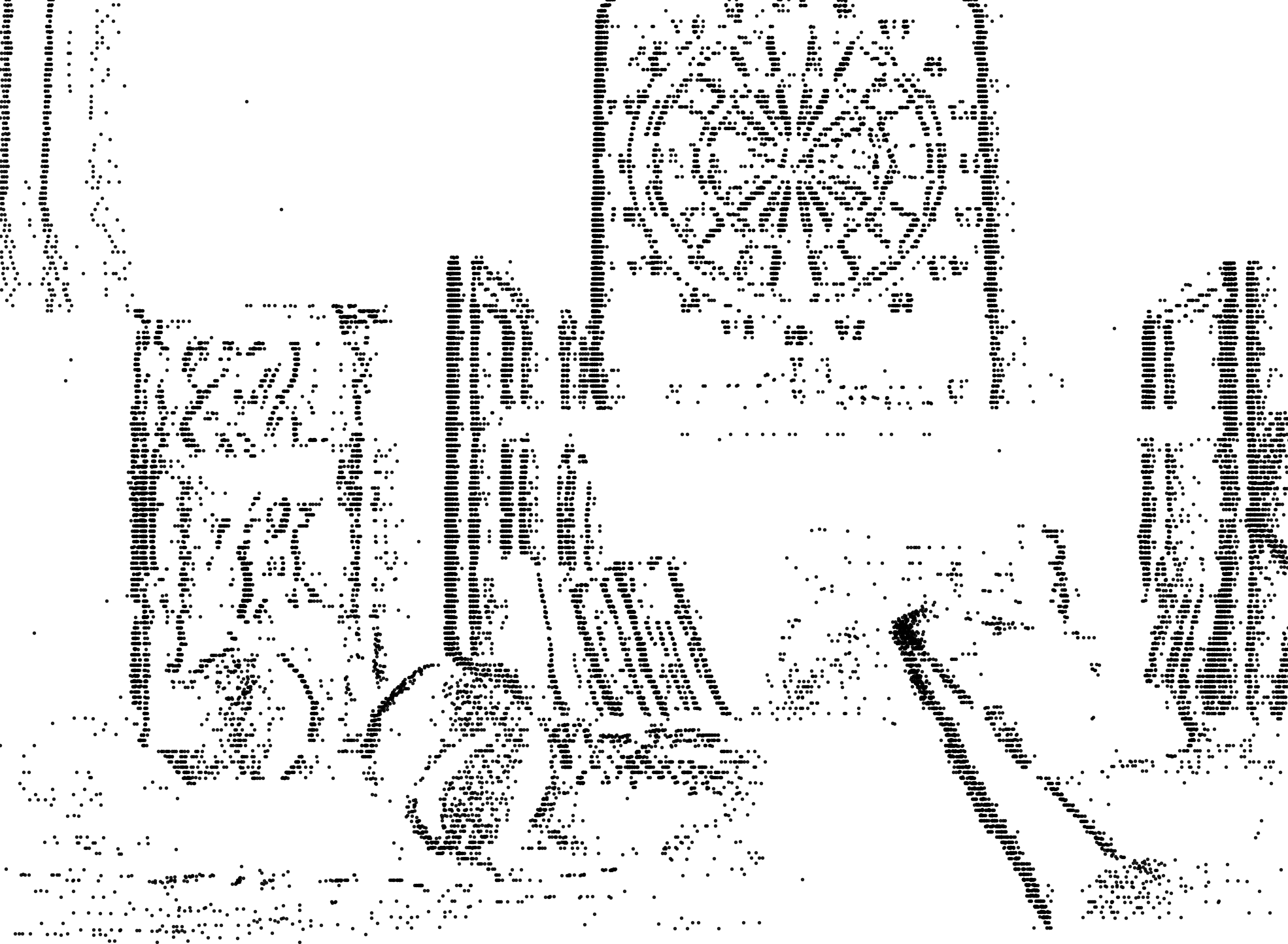} \end{tabular} \\ \vspace{-2pt}
    \rotatebox[origin=c]{90}{\begin{adjustbox}{max width=0.108\textwidth} \texttt{\phantom{\ \ }thun\_01\_a\phantom{\ \ }} \end{adjustbox}} &
    \begin{tabular}{@{}c@{}} \includegraphics[width=0.15\textwidth, cfbox=gray 0.1pt 0pt]{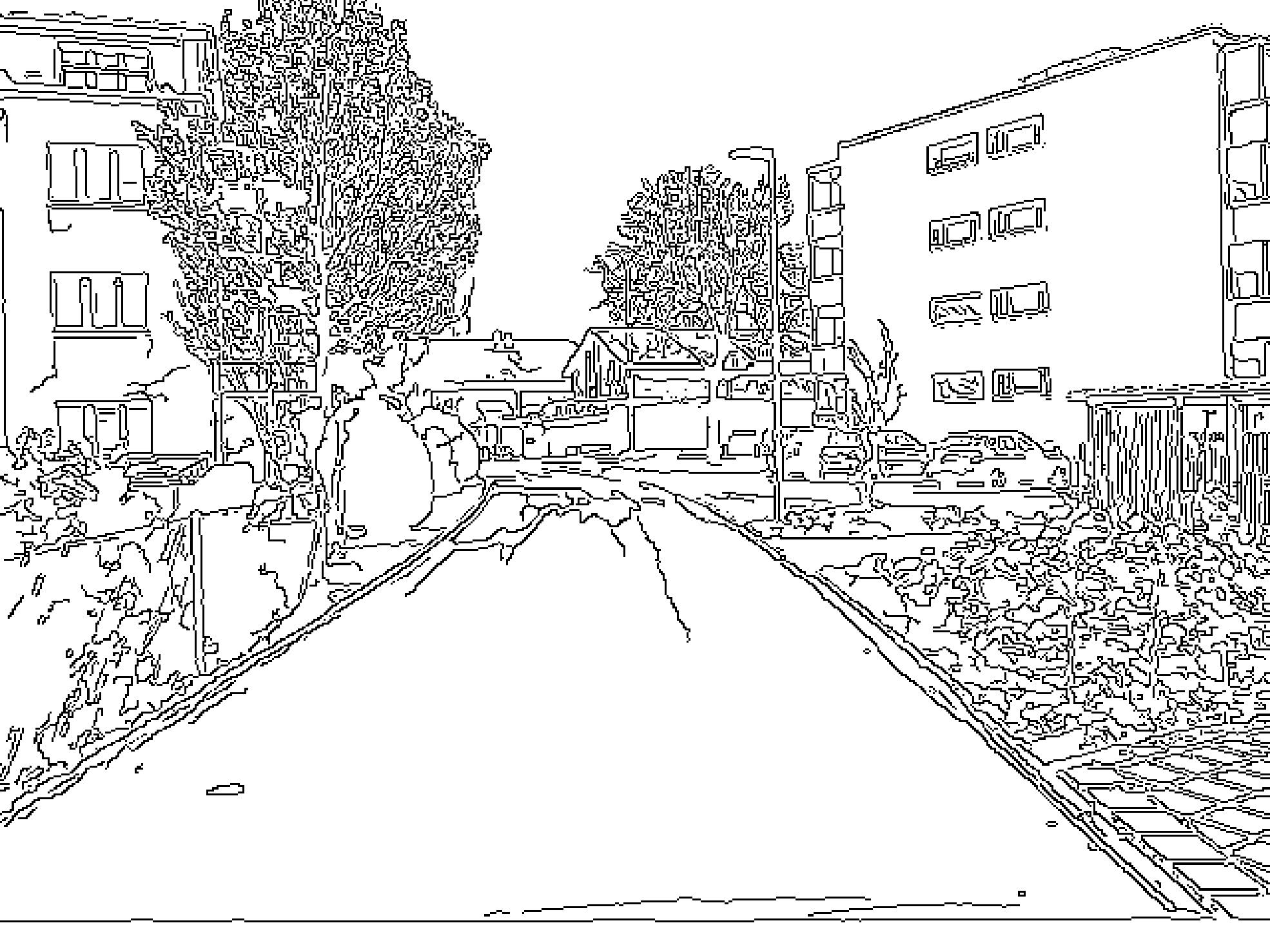} \end{tabular} &
    \begin{tabular}{@{}c@{}} \includegraphics[width=0.15\textwidth, cfbox=gray 0.1pt 0pt]{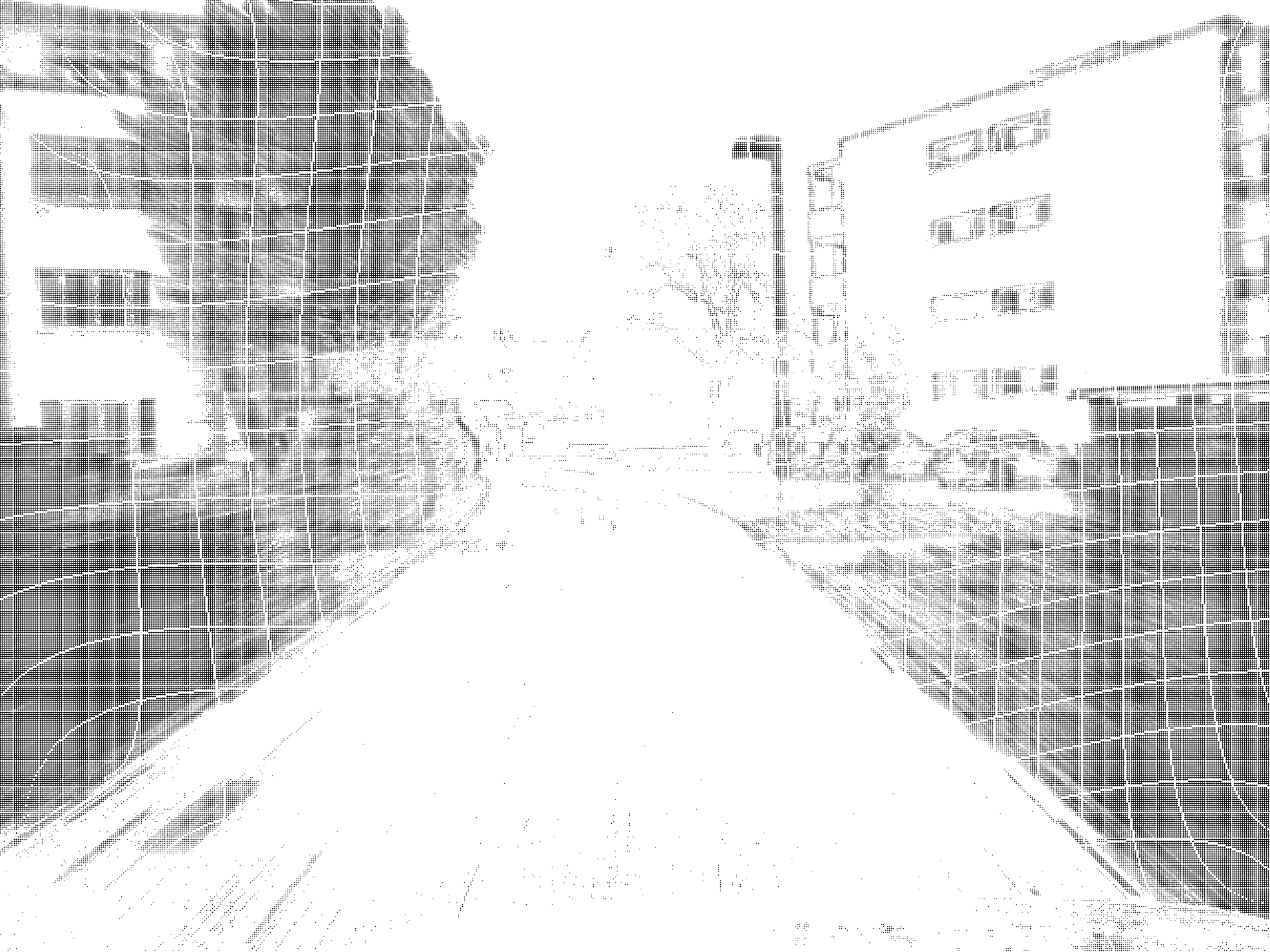} \end{tabular} &
    \begin{tabular}{@{}c@{}} \includegraphics[width=0.15\textwidth, cfbox=gray 0.1pt 0pt]{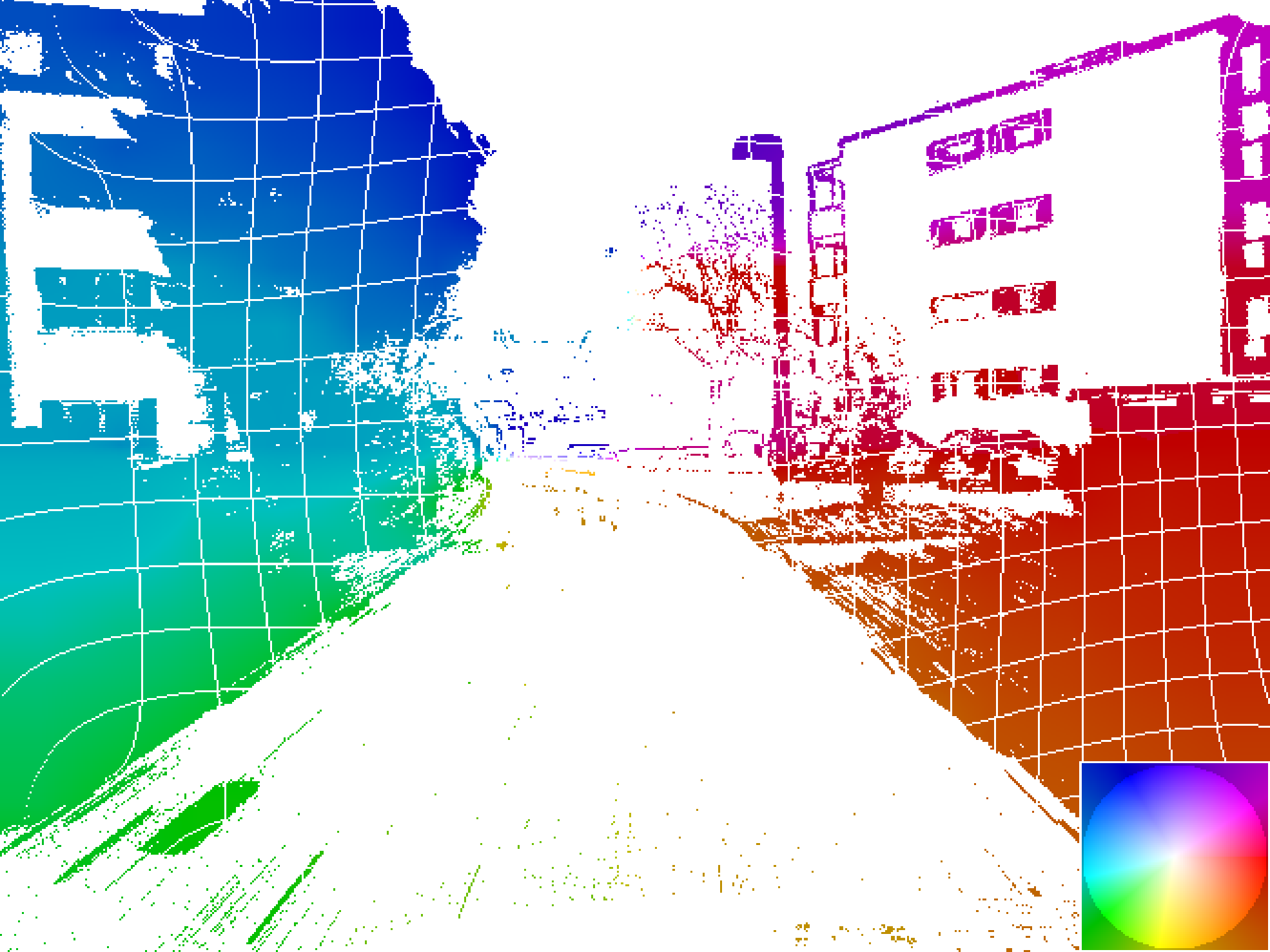} \end{tabular} &
    \begin{tabular}{@{}c@{}} \includegraphics[width=0.15\textwidth, cfbox=gray 0.1pt 0pt]{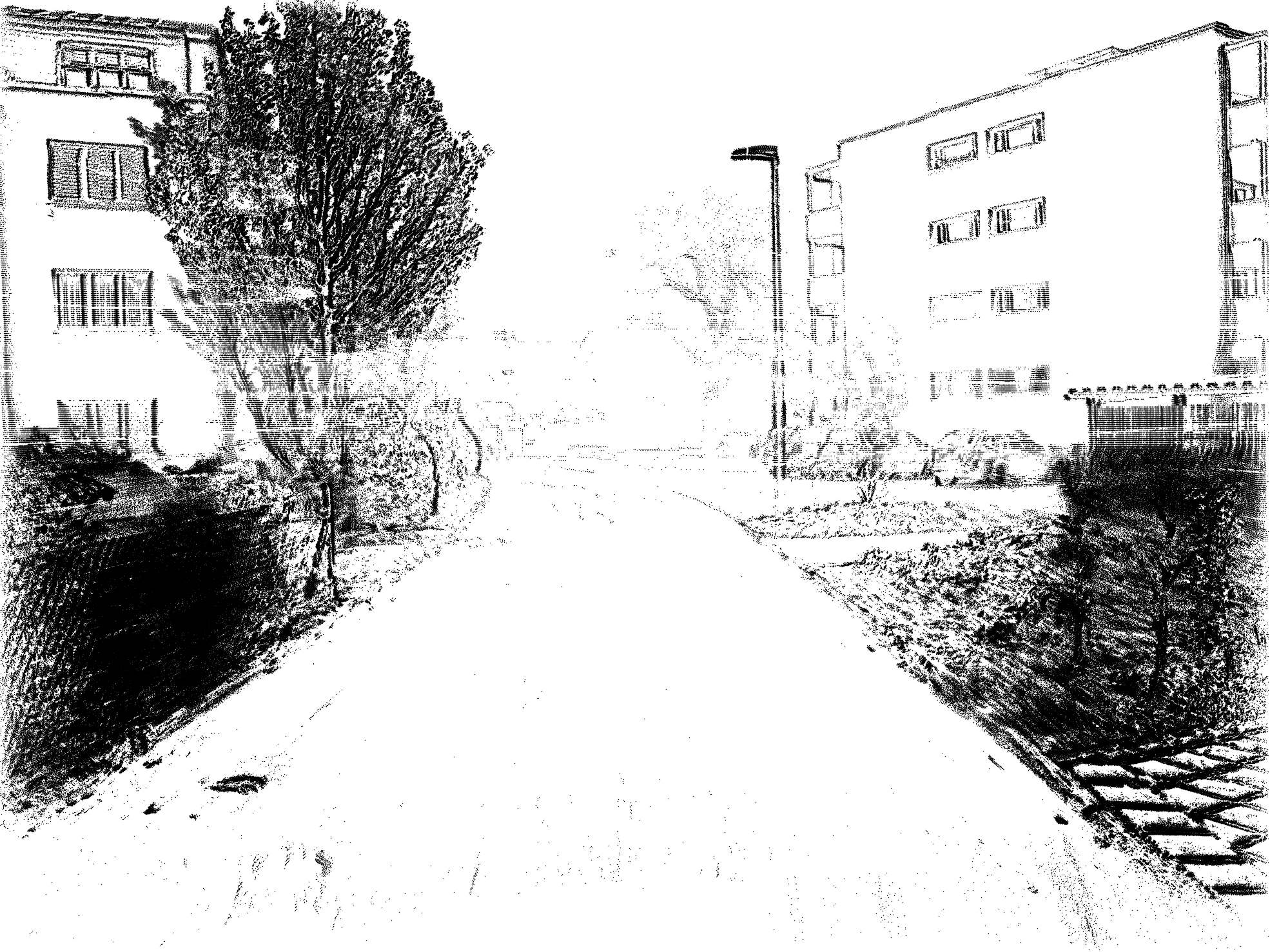} \end{tabular} \\
    \rotatebox[origin=c]{90}{\begin{adjustbox}{max width=0.108\textwidth} \texttt{\phantom{\ \ }thun\_01\_b\phantom{\ \ }} \end{adjustbox}} &
    \begin{tabular}{@{}c@{}} \includegraphics[width=0.15\textwidth, cfbox=gray 0.1pt 0pt]{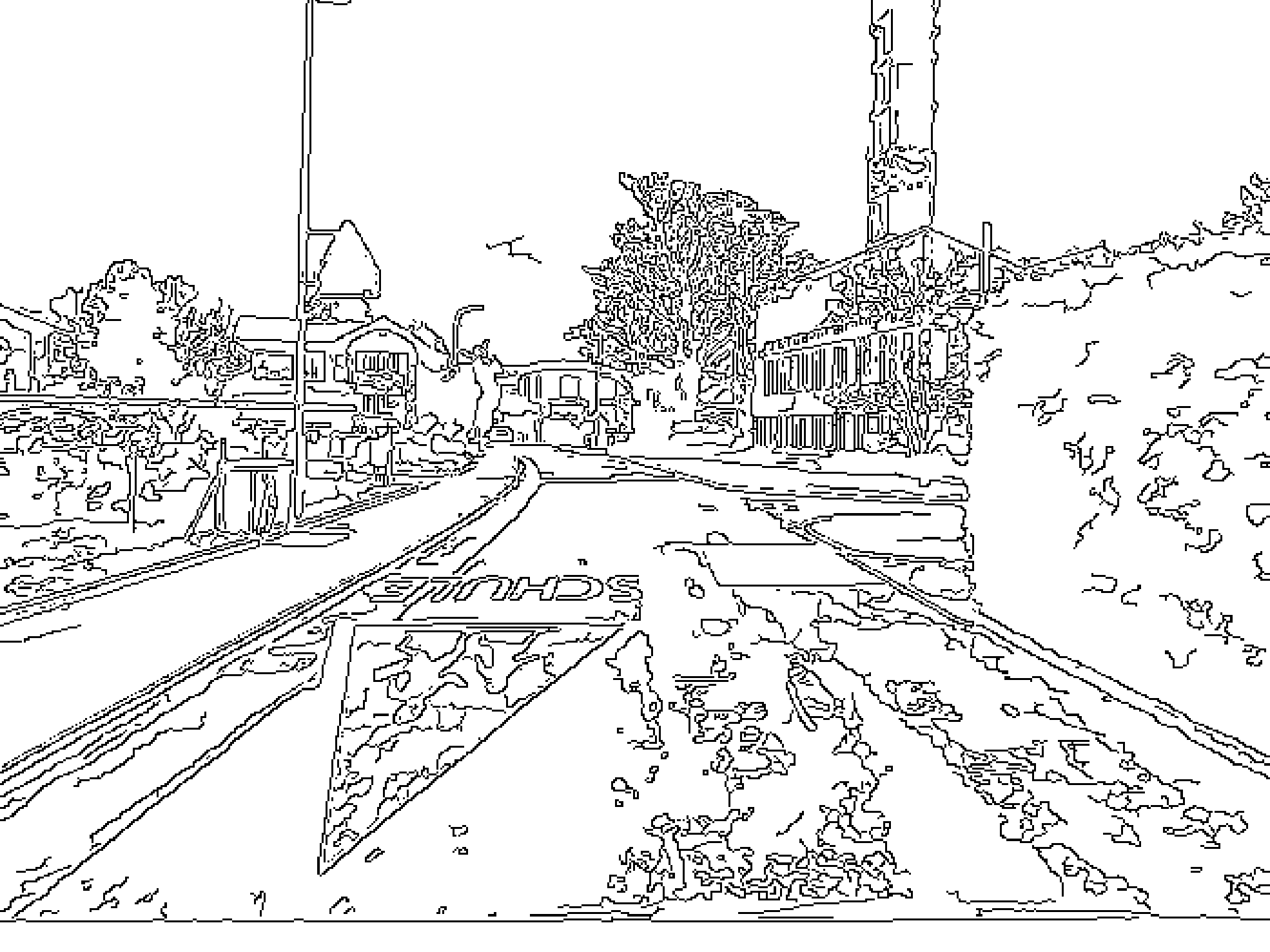} \end{tabular} &
    \begin{tabular}{@{}c@{}} \includegraphics[width=0.15\textwidth, cfbox=gray 0.1pt 0pt]{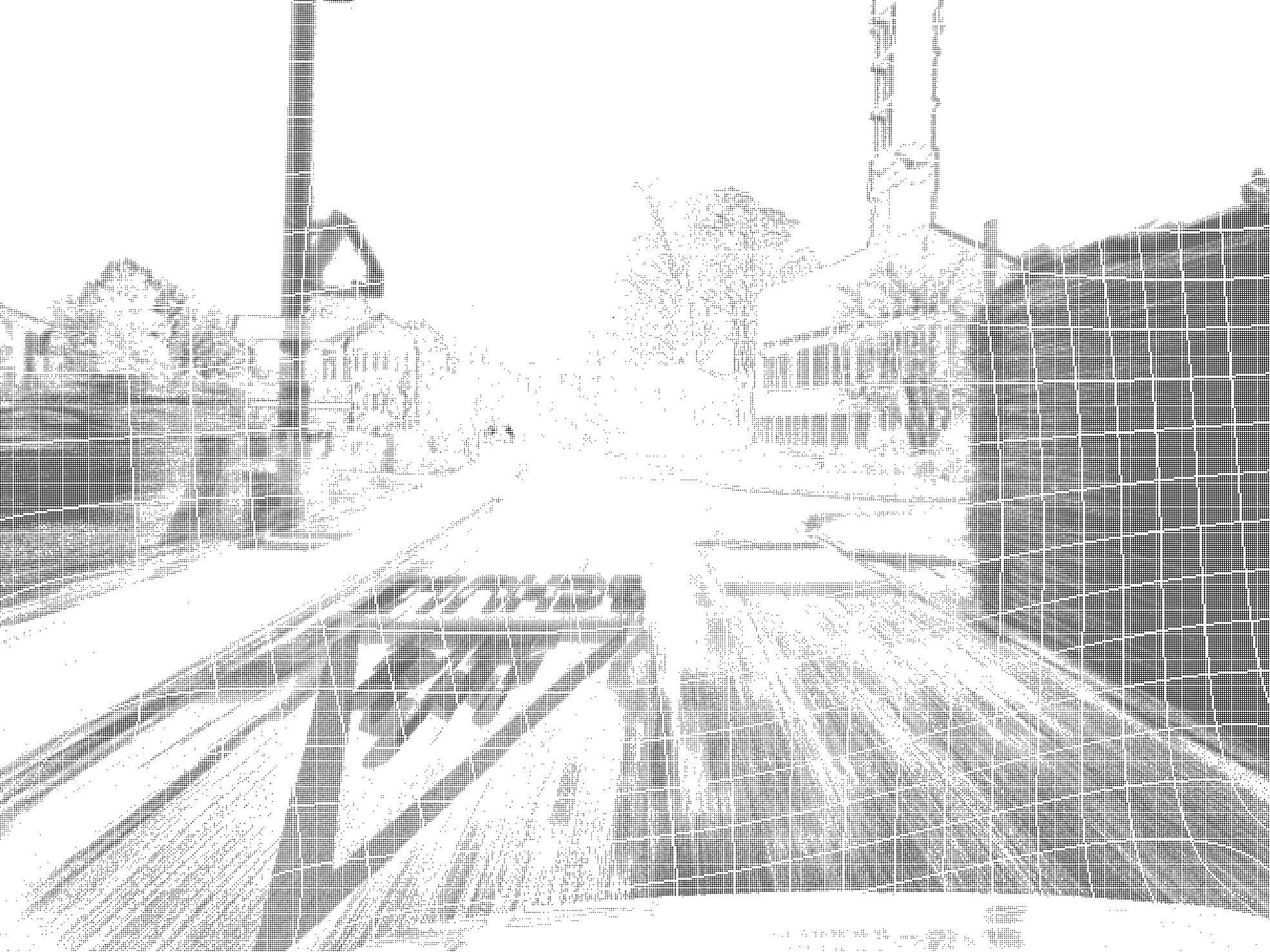} \end{tabular} &
    \begin{tabular}{@{}c@{}} \includegraphics[width=0.15\textwidth, cfbox=gray 0.1pt 0pt]{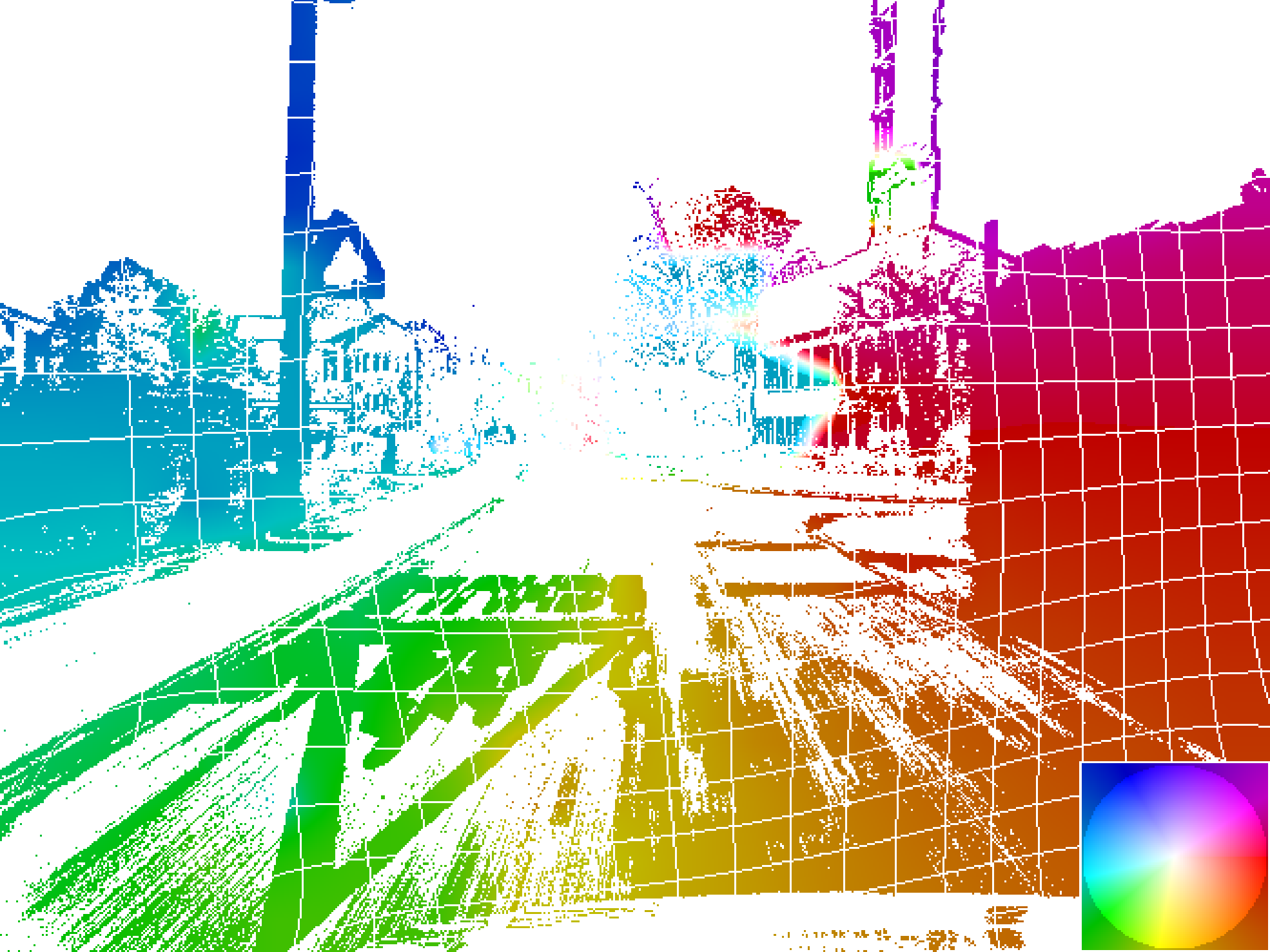} \end{tabular} &
    \begin{tabular}{@{}c@{}} \includegraphics[width=0.15\textwidth, cfbox=gray 0.1pt 0pt]{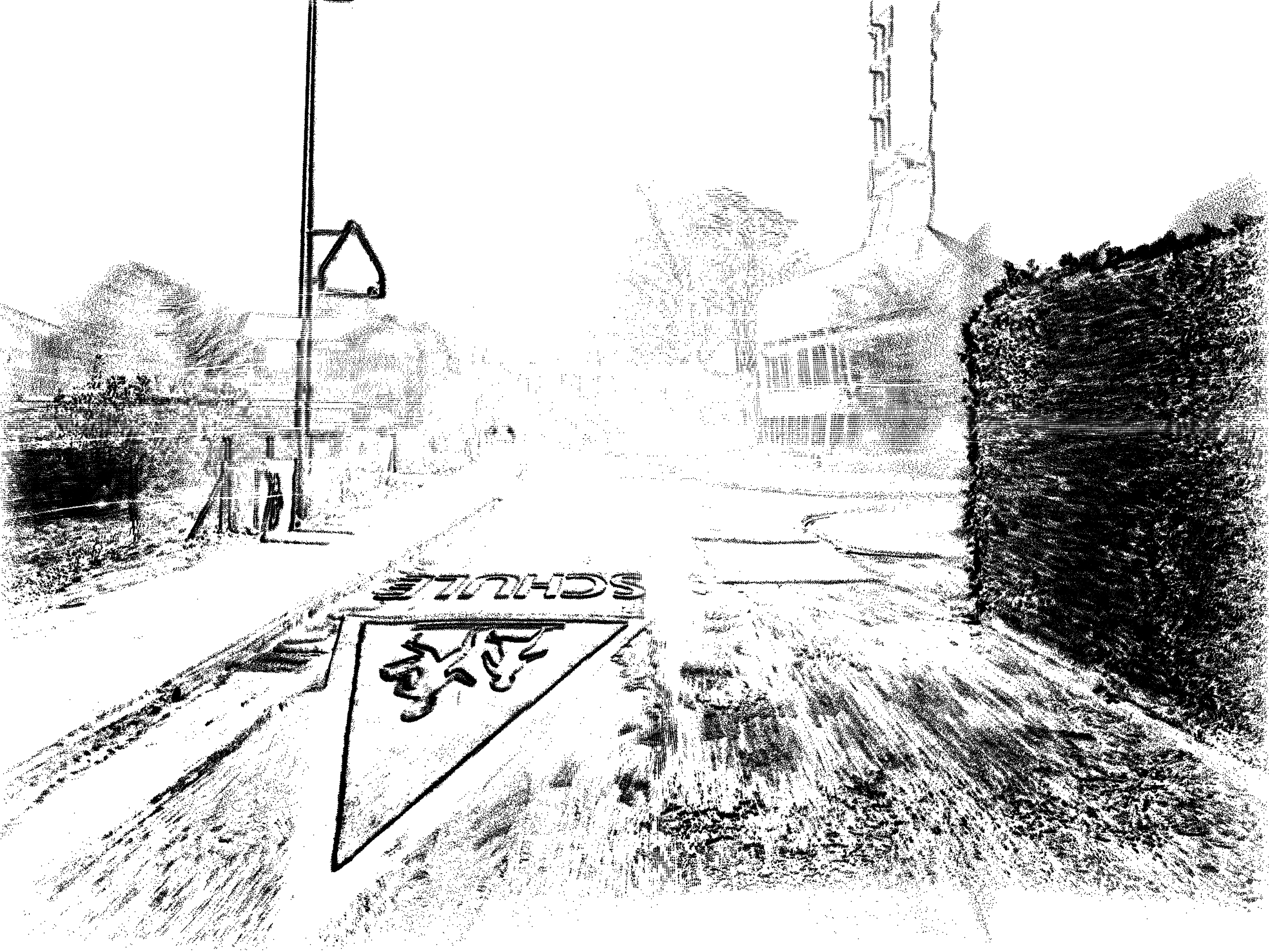} \end{tabular} \\

    &
    \begin{tabular}{@{}c@{}} \begin{adjustbox}{max width=0.15\textwidth}(a) $E$       \end{adjustbox}\end{tabular} &
    \begin{tabular}{@{}c@{}} \begin{adjustbox}{max width=0.15\textwidth}(b) $I_{\text{events}}$       \end{adjustbox}\end{tabular} &
    \begin{tabular}{@{}c@{}} \begin{adjustbox}{max width=0.15\textwidth}(c) Pred. Flow     \end{adjustbox}\end{tabular} &
    \begin{tabular}{@{}c@{}} \begin{adjustbox}{max width=0.15\textwidth}(d) IWE       \end{adjustbox}\end{tabular} \\
    
    \end{tabular}
    \end{adjustbox}
    \caption{Qualitative results for the ECD sequence \texttt{slider\_depth}
    and DSEC test set sequences \texttt{thun\_01\_a} and \texttt{thun\_01\_b}
    in rows 1, 2 and 3, respectively. Columns 1, 2, 3, and 4 depict edges,
    IUEs, predicted flows, and IWEs, respectively.}
    \label{fig:dsec_ecd_evaluations}
\end{figure}

We evaluated our approach on the DSEC benchmark and report the results in
\cref{tab:dsec_testset_results}. With respect to ground-truth flow accuracy,
EINCM is competitive with the previous MB state of the art
\cite{shiba2022secrets} and provides higher sharpness in constructed IWEs (\ie,
higher FWL scores). Recall that in DSEC at least 20\% of the pixel
displacements are greater than 22 pixels with maximum displacements up to 210
pixels. All MB methods \cite{shiba2022secrets, brebion2021realtimeflow},
including ours, have significantly worse evaluation accuracies (AEE $\approx$
2-3) with respect to the ground truth. This was largely explained in
\cite{shiba2022secrets} by noting that E-RAFT \cite{gehrig2021eraft} is a
supervised learning method, tailored to predict the same type of signals as the
ground truth. To shed further light on why our evaluation falls behind
supervised learning techniques such as \cite{gehrig2021eraft}, we observe that
most events in the DSEC driving sequences occur on two sides of the road, where
ground-truth signals are inaccurate or not available, as shown in
\cref{fig:dsec_challenging}. We also note that the events and images in DSEC
were recorded on two different cameras and the registration (\ie, mapping
between events and image frames) based on depth is imperfect, which can
contribute to performance degradation. Nevertheless, EINCM produces
significantly better IWEs (\cref{tab:fwl_scores} and
\cref{fig:dsec_ecd_evaluations}) and practically more realistic flows.

%%%%%%%%%%%%%%%%%%%%%%%%%%%%%%%%%%%%%%%%%%%%%%%%%%%%%%%%%%%%%%%%%%%%%%%%%%%%%%%%
% Table: DSEC Accuracies and FWLS
\begin{table}
    \centering
    \begin{adjustbox}{max width=\columnwidth}
        \begin{tabular}{@{\thinspace}c@{\medspace}l@{\thickspace}c@{\thickspace}
        c@{\thickspace}c@{\thickspace}c@{\thickspace}c@{\thickspace}c@{\thickspace}c
        c@{\thickspace}c@{\thickspace}c@{\thickspace}c@{\thickspace}c@{\thickspace}c
        c@{\thickspace}c@{\thickspace}c@{\thickspace}c@{\thickspace}c} 
    
            % ================================================================================
            \toprule
            & & &
            \multicolumn{5}{c}{\texttt{thun\_01\_a}}              & \thickspace &     % thun_01_a
            \multicolumn{5}{c}{\texttt{thun\_01\_b}}              & \thickspace &     % thun_01_b
            \multicolumn{5}{c}{\texttt{zurich\_city\_15\_a}}      \\                  % zurich_city_15_a    
            % --------------------------------------------------------------------------------
            \cmidrule(){4-8} \cmidrule(){10-14} \cmidrule(){16-20}

            & & &
            AEE $\downarrow$         & \thickspace & \%Out $\downarrow$         & \thickspace & FWL $\uparrow$        & \thickspace &     % thun_01_a 
            AEE $\downarrow$         & \thickspace & \%Out $\downarrow$         & \thickspace & FWL $\uparrow$        & \thickspace &     % thun_01_b 
            AEE $\downarrow$         & \thickspace & \%Out $\downarrow$         & \thickspace & FWL $\uparrow$        \\                  % zurich_city_15_a 
            % --------------------------------------------------------------------------------
            \midrule 
            
            \multirow{1}{*}{\rotatebox[origin=c]{90}{SL}}
            & E-RAFT \cite{gehrig2021eraft} & &
            \textbf{0.65}             & \thickspace & \textbf{1.87}            & \thickspace & 1.20                       & \thickspace &     % thun_01_a 
            \textbf{0.58}             & \thickspace & \textbf{1.52}            & \thickspace & 1.18                       & \thickspace &     % thun_01_b 
            \textbf{0.59}             & \thickspace & \textbf{1.30}            & \thickspace & 1.34                       \\                  % zurich_city_15_a 
            \midrule

            \multirow{3}{*}{\rotatebox[origin=c]{90}{MB}} 
            & Brebion \etal \cite{brebion2021realtimeflow} & &
            3.01                     & \thickspace & 29.69\tiny{7}             & \thickspace & --                         & \thickspace &     % thun_01_a 
            3.91\tiny{3}             & \thickspace & 34.69                     & \thickspace & --                         & \thickspace &     % thun_01_b 
            3.78\tiny{1}             & \thickspace & 37.98\tiny{7}             & \thickspace & --                         \\                  % zurich_city_15_a
            
            & Shiba \etal \cite{shiba2022secrets} & &
            2.12                     & \thickspace & 17.68                     & \thickspace & 1.24                       & \thickspace &     % thun_01_a 
            \underline{2.48}         & \thickspace & \underline{23.56}         & \thickspace & 1.124                      & \thickspace &     % thun_01_b 
            \underline{2.35}         & \thickspace & \underline{20.99}         & \thickspace & 1.41                       \\                  % zurich_city_15_a
            
            & Ours (EINCM) & &
            \underline{2.01\tiny{5}} & \thickspace & \underline{16.17\tiny{4}}  & \thickspace & \textbf{1.40}      & \thickspace &     % thun_01_a 
            2.77\tiny{8}             & \thickspace & 26.56                     & \thickspace & \textbf{1.39\tiny{6}}      & \thickspace &     % thun_01_b 
            3.00\tiny{5}             & \thickspace & 26.63\tiny{3}             & \thickspace & \textbf{1.60\tiny{3}}      \\                  % zurich_city_15_a
            % ================================================================================
            \bottomrule
        \end{tabular}
    \end{adjustbox}
    \caption{Accuracy and FWL scores on the DSEC test sequences (full results
    are reported in the supplementary material). \emph{Bold} and
    \emph{underline} typefaces indicate the \textbf{best} and the
    \underline{second best}, respectively.}
    \label{tab:dsec_testset_results}
\end{table}

%%%%%%%%%%%%%%%%%%%%%%%%%%%%%%%%%%%%%%%%%%%%%%%%%%%%%%%%%%%%%%%%%%%%%%%%%%%%%%%%

% ------------------------------------------------------------------------------
\subsection{Effect of Different Handover Strategies}
\label{subsec:effect_of_different_handover_strategies}
We conducted an ablation study on handover sub-strategies by evaluating the FWL
on several sequences of the DSEC test set
(\cref{tab:dsec_testset_fwls_ablation}). The all-solved weights (SHO) case
unexpectedly underperformed compared to fixed weights (FHO). We hypothesize
that the solved weights become overfitted at coarser pyramid levels. This was
observed in the form of high magnitude and variance for $w_{\text{ho}}$
estimates when subjected to unbounded, unconstrained optimization. Furthermore,
in the MVSEC experiments we found that for a fixed weight $w_{\text{ho}}$, a
higher value (\eg, 0.67) gave better performance over smaller values (\eg, 0.5)
when there was less overall motion in the scene and therefore the subsequent
predicted flow needed to differ by only a small amount. In general, we found
that the best strategy is a combination of fixed and solved weights for the
handover (FSHO) operation across the pyramid levels, where we used fixed
weights on coarser scales and solved weights on finer scales.

%%%%%%%%%%%%%%%%%%%%%%%%%%%%%%%%%%%%%%%%%%%%%%%%%%%%%%%%%%%%%%%%%%%%%%%%%%%%%%%%
% Table: DSEC FWLs Ablation
\begin{table}
    \centering
    \begin{adjustbox}{max width=\columnwidth}
        \begin{tabular}{@{\thinspace}l@{\thickspace}
        c@{\thickspace}c@{\thickspace} 
        c@{\thickspace}c@{\thickspace} 
        c@{\thickspace}c@{\thickspace} 
        c@{\thickspace}c@{\thickspace} 
        c@{\thickspace}c@{\thickspace} 
        c@{\thickspace}c@{\thickspace} 
        c}
    
            % ================================================================================
            \toprule
            & 
            \multicolumn{1}{c}{\texttt{int\_00\_b}}       & \thickspace &     % interlaken_00_b
            \multicolumn{1}{c}{\texttt{int\_01\_a}}       & \thickspace &     % interlaken_01_a
            \multicolumn{1}{c}{\texttt{thu\_01\_a}}       & \thickspace &     % thun_01_a
            \multicolumn{1}{c}{\texttt{thu\_01\_b}}       & \thickspace &     % thun_01_b
            \multicolumn{1}{c}{\texttt{zur\_12\_a}}       & \thickspace &     % zurich_city_12_a
            \multicolumn{1}{c}{\texttt{zur\_14\_c}}       & \thickspace &     % zurich_city_14_c
            \multicolumn{1}{c}{\texttt{zur\_15\_a}}       \\                  % zurich_city_15_a    
            % --------------------------------------------------------------------------------
            \cmidrule(){2-2} \cmidrule(){4-4} \cmidrule(){6-6} \cmidrule(){8-8} \cmidrule(){10-10} \cmidrule(){12-12} \cmidrule(){14-14}  

            & 
            FWL $\uparrow$                                & \thickspace &     % interlaken_00_b
            FWL $\uparrow$                                & \thickspace &     % interlaken_01_a
            FWL $\uparrow$                                & \thickspace &     % thun_01_a 
            FWL $\uparrow$                                & \thickspace &     % thun_01_b 
            FWL $\uparrow$                                & \thickspace &     % zurich_city_12_a
            FWL $\uparrow$                                & \thickspace &     % zurich_city_14_c
            FWL $\uparrow$                                \\                  % zurich_city_15_a 
            % --------------------------------------------------------------------------------
            \midrule

            Ours (EINCM-SHO) & 
            1.51\tiny{3}                                  & \thickspace &     % interlaken_00_b
            1.70\tiny{2}                                  & \thickspace &     % interlaken_01_a
            1.30\tiny{3}                                  & \thickspace &     % thun_01_a 
            1.32\tiny{9}                                  & \thickspace &     % thun_01_b 
            0.6744                                        & \thickspace &     % zurich_city_12_a
            \underline{1.52\tiny{6}}                      & \thickspace &     % zurich_city_14_c
            1.47\tiny{3}                                  \\                  % zurich_city_15_a
            Ours (EINCM-FHO) & % F(0.5)
            \underline{1.65\tiny{9}}                      & \thickspace &     % interlaken_00_b
            \underline{1.74\tiny{6}}                      & \thickspace &     % interlaken_01_a
            \underline{1.32\tiny{6}}                      & \thickspace &     % thun_01_a 
            \underline{1.36\tiny{2}}                      & \thickspace &     % thun_01_b 
            \underline{1.15\tiny{9}}                      & \thickspace &     % zurich_city_12_a
            1.38\tiny{6}                                  & \thickspace &     % zurich_city_14_c
            \underline{1.53\tiny{7}}                      \\                  % zurich_city_15_a
            Ours (EINCM-FSHO) & % F(0.5) 4-2, S 1,0,
            \textbf{1.75\tiny{8}}                         & \thickspace &     % interlaken_00_b
            \textbf{1.75\tiny{5}}                         & \thickspace &     % interlaken_01_a
            \textbf{1.45\tiny{8}}                         & \thickspace &     % thun_01_a 
            \textbf{1.40\tiny{5}}                         & \thickspace &     % thun_01_b 
            \textbf{1.34\tiny{3}}                         & \thickspace &     % zurich_city_12_a
            \textbf{1.54\tiny{2}}                         & \thickspace &     % zurich_city_14_c
            \textbf{1.61\tiny{3}}                         \\                  % zurich_city_15_a
            % ================================================================================
            \bottomrule
        \end{tabular}
    \end{adjustbox}
    \caption{\emph{Handover} sub-strategy ablations evaluated on the DSEC test
    set. We report FWL scores using three \emph{handover} sub-strategies:
    solved handover (SHO), fixed handover (FHO), and fixed+solved handover
    (FSHO). For SHO, $w_{\text{ho}}^{\ast}$ is solved by optimizing for the CM
    objective at all pyramid levels. For FHO, $w_{\text{ho}}=0.5$ for all
    pyramid levels. For FSHO, $w_{\text{ho}}=0.5$ for levels 4, 3, 2, and
    solved $w_{\text{ho}}^{\ast}$ at levels 1 and 0. \emph{Bold} and
    \emph{underline} typefaces indicate the \textbf{best} and the
    \underline{second best}, respectively.}
    \label{tab:dsec_testset_fwls_ablation}
\end{table}

%%%%%%%%%%%%%%%%%%%%%%%%%%%%%%%%%%%%%%%%%%%%%%%%%%%%%%%%%%%%%%%%%%%%%%%%%%%%%%%%

%%%%%%%%%%%%%%%%%%%%%%%%%%%%%%%%%%%%%%%%%%%%%%%%%%%%%%%%%%%%%%%%%%%%%%%%%%%%%%%%
% Figure: GT Challenges

%%%%%%%%%%%%%%%%%%%%%%%%%%%%%%%%%%%%%%%%%%%%%%%%%%%%%%%%%%%%%%%%%%%%%%%%%%%%%%%%

%%%%%%%%%%%%%%%%%%%%%%%%%%%%%%%%%%%%%%%%%%%%%%%%%%%%%%%%%%%%%%%%%%%%%%%%%%%%%%%%
% Figure: DSEC GT challenges

%%%%%%%%%%%%%%%%%%%%%%%%%%%%%%%%%%%%%%%%%%%%%%%%%%%%%%%%%%%%%%%%%%%%%%%%%%%%%%%%

%%%%%%%%%%%%%%%%%%%%%%%%%%%%%%%%%%%%%%%%%%%%%%%%%%%%%%%%%%%%%%%%%%%%%%%%%%%%%%%%
% Figure: ECD, DSEC Qualitative Evaluations

%%%%%%%%%%%%%%%%%%%%%%%%%%%%%%%%%%%%%%%%%%%%%%%%%%%%%%%%%%%%%%%%%%%%%%%%%%%%%%%%

% ------------------------------------------------------------------------------
% CONCLUSION
% ------------------------------------------------------------------------------
\section{Conclusion}
\label{sec:conclusion}
In this work we introduced an approach to estimate dense optical flow from
events and edges obtained from image frames. Our method provides insight on how
to take full advantage of bi-modal data from publicly available datasets or
data collected from event cameras such as the DAVIS 346B and its variants.
Moreover, we revisited and refined key components for a successful application
of CM for optical flow estimation without the need to convert events into
voxel-like representations or have ground-truth signals. Experimental results
demonstrate that our framework can accurately model the flow and motion
trajectories of events with the additional edge-consistency imposition.
However, since our approach leverages frame data, low-quality images and the
registration problem between events and images can involve more effort in the
preprocessing step to extract viable edges. Our method also does not directly
solve the aperture problem and may struggle in places where events are not due
to motion or are extremely noisy. Future work will further enhance our hybrid
event-frame CM framework and allow for practical combinations of event cameras
with conventional cameras.

% ------------------------------------------------------------------------------
% ACKNOWLEDGMENTS 
% ------------------------------------------------------------------------------
\section*{Acknowledgments}
This material is based upon work supported by the Air Force Research Laboratory
under award number FA8571-23-C-0041.

% ------------------------------------------------------------------------------
% REFERENCES
{\small
\bibliographystyle{ieee_fullname}
\bibliography{secrets_of_edge-informed_contrast_maximization_for_event-based_vision}
}

% ------------------------------------------------------------------------------
% SUPPLEMENTARY MATERIAL 
% ------------------------------------------------------------------------------
\appendix
\section*{Supplementary Material}

% ------------------------------------------------------------------------------
% FULL RESULTS ON THE DSEC TEST SEQUENCES
% ------------------------------------------------------------------------------
\section{Full Results on the DSEC Test Sequences}
\label{sec:full_results_on_the_dsec_test_sequences}
\vspace{-1mm}
We provide a full report of our accuracy evaluation results on the DSEC
benchmark in \cref{tab:dsec_testset_accuracies_supp}. In addition, a complete
overview of the sharpness results in terms of flow warp loss (FWL) scores on
the DSEC test set are shown in \cref{tab:dsec_testset_fwls_supp}. At the time
of this writing, Liu \etal \cite{liu2023tma} had the best-known supervised
learning (SL) method on the DSEC-Flow benchmark in terms of accuracy. However,
Liu \etal \cite{liu2023tma} did not report their FWL scores. Conversely, Gehrig
\etal \cite{gehrig2021eraft} had the best-known SL method in terms of FWL
scores.

\cref{tab:dsec_testset_accuracies_supp} provides a summary of the accuracy
comparisons against these SL techniques as well as the best-known model-based
(MB) methods. Similarly, \cref{tab:dsec_testset_fwls_supp} summarizes the
comparisons of the FWL scores (sharpness). We note that no MB method, including
ours, produces accuracy scores comparable to state-of-the-art SL approaches on
the DSEC test set. Nonetheless, when compared to other state-of-the-art MB
methods our approach provides comparable average endpoint error (AEE) and
percentage $3$-pixel error (\%3PE). Additionally, our percentage $1$-pixel
error (\%1PE) scores are consistently better than other MB methods.
Interestingly, for \texttt{zurich\_city\_12\_a} (noisy),
\cite{brebion2021realtimeflow} performed better than others due to its event
denoising component.

%%%%%%%%%%%%%%%%%%%%%%%%%%%%%%%%%%%%%%%%%%%%%%%%%%%%%%%%%%%%%%%%%%%%%%%%%%%%%%%%
% Table: DSEC Accuracies
\begin{table}
    \centering
  
    \begin{adjustbox}{max width=\columnwidth}
        \begin{tabular}{@{\thinspace}c@{\medspace}l@{\thickspace}
        c@{\thickspace}c@{\thickspace}c@{\thickspace}c@{\thickspace} 
        c@{\thickspace}c@{\thickspace}c@{\thickspace}c@{\thickspace} 
        c@{\thickspace}c@{\thickspace}c@{\thickspace}c@{\thickspace}
        c@{\thickspace}c@{\thickspace}c} 
    
            % ================================================================================
            \toprule
            & &
            \multicolumn{3}{c}{All}                              & \thickspace &     % All 
            \multicolumn{3}{c}{\texttt{interlaken\_00\_b}}       & \thickspace &     % interlaken_00_b
            \multicolumn{3}{c}{\texttt{interlaken\_01\_a}}       & \thickspace &     % interlaken_01_a
            \multicolumn{3}{c}{\texttt{thun\_01\_a}}             \\                  % thun_01_a
  
            % --------------------------------------------------------------------------------
            \cmidrule(){3-5} \cmidrule(){7-9} \cmidrule(){11-13} \cmidrule(){15-17} 

            & &
            AEE $\downarrow$       & \%1PE $\downarrow$     & \%3PE $\downarrow$    & \thickspace &     % All
            AEE $\downarrow$       & \%1PE $\downarrow$     & \%3PE $\downarrow$    & \thickspace &     % interlaken_00_b
            AEE $\downarrow$       & \%1PE $\downarrow$     & \%3PE $\downarrow$    & \thickspace &     % interlaken_01_a
            AEE $\downarrow$       & \%1PE $\downarrow$     & \%3PE $\downarrow$    \\                  % thun_01_a 
            
            % --------------------------------------------------------------------------------
            \midrule 

            \multirow{2}{*}{\rotatebox[origin=c]{90}{SL}}
            & TMA \cite{liu2023tma} & 
            \textbf{0.74\tiny{3}}  & \textbf{10.86\tiny{3}} & \textbf{2.30\tiny{1}} & \thickspace &     % All
            \textbf{1.38\tiny{5}}  & \textbf{18.12}         & \textbf{5.78\tiny{5}} & \thickspace &     % interlaken_00_b
            \textbf{0.80\tiny{9}}  & \textbf{12.89\tiny{4}} & \textbf{3.10\tiny{8}} & \thickspace &     % interlaken_01_a
            \textbf{0.61\tiny{6}}  & \textbf{8.84\tiny{4}}  & \textbf{1.60\tiny{5}} \\                  % thun_01_a 
            & E-RAFT \cite{gehrig2021eraft} & 
            0.78\tiny{8}           & 12.74\tiny{2}          & 2.68\tiny{4}          & \thickspace &     % All
            1.39\tiny{4}           & 20.41\tiny{5}          & 6.18\tiny{9}          & \thickspace &     % interlaken_00_b
            0.89\tiny{9}           & 15.48\tiny{3}          & 3.90\tiny{7}           & \thickspace &     % interlaken_01_a
            0.65\tiny{4}           & 10.95\tiny{4}          & 1.87                  \\                  % thun_01_a 
            % --------------------------------------------------------------------------------
            \midrule 

            \multirow{3}{*}{\rotatebox[origin=c]{90}{MB}}
            & Brebion \etal \cite{brebion2021realtimeflow} & 
            4.88\tiny{1}          & 82.81\tiny{2}           & 41.95\tiny{2}         & \thickspace &     % All
            8.58\tiny{8}          & 90.12                   & 59.84\tiny{1}         & \thickspace &     % interlaken_00_b
            5.94                  & 86.63                   & 47.33                 & \thickspace &     % interlaken_01_a
            3.01                  & 71.66\tiny{3}           & 29.69\tiny{7}         \\                  % thun_01_a 
            & Shiba \etal \cite{shiba2022secrets} & 
            \uln{3.47\tiny{2}}    & 76.57                   & \uln{30.85\tiny{5}}   & \thickspace &     % All
            \uln{5.74}            & 78.08\tiny{6}           & \uln{38.92\tiny{5}}   & \thickspace &     % interlaken_00_b
            \uln{3.74}            & 75.40\tiny{2}           & \uln{31.36\tiny{6}}   & \thickspace &     % interlaken_01_a
            2.12                  & 64.73                   & 17.68\tiny{4}         \\                  % thun_01_a 
            & Ours (EINCM) & 
            5.00\tiny{3}          & \uln{68.66\tiny{8}}     & 35.87\tiny{2}         & \thickspace &     % All
            6.39\tiny{6}          & \uln{72.63}             & 43.6                  & \thickspace &     % interlaken_00_b
            5.48\tiny{2}          & \uln{70.00\tiny{8}}     & 41.32\tiny{8}         & \thickspace &     % interlaken_01_a
            \uln{2.01\tiny{5}}    & \uln{51.83\tiny{2}}     & \uln{16.17\tiny{4}}    \\                  % thun_01_a 
            
            % ================================================================================
            \toprule

            & &
            \multicolumn{3}{c}{\texttt{thun\_01\_b}}             & \thickspace &     % thun_01_b
            \multicolumn{3}{c}{\texttt{zurich\_city\_12\_a}}     & \thickspace &     % zurich_city_12_a
            \multicolumn{3}{c}{\texttt{zurich\_city\_14\_c}}     & \thickspace &     % zurich_city_14_c
            \multicolumn{3}{c}{\texttt{zurich\_city\_15\_a}}     \\                  % zurich_city_15_a    
            % --------------------------------------------------------------------------------
            \cmidrule(){3-5} \cmidrule(){7-9} \cmidrule(){11-13} \cmidrule(){15-17} 

            & &
            AEE $\downarrow$       & \%1PE $\downarrow$      & \%3PE $\downarrow$    & \thickspace &     % thun_01_b 
            AEE $\downarrow$       & \%1PE $\downarrow$      & \%3PE $\downarrow$    & \thickspace &     % zurich_city_12_a
            AEE $\downarrow$       & \%1PE $\downarrow$      & \%3PE $\downarrow$    & \thickspace &     % zurich_city_14_c
            AEE $\downarrow$       & \%1PE $\downarrow$      & \%3PE $\downarrow$    \\                  % zurich_city_15_a 
            % --------------------------------------------------------------------------------
            \midrule

            \multirow{2}{*}{\rotatebox[origin=c]{90}{SL}}
            & TMA \cite{liu2023tma} & 
            \textbf{0.55\tiny{2}}  & \textbf{7.44\tiny{9}}   & \textbf{1.31}         & \thickspace &     % thun_01_b 
            \textbf{0.57\tiny{2}}  & \textbf{9.6}            & \textbf{8.66}         & \thickspace &     % zurich_city_12_a
            \textbf{0.65\tiny{7}}  & \textbf{14.10\tiny{7}}  & \textbf{1.99}         & \thickspace &     % zurich_city_14_c
            \textbf{0.55\tiny{4}}  & \textbf{6.95\tiny{4}}   & \textbf{1.07\tiny{9}} \\                  % zurich_city_15_a 
            & E-RAFT \cite{gehrig2021eraft} & 
            0.57\tiny{7}           & 8.322                   & 1.52                  & \thickspace &     % thun_01_b 
            0.61\tiny{2}           & 11.24                   & 1.05\tiny{7}          & \thickspace &     % zurich_city_12_a
            0.71\tiny{3}           & 15.5                    & 1.91\tiny{3}          & \thickspace &     % zurich_city_14_c
            0.58\tiny{9}           & 8.74\tiny{8}            & 1.30\tiny{3}          \\                  % zurich_city_15_a 
            % --------------------------------------------------------------------------------
            \midrule 

            \multirow{3}{*}{\rotatebox[origin=c]{90}{MB}}
            & Brebion \etal \cite{brebion2021realtimeflow} & 
            3.91\tiny{3}           & 77.56\tiny{7}           & 34.69                 & \thickspace &     % thun_01_b 
            \uln{3.13\tiny{9}}     & 80.27\tiny{7}           & 34.07\tiny{8}         & \thickspace &     % zurich_city_12_a
            3.99\tiny{8}           & 88.30\tiny{4}           & 45.67                 & \thickspace &     % zurich_city_14_c
            3.78\tiny{1}           & 81.35\tiny{3}           & 37.98\tiny{7}         \\                  % zurich_city_15_a
            & Shiba \etal \cite{shiba2022secrets} &
            \uln{2.48}             & 73.63\tiny{2}           & \uln{23.56\tiny{4}}   & \thickspace &     % thun_01_b 
            3.86                   & 86.39\tiny{8}           & \uln{43.96\tiny{1}}   & \thickspace &     % zurich_city_12_a
            \uln{2.72}             & 76.85\tiny{1}           & \uln{30.53}           & \thickspace &     % zurich_city_14_c
            \uln{2.35}                   & 72.86\tiny{4}           & \uln{20.98\tiny{7}}   \\                  % zurich_city_15_a
            & Ours (EINCM) & 
            2.77\tiny{8}           & \uln{63.63\tiny{3}}     & 26.56\tiny{ }         & \thickspace &     % thun_01_b 
            8.37                   & \uln{79.59\tiny{7}}     & 45.78\tiny{6}         & \thickspace &     % zurich_city_12_a
            3.15\tiny{3}           & \uln{64.68\tiny{7}}     & 30.87\tiny{9}         & \thickspace &     % zurich_city_14_c
            3.00\tiny{5}           & \uln{62.19\tiny{9}}     & 26.63\tiny{3}         \\                  % zurich_city_15_a
            % --------------------------------------------------------------------------------
            \bottomrule
        
        \end{tabular}
    \end{adjustbox}
    \caption{DSEC test set accuracy results. \emph{Bold} and \emph{underline}
    typefaces indicate the best among supervised learning and model-based
    methods, respectively.}
    \label{tab:dsec_testset_accuracies_supp}
\end{table}

%%%%%%%%%%%%%%%%%%%%%%%%%%%%%%%%%%%%%%%%%%%%%%%%%%%%%%%%%%%%%%%%%%%%%%%%%%%%%%%%

%%%%%%%%%%%%%%%%%%%%%%%%%%%%%%%%%%%%%%%%%%%%%%%%%%%%%%%%%%%%%%%%%%%%%%%%%%%%%%%%
% Table: DSEC FWLs
\begin{table}
    \centering
    \begin{adjustbox}{max width=\columnwidth}
        \begin{tabular}{@{\thinspace}c@{\medspace}l@{\thickspace}
        c@{\thickspace}c@{\thickspace} 
        c@{\thickspace}c@{\thickspace} 
        c@{\thickspace}c@{\thickspace} 
        c@{\thickspace}c@{\thickspace} 
        c@{\thickspace}c@{\thickspace} 
        c@{\thickspace}c@{\thickspace} 
        c@{\thickspace}c@{\thickspace} 
        c}
    
            % ================================================================================
            \toprule
            & &
            \multicolumn{1}{c}{All}                       & \thickspace &     % All
            \multicolumn{1}{c}{\texttt{int\_00\_b}}       & \thickspace &     % interlaken_00_b
            \multicolumn{1}{c}{\texttt{int\_01\_a}}       & \thickspace &     % interlaken_01_a
            \multicolumn{1}{c}{\texttt{thu\_01\_a}}       & \thickspace &     % thun_01_a
            \multicolumn{1}{c}{\texttt{thu\_01\_b}}       & \thickspace &     % thun_01_b
            \multicolumn{1}{c}{\texttt{zur\_12\_a}}       & \thickspace &     % zurich_city_12_a
            \multicolumn{1}{c}{\texttt{zur\_14\_c}}       & \thickspace &     % zurich_city_14_c
            \multicolumn{1}{c}{\texttt{zur\_15\_a}}       \\                  % zurich_city_15_a    
            % --------------------------------------------------------------------------------
            \cmidrule(){3-3} \cmidrule(){5-5} \cmidrule(){7-7} \cmidrule(){9-9} \cmidrule(){11-11} \cmidrule(){13-13} \cmidrule(){15-15} \cmidrule(){17-17} 

            & &
            FWL $\uparrow$                                & \thickspace &     % All
            FWL $\uparrow$                                & \thickspace &     % interlaken_00_b
            FWL $\uparrow$                                & \thickspace &     % interlaken_01_a
            FWL $\uparrow$                                & \thickspace &     % thun_01_a 
            FWL $\uparrow$                                & \thickspace &     % thun_01_b 
            FWL $\uparrow$                                & \thickspace &     % zurich_city_12_a
            FWL $\uparrow$                                & \thickspace &     % zurich_city_14_c
            FWL $\uparrow$                                \\                  % zurich_city_15_a 
            % --------------------------------------------------------------------------------
            \midrule 

            \multirow{1}{*}{\rotatebox[origin=c]{90}{SL}}
            & E-RAFT \cite{gehrig2021eraft} & 
            1.29                                          & \thickspace &     % All
            1.32                                          & \thickspace &     % interlaken_00_b
            1.42                                          & \thickspace &     % interlaken_01_a
            1.20                                          & \thickspace &     % thun_01_a 
            1.18                                          & \thickspace &     % thun_01_b 
            1.12                                          & \thickspace &     % zurich_city_12_a
            1.47                                          & \thickspace &     % zurich_city_14_c
            1.34                                          \\                  % zurich_city_15_a 
            \midrule
            
            \multirow{2}{*}{\rotatebox[origin=c]{90}{MB}}
            & Shiba \etal \cite{shiba2022secrets} & 
            1.36                                          & \thickspace &     % All
            1.50                                          & \thickspace &     % interlaken_00_b
            1.51                                          & \thickspace &     % interlaken_01_a
            1.24                                          & \thickspace &     % thun_01_a 
            1.24                                          & \thickspace &     % thun_01_b 
            1.14                                          & \thickspace &     % zurich_city_12_a
            1.50                                          & \thickspace &     % zurich_city_14_c
            1.41                                          \\                  % zurich_city_15_a 
            & Ours (EINCM) & 
            \textbf{1.61\tiny{5}}                         & \thickspace &     % All
            \textbf{1.94}                                 & \thickspace &     % interlaken_00_b
            \textbf{1.86\tiny{8}}                         & \thickspace &     % interlaken_01_a
            \textbf{1.40}                                 & \thickspace &     % thun_01_a 
            \textbf{1.39\tiny{6}}                         & \thickspace &     % thun_01_b 
            \textbf{1.28\tiny{9}}                         & \thickspace &     % zurich_city_12_a
            \textbf{1.60\tiny{5}}                         & \thickspace &     % zurich_city_14_c
            \textbf{1.60\tiny{3}}                         \\                  % zurich_city_15_a
            % ================================================================================
            \bottomrule
        \end{tabular}
    \end{adjustbox}
    \caption{DSEC test set sharpness results (FWL scores). \emph{Bold} typeface
    is used to indicate the \textbf{best}.}
    \label{tab:dsec_testset_fwls_supp}
\end{table}

%%%%%%%%%%%%%%%%%%%%%%%%%%%%%%%%%%%%%%%%%%%%%%%%%%%%%%%%%%%%%%%%%%%%%%%%%%%%%%%%

% ------------------------------------------------------------------------------
% ADDITIONAL SHARPNESS RESULTS ON MVSEC
% ------------------------------------------------------------------------------
\vspace{-1mm}
\section{Additional Sharpness Results on MVSEC}
\label{sec:additional_shaprness_results_on_mvsec}
\vspace{-1mm}
For the $dt=1$ setting on MVSEC, each data sample contains very few events
($\approx6.5$ K, 9.4 K, 7.8 K, and 8.7 K on average in
\texttt{indoor\_flying1}, \texttt{indoor\_flying2}, \texttt{indoor\_flying3},
and \texttt{outdoor\_day1}, respectively). In this scenario, MultiCM
\cite{shiba2022secrets} reported (sharpness) FWL scores of $\approx$1 for each
sequence. We report further comparisons for the MVSEC $dt=1$ case with exact
FWL scores in \cref{tab:mvsec_fwls_supp}. The FWL scores of MultiCM were
obtained using the open-source code provided by the authors. We observe that
although small, the FWL scores for both indoor and outdoor sequences were all
$>1$ and better than MultiCM. We also note that the average FWL score for
\texttt{indoor\_flying2} is higher than other sequences, which can be
correlated with it comprising a larger average number of events.

%%%%%%%%%%%%%%%%%%%%%%%%%%%%%%%%%%%%%%%%%%%%%%%%%%%%%%%%%%%%%%%%%%%%%%%%%%%%%%%%
% Table: MVSEC FWLS SUPP
\begin{table}
    \centering
    \begin{adjustbox}{max width=\columnwidth}
        \begin{tabular}{@{\thinspace}l@{\thickspace}c@{\thickspace}
        c@{\thickspace}c@{\thickspace}c@{\thickspace}c@{\thickspace}c@{\thickspace}c@{\thickspace}c}
    
            % ================================================================================
            \toprule
            & \thickspace &
            \multicolumn{7}{c}{MVSEC ($dt$=1)}             \\                  % MVSEC
            % --------------------------------------------------------------------------------
            \cmidrule(){3-9} %\cmidrule(){11-11} \cmidrule(){13-15} 

            & \thickspace & 
            \texttt{indoor\_flying1}                             & \thickspace &     % indoor_flying1
            \texttt{indoor\_flying2}                             & \thickspace &     % indoor_flying2
            \texttt{indoor\_flying3}                             & \thickspace &     % indoor_flying2 
            \phantom{-}\texttt{outdoor\_day1}\phantom{-}         \\                  % outdoor_day1 
            % --------------------------------------------------------------------------------
            \midrule 
            
            Ground truth & \thickspace &
            1.02\tiny{6}                                  & \thickspace &     % indoor_flying1
            0.98\tiny{6}                                  & \thickspace &     % indoor_flying2
            1.00\tiny{6}                                  & \thickspace &     % indoor_flying3 
            0.99\tiny{6}                                  \\                  % outdoor_day1
            Shiba \etal \cite{shiba2022secrets} & \thickspace &
            1.01\tiny{9}                                  & \thickspace &     % indoor_flying1
            0.96\tiny{8}                                  & \thickspace &     % indoor_flying2
            0.98\tiny{9}                                  & \thickspace &     % indoor_flying3 
            0.98\tiny{5}                                  \\                  % outdoor_day1 
            Ours (EINCM) & \thickspace &
            \textbf{1.03\tiny{4}}                         & \thickspace &     % indoor_flying1
            \textbf{1.16\tiny{1}}                         & \thickspace &     % indoor_flying2
            \textbf{1.03\tiny{8}}                         & \thickspace &     % indoor_flying3 
            \textbf{1.00\tiny{3}}                         \\                  % outdoor_day1 
            % ================================================================================
            \bottomrule
        \end{tabular}
    \end{adjustbox}
    \caption{Flow warp loss (FWL) for MVSEC sequences with $dt=1$ on grayscale
    frames. \emph{Bold} typeface indicates the \textbf{best}.}
    \label{tab:mvsec_fwls_supp}
\end{table}

%%%%%%%%%%%%%%%%%%%%%%%%%%%%%%%%%%%%%%%%%%%%%%%%%%%%%%%%%%%%%%%%%%%%%%%%%%%%%%%%

% ------------------------------------------------------------------------------
% MVSEC OUTDOOR EVALUATIONS 
% ------------------------------------------------------------------------------
\vspace{-1mm}
\section{MVSEC Outdoor Evaluations}
\label{sec:mvsec_outdoor_evaluations}
\vspace{-1mm}
The MVSEC outdoor sequence \texttt{outdoor\_day1} consists of 11,440 image
frames. Yet, optical flow is only evaluated on a small subset of this sequence.
To compare their results with UnFlow \cite{meister2018unflow}, Zhu \etal
\cite{zhu2018evflownet} evaluated on 800 frames from \texttt{outdoor\_day1}
spanning a time window from $222.4\,$s to $240.4\,$s. These start and end
times, interpreted as image timestamps, correspond to
$1,506,118,124.733064\,$s and $1,506,118,142.7177844\,$s,
respectively. Equivalently, interpreted as image indices, they correspond to
the 10,138\textsuperscript{th} and the 10,958\textsuperscript{th} (with
starting index 0), respectively. Following Zhu \etal \cite{zhu2018evflownet},
other works that benchmarked their evaluations on \texttt{outdoor\_day1} fall
short on consistently reporting and/or using the same evaluation points. To the
authors' knowledge, there are at least two sets of evaluation points for the
MVSEC \texttt{outdoor\_day1} sequence in the literature.  

% ------------------------------------------------------------------------------
\subsection{Discrepancies}
\label{subsec:discrepancies}
We summarize discrepancies in prior works as follows.
\begin{itemize}
    \item Although Zhu \etal \cite{zhu2018evflownet} reported a usage of 800
    frames, the provided timestamps indicate 820 frames instead. On the other
    hand, their publicly available code and assets suggest the use of exactly
    800 frames.
    \item Lee \etal \cite{lee2020spikeflownet} and Ding \etal
    \cite{ding2022steflownet} used two sets of 401 frames, one between the
    image indices $[9200, 9600]$ and the other between $[10500, 10900]$.
    \item Shiba \etal \cite{shiba2022secrets} mentioned using the same 800
    frames as \cite{zhu2018evflownet}. However, the reported results were not
    reasonably reproducible on our local machine. Therefore, in Tab.~1 of the
    main paper, the accuracy scores for \cite{shiba2022secrets} were obtained
    by running their code locally on the 800 frames as suggested by
    \cite{zhu2018evflownet}. This corresponds to image indices 10,148 to
    10,948.
\end{itemize}
Our evaluations on MVSEC \texttt{outdoor\_day1} were performed on the 800
frames corresponding to the image indices $[10148, 10948]$ (starting at 0).

% ------------------------------------------------------------------------------
% EDGE SMOOTHING SENSITIVITY ANALYSIS
% ------------------------------------------------------------------------------
\vspace{-1mm}
\section{Edge Smoothing Sensitivity Analysis}
\label{sec:edge_smoothing_sensitivity_analysis}
\vspace{-1mm}
%%%%%%%%%%%%%%%%%%%%%%%%%%%%%%%%%%%%%%%%%%%%%%%%%%%%%%%%%%%%%%%%%%%%%%%%%%%%%%%%
% Figure: Edge Smoothing
\begin{figure}
    \centering
    \begin{adjustbox}{max width=\columnwidth}
    \begin{tabular}{@{}c@{\thinspace}c@{\thinspace}c@{\thinspace}c@{\thinspace}c@{\thinspace}c@{\thinspace}c@{}}
    % \begin{tabular}{@{}c@{}} \begin{adjustbox}{max width=0.25\textwidth} Original \end{adjustbox}\end{tabular} & 
    % \begin{tabular}{@{}c@{}} \begin{adjustbox}{max width=0.25\textwidth} $k=1$  \end{adjustbox}\end{tabular} &
    % \begin{tabular}{@{}c@{}} \begin{adjustbox}{max width=0.25\textwidth} $k=5$    \end{adjustbox}\end{tabular} &
    % \begin{tabular}{@{}c@{}} \begin{adjustbox}{max width=0.25\textwidth} IEDT     \end{adjustbox}\end{tabular} \\

    \begin{tabular}{@{}c@{}} \includegraphics[width=0.25\textwidth, cfbox=gray 0.1pt 0pt]{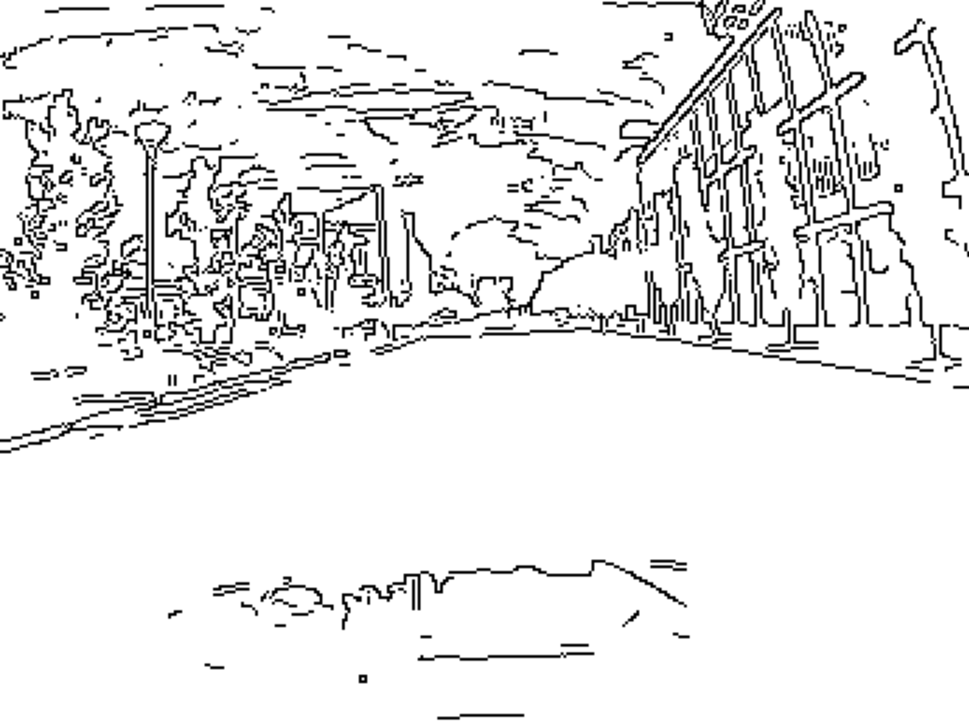} \end{tabular} &
    \begin{tabular}{@{}c@{}} \includegraphics[width=0.25\textwidth, cfbox=gray 0.1pt 0pt]{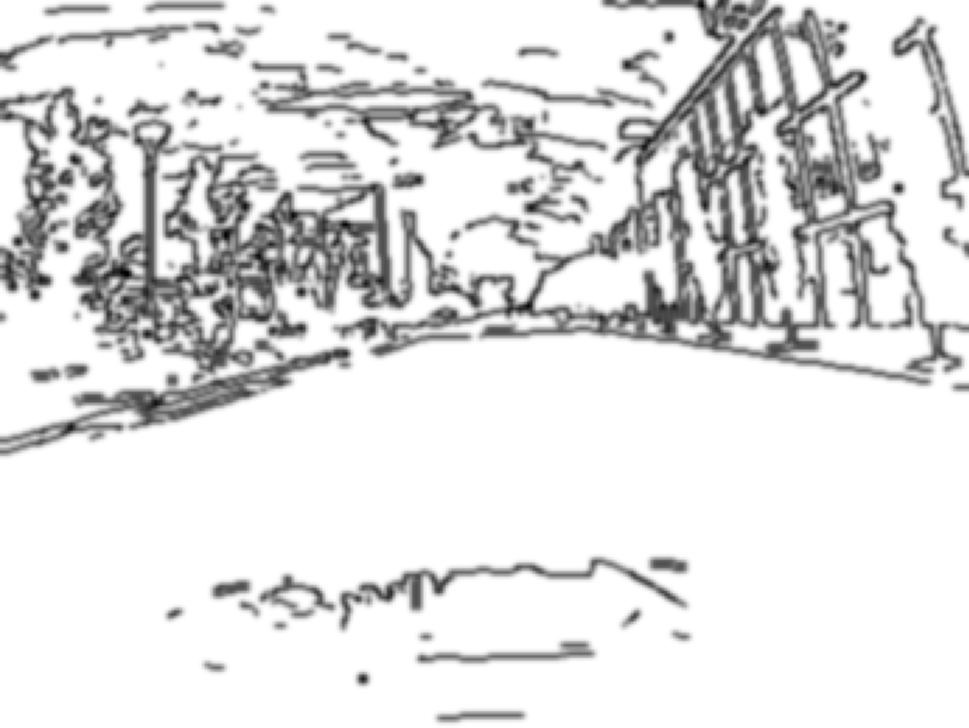} \end{tabular} &
    \begin{tabular}{@{}c@{}} \includegraphics[width=0.25\textwidth, cfbox=gray 0.1pt 0pt]{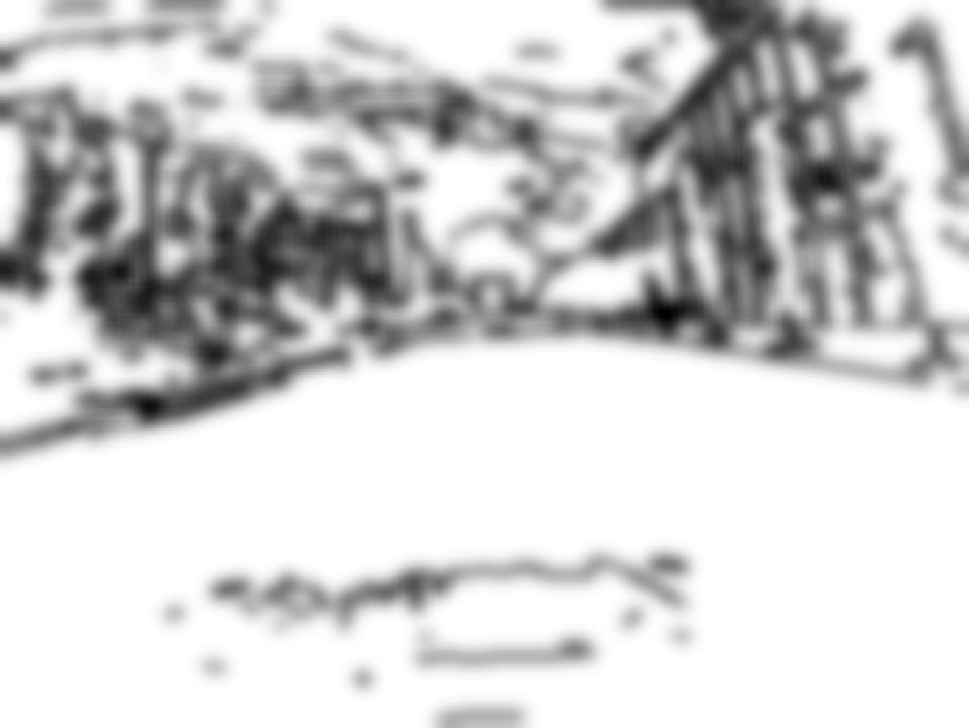} \end{tabular} &
    \begin{tabular}{@{}c@{}} \includegraphics[width=0.25\textwidth, cfbox=gray 0.1pt 0pt]{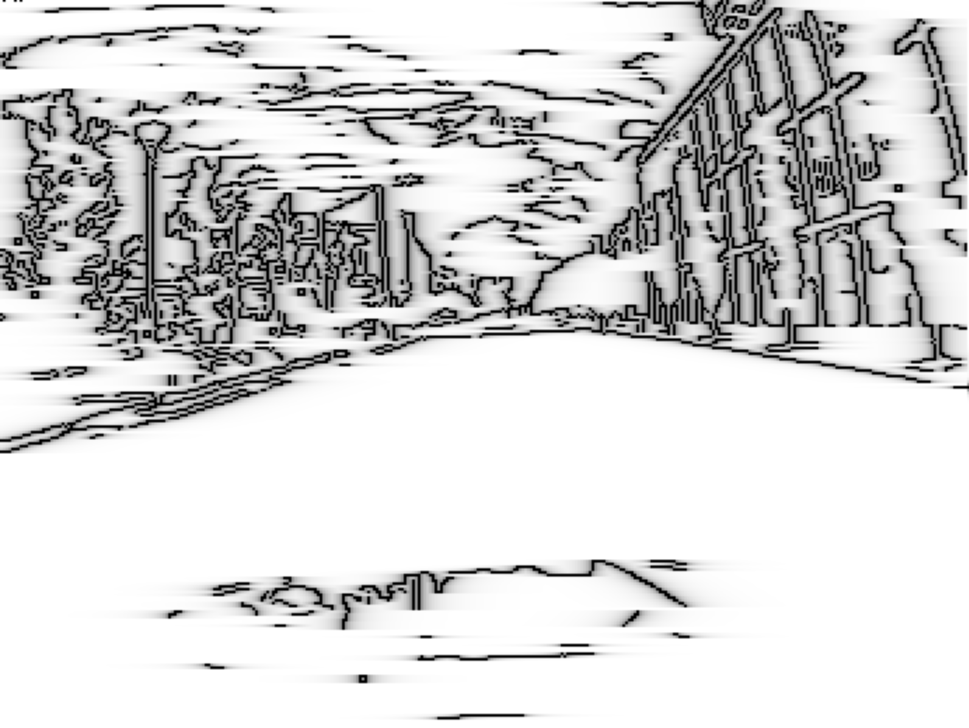} \end{tabular} \\
    
    \begin{tabular}{@{}c@{}} (a) Original \end{tabular} &
    \begin{tabular}{@{}c@{}} (b) $k=1$ \end{tabular} &
    \begin{tabular}{@{}c@{}} (c) $k=5$ \end{tabular} &
    \begin{tabular}{@{}c@{}} (d) IEDT \end{tabular} \\
    
    \end{tabular}
    \end{adjustbox}
    \caption{Edge smoothing operations.}
    \label{fig:edge_smoothing}
\end{figure}

%%%%%%%%%%%%%%%%%%%%%%%%%%%%%%%%%%%%%%%%%%%%%%%%%%%%%%%%%%%%%%%%%%%%%%%%%%%%%%%%

In \cref{tab:edge_smooth_sensitivity}, we present a sensitivity analysis on the
choice of edge smoothing methods. Observe that we obtained the best performance
by using a Gaussian kernel size of $k=1$ (\cref{fig:edge_smoothing}).
Increasing the kernel size to $k=5$ resulted in enlarging the reach of an edgel
to non-edge pixel regions. Yet, it also simultaneously increased the softness
of the edgels, which resulted in performance degradation. The inverse
exponential distance transform (IEDT) \cite{brebion2021realtimeflow} can smooth
edges in a manner where the reach of edgels can be extended to the non-edge
pixel regions without softening the edgel itself. Edges smoothed using the IEDT
yielded better performance when compared to Gaussian blurring with $k=5$. Note
that the IWEs for all three settings were consistently obtained using $k=1$.
Nevertheless, the IEDT is computationally expensive (\cref{tab:run_times}).
Consequently, we used a Gaussian blur with $k=1$ for edge smoothing.

%%%%%%%%%%%%%%%%%%%%%%%%%%%%%%%%%%%%%%%%%%%%%%%%%%%%%%%%%%%%%%%%%%%%%%%%%%%%%%%%
% Table: Edge Sensitivity
\begin{table}
    \centering
    \begin{adjustbox}{max width=0.5\columnwidth}
        \begin{tabular}{@{\thinspace}l@{\thickspace}c@{\thickspace}
        c@{\thickspace}c@{\thickspace}c@{\thickspace}c@{\thickspace}c}
    
            % ================================================================================
            \toprule
            & \thickspace &
            \multicolumn{5}{c}{\texttt{outdoor\_day1} ($dt=4$)}             \\                  % MVSEC
            % --------------------------------------------------------------------------------
            \cmidrule(){3-7}

            & \thickspace & 
            \multicolumn{1}{c}{$k=1$}                        & \thickspace &     % beta=5000
            \multicolumn{1}{c}{$k=5$}                        & \thickspace &     % beta=7000
            \multicolumn{1}{c}{IEDT}                       \\                  % beta=10000 
            % --------------------------------------------------------------------------------
            \midrule 
            
            AEE $\downarrow$ & \thickspace &
            1.70\tiny{4}                & \thickspace &     % beta=5000
            1.76\tiny{7}                & \thickspace &     % beta=7000
            1.73\tiny{6}                \\                  % beta=10000 
            \%3PE $\downarrow$ & \thickspace &
            16.01\tiny{3}                & \thickspace &     % beta=5000
            16.93                & \thickspace &     % beta=7000
            16.71\tiny{9}                \\                  % beta=10000 
            FWL $\uparrow$ & \thickspace &
            1.23                & \thickspace &     % beta=5000
            1.20\tiny{6}                         & \thickspace &     % beta=7000
            1.21\tiny{1}                \\                  % beta=10000 
            % ================================================================================
            \bottomrule
        
        \end{tabular}
    \end{adjustbox}
    \caption{Edge smoothing sensitivity analysis results. We report the
    accuracy and sharpness scores on the MVSEC sequence \texttt{outdoor\_day1}
    ($dt=4$). The first two columns depict a Gaussian blur with kernel size
    $k=1$ and $k=5$. The third column shows results using the inverse
    exponential distance transform (IEDT).}
    \label{tab:edge_smooth_sensitivity}
\end{table}

%%%%%%%%%%%%%%%%%%%%%%%%%%%%%%%%%%%%%%%%%%%%%%%%%%%%%%%%%%%%%%%%%%%%%%%%%%%%%%%%

% ------------------------------------------------------------------------------
% HYPERPARAMETERS
% ------------------------------------------------------------------------------
\vspace{-1mm}
\section{Hyperparameters}
\label{sec:hyperparameters}
\vspace{-1mm}
As discussed in the main paper, all the experiments used five pyramid levels to
take advantage of multiscaling. With regards to multiple references for MVSEC
$dt=1$, reference times $t_0$, $t_{\text{mid}}$, and $t_1$ were used to compute
contrasts, while the image timestamps $\mathcal{T}^{(i)}$ were utilized to
compute correlations. In the MVSEC $dt=4$ case, there were three images within
the duration of each data sample. Therefore, the image timestamps
$\mathcal{T}^{(i)}$ were used as reference times to compute both contrasts and
correlations. For the ECD sequence \texttt{slider\_depth}, $dt=2$ was chosen
(with on average $\approx24$ K events per data sample) for the evaluations.
Each data sample consisted of three images: two at the boundaries and one in
between. Contrasts and correlations were calculated at the three image
timestamps $\mathcal{T}^{(i)}$. Similarly, in the DSEC sequences each data
sample consisted of three images and the timestamps $\mathcal{T}^{(i)}$ served
as reference times for computing both contrasts and correlations.

The accuracy and FWL scores were evaluated for each sequence using the
corresponding events within a data sample. However, for optimization we ensured
a fixed number of events per data sample $\mathcal{D}^{(i)}$. Specifically, we
used 30 K and 40 K events for the indoor and outdoor sequences from MVSEC,
respectively. For DSEC and ECD, we used 1.5 M and 30 K events, respectively.
For the MVSEC sequences, we set $\alpha=20, \beta=35$, for ECD we used
$\alpha=60, \beta=60$, and for DSEC $\alpha=2000, \beta=4000$ were used.

Extracting image edges via OpenCV's
\texttt{Canny}\footnote{\href{https://docs.opencv.org/4.x/da/d22/tutorial\_py\_canny.html}{https://docs.opencv.org/4.x/da/d22/tutorial\_py\_canny.html}}
involves using a pair of threshold values $(\texttt{thresh\_1},
\texttt{thresh\_2})$. We used $(100, 200)$ and $(30, 80)$ for the MVSEC indoor
and outdoor sequences, respectively. For ECD, $(100, 200)$ was used. Finally,
for DSEC $(30, 80)$ was used for all sequences except for
\texttt{zurich\_city\_12\_a} (night-time images with extremely noisy events),
where the thresholds $(10, 60)$ were used. The coefficient $\gamma$ for the
regularizer term in our objective function was fixed to $0.0025$ for the MVSEC
sequences, while it was set to $0.0$ for both the ECD and DSEC sequences.

% ------------------------------------------------------------------------------
% ENINCM MULTISCALE PSEUDOCODE
% ------------------------------------------------------------------------------
\vspace{-1mm}
\section{EINCM Multiscale Pseudocode}
\label{sec:eincm_multiscale_pseudocode}
\vspace{-1mm}
In \cref{alg:eincm_multiscale_pseudocode}, we present the high-level pseudocode
of the multiscaling scheme used by our method. The $i$-th input data sample
$\mathcal{D}^{(i)}$ consists of the corresponding events $\mathcal{E}^{(i)}$,
edge images $\mathcal{I}^{(i)}$, and image timestamps $\mathcal{T}^{(i)}$. The
outer loop (lines 5-19) reflects the fact that we used five scales in the
multiscale scheme. The number of scales as well as the resolution of the motion
parameters at each scale are preset and can be adjusted. The main contrast and
correlation maximization (CCM, line 6), where we optimize for the motion
parameters, requires a loss function and an initial
$\prescript{}{l}{\boldsymbol{\Theta}_i}$ (\ie, the first argument). To solve
for \textit{handovers} (line 12), we essentially solve for the coefficient
$w_{\text{ho}}$. This coefficient linearly combines the optimized parameters at
the current index and scale (result of line 6), and the downsampled optimized
parameters from the previous index at the current scale (result of line 7). We
optimize for $w_{\text{ho}}$ in the same manner as the main CCM optimization
where we replace $\prescript{}{l}{\boldsymbol{\Theta}_i}$ by the aforementioned
weighted sum.

%%%%%%%%%%%%%%%%%%%%%%%%%%%%%%%%%%%%%%%%%%%%%%%%%%%%%%%%%%%%%%%%%%%%%%%%%%%%%%%%
% Algorithm: EINCM Multiscale Pseudo Code
\vspace{2mm}
\begin{adjustbox}{max width=\columnwidth}
\centering
\SetAlgoLined
\SetArgSty{textnormal}
\begin{minipage}{1.33\columnwidth}
\removelatexerror%
\begin{algorithm}[H]

    \SetKwInOut{Hparams}{Hyperparameters}
    \SetKw{In}{in}
    \SetKw{From}{from}
    \SetKw{To}{to}
    \KwData{
        $\mathcal{E}^{(i)}$, 
        $\mathcal{I}^{(i)}$, 
        $\mathcal{T}^{(i)}$, 
        and optionally $\fourIdx{}{0}{\ast}{i-1}{\boldsymbol{\Theta}}$\\}
    \Hparams{$\boldsymbol{a}$\\}
    \KwResult{$\fourIdx{}{0}{\ast}{i}{\boldsymbol{\Theta}}$\\}

    \uIf{$\fourIdx{}{0}{\ast}{i-1}{{\boldsymbol{\Theta}}}$ is available}{
    $\fourIdx{}{4}{0}{i}{\boldsymbol{\Theta}} \leftarrow 
        \texttt{downscale} (\fourIdx{}{0}{\ast}{i-1}{\boldsymbol{\Theta}})$ 
    }
    \uElse{
        $\fourIdx{}{4}{0}{i}{\boldsymbol{\Theta}} \leftarrow 
            \textbf{zero}$
    }
    
    \For{$\texttt{lvl} = 4$ \KwTo $0$}{
        
        % CM-opt 
        $\fourIdx{}{\texttt{lvl}}{\ast}{i}{\boldsymbol{\Theta}} \leftarrow 
        \argmax_{\fourIdx{}{\texttt{lvl}}{}{i}{\boldsymbol{\Theta}}} \texttt{loss}(\fourIdx{}{\texttt{lvl}}{0}{i}{\boldsymbol{\Theta}}; \mathcal{E}^{(i)}, \mathcal{I}^{(i)}, \mathcal{T}^{(i)})$ 

        % downscale preceding
        $\fourIdx{}{\texttt{lvl}}{\downarrow}{i-1}{\boldsymbol{\Theta}} \leftarrow 
        \texttt{downscale} (\fourIdx{}{0}{\ast}{i-1}{\boldsymbol{\Theta}})$ 

        $w_{\text{ho}} \leftarrow 0$

        \uIf{$\texttt{handover\_flag}_{\texttt{lvl}}$}{
            \uIf{$\texttt{solve\_flag}_{\texttt{lvl}}$}{
                $w_{\text{ho}}^{0} \leftarrow 0.5$
                
                %alpha-opt
                $w_{\text{ho}}^{\ast} \leftarrow 
                    \argmax_{w_{\text{ho}}}\texttt{loss}_{w_{\text{ho}}}(w_{\text{ho}}^{0}; 
                    \fourIdx{}{\texttt{lvl}}{\ast}{i}{\boldsymbol{\Theta}}, 
                    \fourIdx{}{\texttt{lvl}}{\downarrow}{i-1}{\boldsymbol{\Theta}}, 
                    \mathcal{E}^{(i)}, 
                    \mathcal{I}^{(i)}, 
                    \mathcal{T}^{(i)})$ 

                $w_{\text{ho}} \leftarrow w_{\text{ho}}^{\ast}$
            }
            \uElse{ 
                $w_{\text{ho}} \leftarrow \boldsymbol{a}$ 
            }
        }
        % \uIf{not $\texttt{handover\_flag}_{\texttt{lvl}}$ and  $\texttt{lvl} = 0$}{
        %     $w_{\text{ho}} \leftarrow 0$ 
        % }
        $\fourIdx{}{\texttt{lvl}}{\ast}{i}{{\boldsymbol{\Theta}}} \leftarrow 
            w_{\text{ho}} \cdot \fourIdx{}{\texttt{lvl}}{\downarrow}{i-1}{\boldsymbol{\Theta}} + 
            (1-w_{\text{ho}}) \cdot \fourIdx{}{\texttt{lvl}}{\ast}{i}{\boldsymbol{\Theta}}$ 

        %prepare for next level
        \uIf{$\texttt{lvl} \neq 0$}{
            $\fourIdx{}{\texttt{lvl}-1}{0}{i}{\boldsymbol{\Theta}} \leftarrow \texttt{upscale} (\fourIdx{}{\texttt{lvl}}{\ast}{i}{\boldsymbol{\Theta}})$  
        }
         
    }
    
    \Return $\fourIdx{}{0}{\ast}{i}{\boldsymbol{\Theta}}$
    
    \caption{EINCM Multiscale Pseudocode}
    \label{alg:eincm_multiscale_pseudocode}
\end{algorithm}
\end{minipage}
\end{adjustbox}

%%%%%%%%%%%%%%%%%%%%%%%%%%%%%%%%%%%%%%%%%%%%%%%%%%%%%%%%%%%%%%%%%%%%%%%%%%%%%%%%

% ------------------------------------------------------------------------------
% RUNTIME ANALYSIS
% ------------------------------------------------------------------------------
\section{Runtime Analysis}
\label{sec:runtime_analysis}
\vspace{-1mm}
In \cref{tab:run_times}, we present a detailed runtime report of our image
preprocessing as well as the optimization (including and excluding the first
\texttt{jit}\footnote{\href{https://jax.readthedocs.io/en/latest/\_autosummary/jax.jit.html}{https://jax.readthedocs.io/en/latest/\_autosummary/jax.jit.html}}
compilation) pipeline on the same machine and software suite described in the
main paper.

%%%%%%%%%%%%%%%%%%%%%%%%%%%%%%%%%%%%%%%%%%%%%%%%%%%%%%%%%%%%%%%%%%%%%%%%%%%%%%%%
% Table: Runtimes
\begin{table}
    \centering
    \begin{adjustbox}{max width=\columnwidth}
        \begin{tabular}{@{\thinspace}l@{\thickspace}c@{\thickspace}
        c@{\thickspace}c@{\thickspace}c@{\thickspace}c@{\thickspace}c}
    
            % ================================================================================
            \toprule
            
            & \thickspace & 
            \multicolumn{1}{c}{ECD ($176\times240$)}                        & \thickspace &     % ECD
            \multicolumn{1}{c}{MVSEC ($260\times346$)}                      & \thickspace &     % MVSEC
            \multicolumn{1}{c}{DSEC ($480\times640$)}                       \\                  % DSEC 
            % --------------------------------------------------------------------------------
            % \cmidrule(){3-3} \cmidrule(){5-5} \cmidrule(){7-7}

            % --------------------------------------------------------------------------------
            \midrule 
            
            Preprocessing & \thickspace &
            $17.4\,$ms $\pm$ $588\,\mu$s                 & \thickspace &     % ECD
            $33.7\,$ms $\pm$ $1.81\,\mu$s                & \thickspace &     % MVSEC
            $68.7\,$ms $\pm$ $3.59\,\mu$s                \\                  % DSEC
            Edge extraction & \thickspace &
            $146\,\mu$s $\pm$ $27.7\,\mu$s                 & \thickspace &     % ECD
            $162\,\mu$s $\pm$ $18.62\,\mu$s                & \thickspace &     % MVSEC
            $351\,\mu$s $\pm$ $46.1\,\mu$s                \\                  % DSEC
            Gaussian blur & \thickspace &
            $195\,\mu$s $\pm$ $13.9\,\mu$s                 & \thickspace &     % ECD
            $395\,\mu$s $\pm$ $29.6\,\mu$s                & \thickspace &     % MVSEC
            $1.6\,$ms $\pm$ $172\,\mu$s                \\                  % DSEC
            Inverse exponential distance transform & \thickspace &
            $755\,$ms $\pm$ $28.8\,$ms                 & \thickspace &     % ECD
            $1.56\,$s $\pm$ $21.6\,$ms                & \thickspace &     % MVSEC
            $5.36\,$s $\pm$ $86.3\,$ms                \\                  % DSEC
            CCM at pyramid level 0 (include first \texttt{jit} compilation) & \thickspace &
            $356\,$ms $\pm$ $1.02\,$s                 & \thickspace &     % ECD
            $465.59\,$ms $\pm$ $1.354\,$s                & \thickspace &     % MVSEC
            $2.35\,$s $\pm$ $3.65\,$s                \\                  % DSEC
            CCM at pyramid level 0 (exclude first \texttt{jit} compilation) & \thickspace &
            $15.96\,$ms $\pm$ $846.4\,\mu$s                 & \thickspace &     % ECD
            $32.44\,$ms $\pm$ $188.3\,\mu$s                & \thickspace &     % MVSEC
            $1.128\,$s $\pm$ $280.3\,\mu$s                \\                  % DSEC
            Downscale from pyramid level 4 to 0 & \thickspace &
            $96.4\,$ms $\pm$ $62.9\,$ms                 & \thickspace &     % ECD
            $96.4\,$ms $\pm$ $62.9\,$ms                & \thickspace &     % MVSEC
            $96.4\,$ms $\pm$ $62.9\,$ms                \\                  % DSEC
            Upscale to sensor size & \thickspace &
            $47.8\,$ms $\pm$ $39.5\,$ms                 & \thickspace &     % ECD
            $99.3\,$ms $\pm$ $33.2\,$ms                & \thickspace &     % MVSEC
            $122\,$ms $\pm$ $14.3\,$ms                \\                  % DSEC
            
            % ================================================================================
            \bottomrule
        \end{tabular}
    \end{adjustbox}
    \caption{The runtime details of the edge extraction pipeline (Fig.~2 in the
    main paper). This includes the following: (i) preprocessing, (ii) edge
    detection, and (iii) edge smoothing components, the optimization routine,
    and upscaling/downsampling routines.}
    \label{tab:run_times}
\end{table}

%%%%%%%%%%%%%%%%%%%%%%%%%%%%%%%%%%%%%%%%%%%%%%%%%%%%%%%%%%%%%%%%%%%%%%%%%%%%%%%%

\end{document}